\documentclass{article}

\usepackage{microtype}
\usepackage{graphicx}
\usepackage{booktabs} %

\usepackage{mmstyle}

\usepackage[utf8]{inputenc} %
\usepackage[T1]{fontenc}    %
\usepackage{hyperref}       %
\usepackage{url}            %
\usepackage{booktabs}       %
\usepackage{amsfonts}       %
\usepackage{nicefrac}       %
\usepackage{microtype}      %
\usepackage{wrapfig}
\usepackage{lipsum}

\usepackage{natbib}
\usepackage{graphicx}
\usepackage{subcaption}
\usepackage{amssymb}
\usepackage{amsmath}
\usepackage{amsthm}
\usepackage{bbm}
\usepackage{csquotes}
\usepackage{color}
\usepackage{xcolor}
\usepackage{mathtools}
\usepackage{multirow}
\usepackage[toc,page]{appendix}
\usepackage[font=small,labelfont=bf]{caption}
\usepackage{enumitem}

\usepackage{hyperref}

\usepackage[accepted]{icml2019}
\definecolor{mydarkblue}{rgb}{0,0.1,0.45}
\hypersetup{
    colorlinks=true,
    linkcolor=mydarkblue,
    citecolor=mydarkblue,
    filecolor=mydarkblue,
    urlcolor=mydarkblue}

\newcommand{\expect}{\mathbb{E}}

\newcommand{\loss}{\mathcal{L}}
\newcommand{\kldiv}{\mathrm{D}_{\mathrm{KL}}}
\newcommand{\newvec}{\mathrm{vec}}
\newcommand{\weight}{\mathbf{W}}
\newcommand{\fisher}{\mathbf{F}}
\newcommand{\kron}{\otimes}
\newcommand{\AMat}{\mathbf{A}}
\newcommand{\SMat}{\mathbf{S}}
\newcommand{\ba}{\mathbf{a}}
\newcommand{\bs}{\mathbf{s}}

\newcommand{\normal}{\mathcal{N}}

\newtheorem{exampp}{Example}

\usepackage{xcolor}
\usepackage{framed}
\colorlet{shadecolor}{gray!30}
\newenvironment{examp}
   {\begin{shaded}\begin{exampp}}
   {\end{exampp}\end{shaded}}

\icmltitlerunning{EigenDamage: Structured Pruning in the Kronecker-Factored Eigenbasis}

\begin{document}

\twocolumn[
\icmltitle{EigenDamage: Structured Pruning in the Kronecker-Factored Eigenbasis}

\icmlsetsymbol{equal}{*}

\begin{icmlauthorlist}
\icmlauthor{Chaoqi Wang}{uoft,vector}
\icmlauthor{Roger Grosse}{uoft,vector}
\icmlauthor{Sanja Fidler}{uoft,vector,nvidia}
\icmlauthor{Guodong Zhang}{uoft,vector}
\end{icmlauthorlist}

\icmlaffiliation{uoft}{Department of Computer Science, University of Toronto, Toronto, Canada}
\icmlaffiliation{vector}{Vector Institute, Toronto, Canada}
\icmlaffiliation{nvidia}{NVIDIA}

\icmlcorrespondingauthor{Chaoqi Wang}{cqwang@cs.toronto.edu}
\icmlcorrespondingauthor{Guodong Zhang}{gdzhang@cs.toronto.edu}

\icmlkeywords{Machine Learning, ICML}

\vskip 0.3in
]

\printAffiliationsAndNotice{}  %

\begin{abstract}
Reducing the test time resource requirements of a neural network while preserving test accuracy is crucial for running inference on resource-constrained devices. To achieve this goal, we introduce a novel network reparameterization based on the Kronecker-factored eigenbasis (KFE), and then apply Hessian-based structured pruning methods in this basis. As opposed to existing Hessian-based pruning algorithms which do pruning in parameter coordinates, our method works in the KFE where different weights are approximately independent, enabling accurate pruning and fast computation.
We demonstrate empirically the effectiveness of the proposed method through extensive experiments. In particular, we highlight that the improvements are especially significant for more challenging datasets and networks. 
With negligible loss of accuracy, an iterative-pruning version gives a 10$\times$ reduction in model size and a 8$\times$ reduction in FLOPs on wide ResNet32. Our code is available at \href{https://github.com/alecwangcq/EigenDamage-Pytorch}{here}.

\end{abstract}

\section{Introduction}
Deep neural networks exhibit good generalization behavior in the over-parameterized regime~\citep{zhang2016understanding, neyshabur2018towards}, where the number of network parameters exceeds the number of training samples. However, over-parameterization leads to high computational cost and memory overhead at test time, making it hard to deploy deep neural networks on a resource-limited device. 

Network pruning~\citep{lecun1990optimal, hassibi1993optimal, han2015learning, dong2017learning, zeng2019mlprune} has been identified as an effective technique to improve the efficiency of deep networks for applications with limited test-time computation and memory budgets. Without much loss in accuracy, classification networks can be compressed by a factor of 10 or even more~\citep{han2015learning, zeng2019mlprune} on ImageNet~\citep{deng2009imagenet}.
A typical pruning procedure consists of three stages: 1) train a large, over-parameterized model, 2) prune the trained model according to a certain criterion, and 3) fine-tune the pruned model to regain the lost performance. 
\begin{figure}
    \centering
    \vspace{-0.2cm}
    \includegraphics[width=0.9\columnwidth]{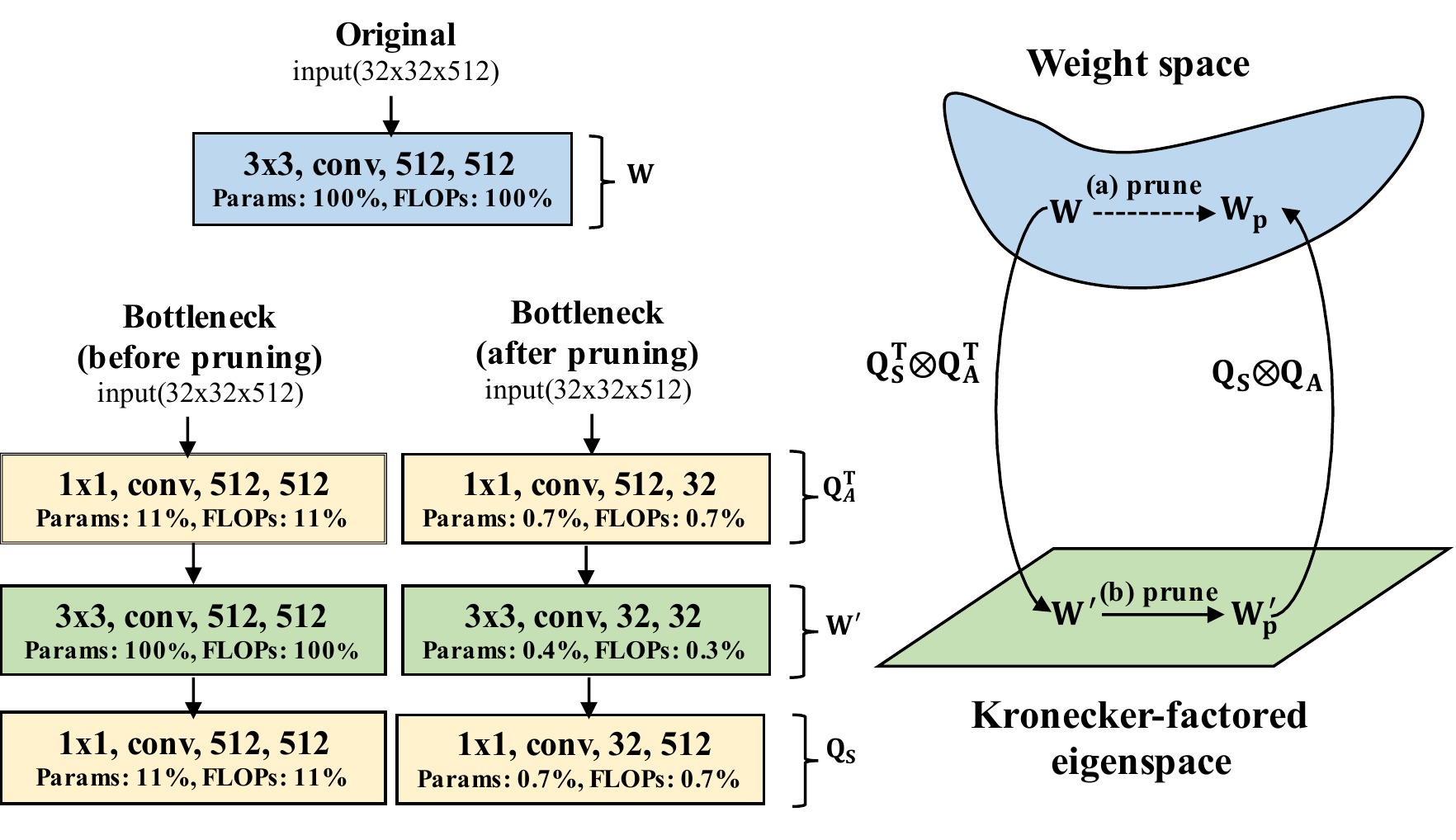}
    \vspace{-0.3cm}
    \caption{On the left-hand side, the proposed bottleneck structure before and after pruning. 
    The number after `Params' and `FLOPs' indicates the remaining portion compared to the original one. On the right-hand side, we highlight the differences of the pruning procedure between traditional methods (a) and our method (b).}
    \label{fig:figure_1}
    \vspace{-0.5cm}
\end{figure}

Most existing work on network pruning focuses on the second stage. A common idea is to select parameters for pruning based on weight magnitudes~\citep{hanson1989comparing, han2015learning}. However, weights with small magnitude are not necessarily unimportant~\citep{lecun1990optimal}. As a consequence, magnitude-based pruning might delete important parameters, or preserve unimportant ones. By contrast, Optimal Brain Damage~(OBD)~\citep{lecun1990optimal} and Optimal Brain Surgeon~(OBS)~\citep{hassibi1993optimal} prune weights based on the Hessian of the loss function; the advantage is that both criteria reflect the sensitivity of the cost to the weight. Though OBD and OBS have proven to be effective for shallow neural networks, it remains challenging to extend them for deep networks because of the high computational cost of computing second derivatives. To solve this issue, several approximations to the Hessian have been proposed recently which assume layerwise independence~\citep{dong2017learning} or Kronecker structure~\citep{zeng2019mlprune}.

All of the aforementioned methods prune individual weights, leading to non-structured architectures which do not enjoy computational speedups unless one employs dedicated hardware~\citep{han2016eie} and software, which is difficult and expensive in real-world applications~\citep{liu2018rethinking}. 
In contrast, structured pruning methods such as channel pruning~\citep{liu2017learning, li2016pruning} aim to preserve the convolutional structure by pruning at the level of channels or even layers, thus automatically enjoy computational gains even with standard software frameowrks and hardware. 

\begin{figure}
    \centering
    \begin{subfigure}[b]{0.3\columnwidth}
        \includegraphics[width=\textwidth]{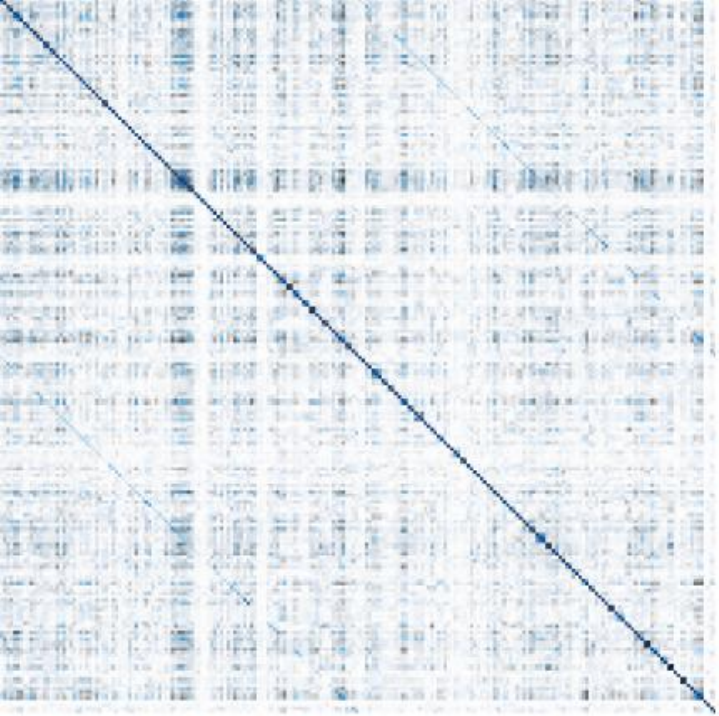}
    \end{subfigure}
    \hspace{0.5cm}
    \begin{subfigure}[b]{0.3\columnwidth}
        \includegraphics[width=\textwidth]{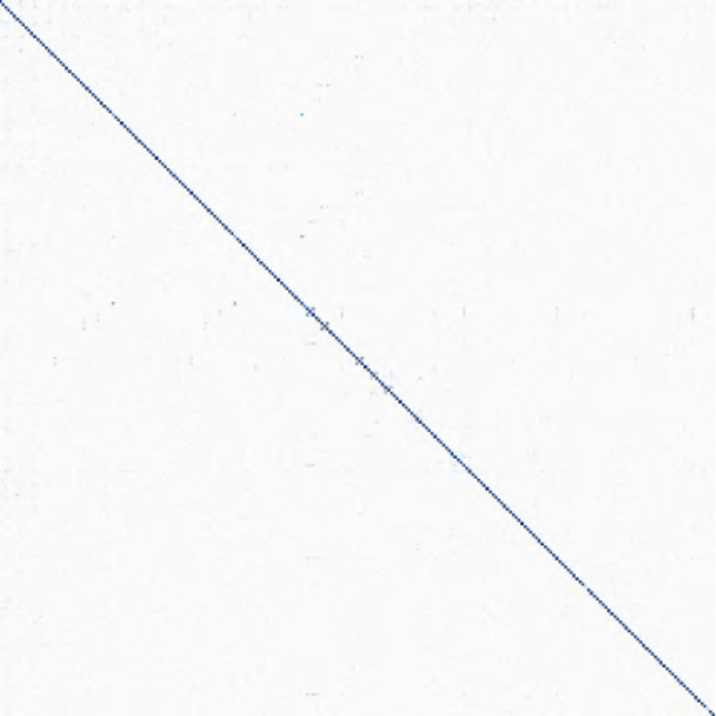}
    \end{subfigure}
    \vspace{-0.2cm}
    \caption{Fisher information matrices measured in initial parameter basis and in the KFE, computed from a small 3-layer ReLU MLP trained on MNIST. We only plot the block for the second layer. Note that we normalize the diagonal elements for visualization.}
    \label{fig:fisher_matrix}
    \vspace{-0.5cm}
\end{figure}

\textbf{Our Contributions.} In this work, we focus on structured pruning.
We first extend OBD and OBS to channel pruning, showing that they can match the performance of a state-of-the-art channel pruning algorithm~\citep{liu2017learning}. 
We then interpret them from the Bayesian perspective, showing that OBD and OBS each approximate the full-covariance Gaussian posterior with factorized Gaussians, but minimizing different variational objectives. 
However, different weights can be highly coupled in Bayesian neural network posteriors~(e.g., see Figure~\ref{fig:fisher_matrix}), suggesting that full-factorization assumptions may hurt the pruning performance. 

Based on this insight, we prune in a different coordinate system in which the posterior is closer to factorial. Specifically, we consider the Kronecker-factored eigenbasis (KFE)~\citep{george2018fast, bae2018eigenvalue}, in which the Hessian for a given layer is closer to diagonal. We propose a novel network reparameterization inspired by~\citet{desjardins2015natural} which explicitly parameterizes each layer in terms of the KFE.
Because the Hessian matrix is closer to diagonal in the KFE, we can apply OBD with less cost to prediction accuracy; we call this method EigenDamage.

Instead of sparse weight matrices, pruning in the KFE leads to a low-rank approximation, or bottleneck structure, in each layer (see Figure~\ref{fig:figure_1}). 
While most existing structured pruning methods~\citep{he2017channel, li2016pruning, liu2017learning, luo2017thinet} require specialized network architectures, EigenDamage can be applied to any fully connected or convolution layers without modifications. 
Furthermore, in contrast to traditional low-rank approximations~\citep{denton2014exploiting, lebedev2014speeding, jaderberg2014speeding} which minimize the Frobenius norm of the weight space error, EigenDamage is \emph{loss aware}.
As a consequence, the user need only choose a single compression ratio parameter, and EigenDamage can automatically determine an appropriate rank for each layer, and thus it is calibrated across layers. Empirically, EigenDamage outperforms strong baselines which do pruning in parameter coordinates, especially in more challenging datasets and networks.

\section{Background}
\label{sec:background_revisit}
In this section, we first introduce some background for understanding and reinterpreting Hessian-based weight pruning algorithms, and then briefly review structured pruning to provide context for the task that we will deal with.

\label{subsec:background}

\textbf{Laplace Approximation.} In general, we can obtain the Laplace approximation~\citep{mackay1992practical} by simply taking the second-order Taylor expansion around a local mode. %
For neural networks, we can find such modes with SGD.
Given a neural network with local MAP parameters $\vtheta^*$ after training on a dataset $\mathcal{D}$, we can obtain the Laplace approximation over the weights around $\vtheta^*$ by:
\begin{equation}\label{eq:laplace_approx}
    \log p(\vtheta|\mathcal{D}) \approx \log p(\vtheta^*|\mathcal{D}) - \frac{1}{2}(\vtheta-\vtheta^*)^\top\mH(\vtheta-\vtheta^*)
\end{equation}
where $\vtheta=[\newvec(\mW_1), ..., \newvec(\mW_L)]$, and $\mH$ is the Hessian matrix of the negative log posterior evaluated at $\vtheta^*$. 
Assuming $\mH$ is p.s.d., the Laplace approximation is equivalent to approximating the posterior over weights as a Gaussian distribution with $\vtheta^*$ and $\mH$ as the mean and precision, respectively. In practice, we can use the Fisher information matrix $\mF$ to approximate $\mH$, as done in~\citet{graves2011practical, zhang2017noisy, ritter2018scalable}. This ensures a p.s.d.~matrix and allows efficient approximation~\citep{martens2014new}.

\textbf{Forward and reverse KL divergence~\citep{murphy2012machine}.} 
Suppose the true distribution is $p(\vtheta)$, and the approximate distribution is $q_\phi(\vtheta)$, the forward and reverse KL divergence are $\kldiv(p(\vtheta)||q_\phi(\vtheta))$ and $\kldiv(q_\phi(\vtheta)||p(\vtheta))$ respectively. 
In general, minimizing the forward KL will arise the mass-covering behavior, and minimizing the reverse KL will arise the zero-forcing/mode-seeking behavior~\citep{minka2005divergence}.
When we use a factorized Gaussian distribution $q_\phi(\vtheta)=\normal(\vtheta | \mathbf{0}, \mSigma)$ to approximate multivariate Gaussian distribution $p(\vtheta)=\normal(\vtheta | \mathbf{0}, \mSigma^*)$, the solutions to minimizing the forward KL and reverse KL are
\begin{align*}
    (a)\ \mSigma &= \mathrm{diag}(\mSigma^*) & (b)\  \mSigma &= \mathrm{diag}(\mLambda^*)^{-1}
\end{align*}
where the precision matrix $\mLambda^* = \mSigma^{*-1}$. 
For the Laplace approximation, the true posterior variance $\mSigma^*$ is $\mH^{-1}$.

\textbf{K-FAC.}
Kronecker-factored approximate curvature (K-FAC)~\citep{martens2015optimizing} uses a Kronecker-factored approximation to the Fisher matrix of fully connected layers, \ie~no weight sharing.
Considering $l$-th layer in a neural network whose input activations are $\ba \in \mathbb{R}^{n}$, weight matrix $\weight \in \mathbb{R}^{n \times m}$, and output $\bs \in \mathbb{R}^{m}$, we have $\bs = \weight^\top \ba$. Therefore, the weight gradient is $\nabla_{\weight}\loss = \ba(\nabla_{\bs} \loss )^\top$. With this formula, K-FAC decomposes this layer's Fisher matrix $\fisher$ with an independence assumption:
\begin{equation}
\label{eq:kfac_for_fc}
\begin{aligned}
	\fisher &= \expect[\newvec \{\nabla_{\weight} \loss\}\newvec\{\nabla_{\weight}\loss \}^\top]\\ &= \expect[\{\nabla_{\bs} \loss \}\{\nabla_{\bs} \loss\}^\top \kron \ba\ba^\top] \\
    &\approx \expect [\{\nabla_{\bs} \loss\}\{\nabla_{\bs} \loss \}^\top] \kron \expect[\ba\ba^\top] = \SMat \kron \AMat,
\end{aligned}
\end{equation}
where $\mA = \expect [\ba\ba^\top]$ and 
$\mS = \expect [\{ \nabla_{\vecs} \loss \} \{\nabla_{\vecs} \loss \}^{\top}]$. 

\citet{grosse2016kronecker} further extended K-FAC to convolutional layers under additional assumptions of spatial homogeneity~(\textbf{SH}) and spatially uncorrelated derivatives~(\textbf{SUD}). Suppose the input $\va \in \mathbb{R}^{c_{\mathrm{in}} \times h \times w}$ and the output $\bs \in \mathbb{R}^{c_{\mathrm{out}} \times h \times w}$, then the gradient of the reshaped weight $\mW \in \mathbb{R}^{c_{\mathrm{out}}\times c_{\mathrm{in}}k^2}$ is $\nabla_{\weight}\loss = \sum\ba_i\nabla_{\bs_{i}}\loss  ^\top$, and the corresponding Fisher matrix is:
\begin{equation}
\begin{aligned}
	\fisher &\approx \sum\expect\left[ \{\nabla_{\bs_{i}}\loss \}\{\nabla_{\bs_{i'}}\loss \}^\top\right] \kron \expect\left[\ba_{i}\ba_{i'}^\top\right] \\
	&\approx \underbrace{\left(\frac{1}{|\mathcal{I}|}\sum\expect\left[ \{\nabla_{\bs_{i}}\loss \}\{\nabla_{\bs_{i}}\loss \}^\top\right]\right)}_{\mS, \mathrm{size}=(c_{\mathrm{out}})^2} \kron \underbrace{\left(\sum\expect\left[\ba_{i}\ba_{i}^\top\right]\right)}_{\mA, \mathrm{size}=(c_{\mathrm{in}} \times k^2)^2} 
\end{aligned}
\end{equation}
where $\mathcal{I} = [h]\times [w]$ is the set of spatial locations, $\va_i \in \mathbb{R}^{c_{\mathrm{in}}k^2}$ is the patch extracted from $\va$, $\nabla_{\bs_{i}}\loss \in \mathbb{R}^{c_{\mathrm{out}}}$ is the gradient to each spatial location in $\bs$ and $i, i' \in \mathcal{I}$. 
Decomposing $\fisher$ into $\mA$ and $\mS$ not only avoids the quadratic storage cost of the exact Fisher, but also enables efficient computation of the Fisher vector product:
\begin{equation}
\begin{aligned}
    \fisher \mathrm{vec} \{\mX\} = \SMat \kron \AMat \mathrm{vec} \{\mX\}
    = \mathrm{vec} \{\mA \mX \mS^\top\}
\end{aligned}
\end{equation}
and fast computation of inverse and eigen-decomposition:
\begin{equation}
\begin{aligned}
&\fisher^{-1} = (\SMat \kron \AMat)^{-1} = \mS^{-1} \kron \mA^{-1} \\
&\fisher = (\mQ_\mS \kron \mQ_\mA) (\mLambda_\mS \kron \mLambda_\mA) (\mQ_\mS \kron \mQ_\mA)^\top
\end{aligned}
\end{equation}
where $\mQ$ and $\mLambda$ are eigenvectors and eigenvalues. Since $\mQ_\mS \kron \mQ_\mA$ gives the eigenbasis of the Kronecker product, we call it the Kronecker-factored Eigenbasis (KFE).

\textbf{Structured Pruning.} 
Structured network pruning~\cite{he2017channel, liu2017learning, li2016pruning, luo2017thinet} is a technique to reduce the size of a network while retaining the original convolutional structure. Among structured pruning methods, channel/filter pruning is the most popular. Let $c_\mathrm{in}$ denote the number of input channels for the $l$-th convolutional layer and $h$/$w$ be the height/width of the input feature maps. The conv layer transforms the input $\va \in \mathbb{R}^{c_\mathrm{in}\times h\times w}$ with $c_\mathrm{out}$ filters $\mathcal{F}_{i}$. All the filters constitute the kernel matrix $\mathcal{F} \in \mathbb{R}^{c_\mathrm{in}\times c_\mathrm{out} \times k\times k}$. When a filter $\mathcal{F}_i$ is pruned, its corresponding feature map in the next layer $\va_i$ is removed, so channel and filter pruning are typically referred to as the same thing. However, most current channel pruning methods either require predefined target models~\citep{li2016pruning, luo2017thinet} or specialized network architectures~\cite{liu2018rethinking}, making them hard to use.

\section{Revisiting OBD and OBS}
\label{sec:revisit}
OBD and OBS share the same basic pruning pipeline: first training a network to (local) minimum in error at weight $\vtheta^*$, and then pruning a weight that leads to the smallest increase in the training error. The predicted increase in the error for a change in full weight vector $\Delta\vtheta$ is: %
\begin{equation}\label{eq:change_of_loss}
    \Delta \loss = \underbrace{\frac{\partial \loss}{\partial \vtheta}^{\top}\Delta \vtheta}_{\approx 0} + \frac{1}{2}\Delta\vtheta^{\top}\mH\Delta\vtheta + \mathcal{O}(||\Delta\vtheta||^3)
\end{equation}
Eqn.~\eqref{eq:change_of_loss} is a simple second order Taylor expansion around the local mode, which is essentially the Laplace approximation.
According to Eqn.~\eqref{eq:laplace_approx}, we can reinterpret the above cost function from a probabilistic perspective:
\begin{equation}
\begin{aligned}
	&\Delta \loss = -\log p_{\mathrm{LA}}(\vtheta^{*}+\Delta \vtheta| \mathcal{D}) + \mathrm{const} \\
	&p_{\mathrm{LA}}(\vtheta^{*}+\Delta \vtheta| \mathcal{D}) = \mathcal{N}(\Delta \vtheta | \mathbf{0},\  \mH^{-1})
\end{aligned}
\end{equation}
where LA denotes Laplace approximation. 

\textbf{OBD.} Due to the intractability of computing full Hessian in deep networks, the Hessian matrix $\mH$ is approximated by a diagonal matrix in OBD. If we prune a weight $\vtheta_q$, then the corresponding change in weights as well as the cost are:
\begin{equation}\label{eq:obd-f1-objective}
    \Delta \vtheta_q = -\vtheta_q^{*} \;\; \mathrm{and} \;\; \Delta \loss_{\text{OBD}} = \frac{1}{2}\left(\vtheta_q^{*}\right)^2 \mH_{qq}
\end{equation}
It regards all the weights as uncorrelated, such that removing one will not affect the others. This treatment can be problematic if the weights are correlated in the posterior.

\textbf{OBS.} In OBS, the importance of each weight is calculated by solving the following constrained optimization problem:
\begin{equation}\label{eq:OBS_objective}
	\min_{q}   \{ \min_{\Delta\vtheta } \  \frac{1}{2}\Delta \vtheta^{\top} \mH \Delta\vtheta  \;\; s.t. \;\; \ve_{q}^{\top}\Delta\vtheta + \vtheta_{q}^{*} = 0 \}
\end{equation}
for considering the correlations among weights, where $\ve_q$ is the unit selecting vector whose $q$-th element is 1 and 0 otherwise. Solving Eqn.~\eqref{eq:OBS_objective} yields the optimal weight change and the corresponding change in error:
\begin{equation} \label{eq:obs-f-1}
	\Delta\vtheta = -\frac{\vtheta_{q}^{*}}{[\mH^{-1}]_{qq}}\mH^{-1}\ve_q \;\; \mathrm{and} \;\; \Delta \loss_{\text{OBS}} = \frac{1}{2}\frac{(\vtheta_{q}^{*})^{2}}{[\mH^{-1}]_{qq}}
\end{equation}
The main difference is that OBS not only prunes a single weight but takes into account the correlation between weights and updates the rest of the weights to compensate.

\subsection{New Insights and Perspectives}
A common belief is that OBS is superior to OBD, though it is only feasible for shallow networks. In the following paragraphs, we will show that this may not be the case in practice even when we can compute exact Hessian inverse.

From Eqn.~\eqref{eq:obd-f1-objective}, we can see that OBD can be seen as OBS with off-diagonal entries of the Hessian ignored. If we prune only one weight each time, OBS is advantageous in the sense that it takes into account the off-diagonal entries. 
However, pruning weights one by one is time consuming and typically infeasible for modern neural networks. It is more common to prune many weights at a time~\citep{zeng2019mlprune,dong2017learning,han2015learning}, especially in structured pruning~\citep{liu2017learning, luo2017thinet, li2016pruning}.  

We note that, when pruning multiple weights simultaneously, both OBD and OBS can be interpreted as using a factorized Gaussian $\mathcal{N}(\Delta \vtheta | \mathbf{0},\  \mSigma)$ to approximate the true posterior over weights, but with different objectives. Specifically, OBD can be obtained by minimizing the reverse KL divergence~($\mSigma=\mathrm{diag}(\mH)^{-1}$), whereas OBS is using the forward KL divergence~($\mSigma=\mathrm{diag}(\mH^{-1})$). 
Reverse KL underestimates the variance of the true distribution and overestimates the importance of each weight. By contrast, forward KL overestimates the variance and prunes more aggressively. 
The following example illustrates that while OBS outperforms OBD when pruning only a single weight, there is no guarantee that OBS is better than OBD when pruning multiple weights simultaneously since OBS may prune highly correlated weights all together.
\begin{center}
\begin{minipage}{.95\columnwidth}
\vspace{-5pt}
\begin{examp}
\label{examp:limitation-obd-obs}
{\small Suppose a neural network converged to a local minima with weight $\vtheta^*=[1, 1, 1]^\top$, and the associated Hessian $\mH=\begin{psmallmatrix} 1 & 0.99 & 0 \\ 0.99 & 1 & 0.01 \\ 0 & 0.01 & 0.5 \end{psmallmatrix}$. Compute the resulting weight and increase in loss of OBD and OBS for the following cases.

\vspace{5pt}
\textbf{Case 1: Prune one weight~(OBS is better).}
\vspace{-2pt}
\begin{itemize}[leftmargin=1.2em] \itemsep -1pt 
    \item OBD: $\Delta \vtheta = [0, 0, -1]^\top$, $\Delta \loss= \Delta \loss_\text{OBD}=0.25$
    \item OBS: $\Delta \vtheta = [-1, 0.99, 0.02]^\top$, $\Delta \loss=\Delta \loss_{\text{OBS}}=0.01$
\end{itemize}
\textbf{Case 2: Prune two weights simultaneously~(OBD is better).}
\vspace{-2pt}
\begin{itemize}[leftmargin=1.2em] \itemsep -1pt
    \item \small OBD: $\Delta \vtheta = [0, -1, -1]^\top$, $\Delta \loss=0.76 (\Delta \loss_\text{OBD}=0.75)$
    \item OBS: $\Delta \vtheta = [-1, -1, 0]^\top$, $\Delta \loss=1.99 (\Delta \loss_\text{OBS}=0.02)$
\end{itemize}
}
\vspace{-10pt}
\end{examp}
\vspace{-5pt}
\end{minipage}
\end{center}

OBD and OBS are equivalent when the true posterior distribution is fully factorized. 
It has been observed that different weights are highly coupled~\citep{zhang2017noisy} and diagonal approximation is too crude. However, the correlations are small in the KFE (see Figure~\ref{fig:fisher_matrix}). %
This motivates us to consider applying OBD in the KFE, where the diagonal approximation is more reasonable.

\section{Methods}
\subsection{Approximating the Hessian with the Fisher Matrix}
We use the Fisher matrix to approximate the Hessian. In the following, we briefly discuss the relationship between these matrices. For more detailed discussion, we refer readers to~\citet{martens2014new, Pascanu2013RevisitingNG}.

Suppose the function $\vz=f(\vx, \vtheta)$ is parameterized by $\vtheta$, and the loss function is $\ell (y, \vz)=-\log p(y|\vz)$. Then the Hessian $\mH$ at (local) minimum is equivalent to the generalized Gauss-Newton matrix $\mG$:
\begin{equation}\label{eq:Hessian}
\begin{aligned}
    \mH &= \expect \Big[ \mJ_f^{\top}\mH_{\ell}\mJ_f +
    \underbrace{\sum_{j=1}^m[\nabla_{\vz}\ell (y, \vz)|_{\vz=f(\vx, \vtheta)}]_j\mH_{[f]_j}}_{\approx 0} \Big]\\
    &= \expect \Big[\mJ_f^\top \mH_\mathcal{\ell} \mJ_f \Big] = \mG
\end{aligned}
\end{equation}
where $\nabla_{\vz}\mathcal{\ell}(y, \vz)|_{\vz=f(\vx, \vtheta)}$ is the gradient of $\ell(y, \vz)$ evaluated at $z=f(\vx, \vtheta)$, $\mH_\ell$ is the Hessian of $\ell(y, \vz)$ w.r.t. $\vz$, and $\mH_{[f]_j}$ is the Hessian of $j$-th component of $f(\vx, \vtheta)$.

\citet{pascanu2013revisiting} showed that the Fisher matrix $\fisher$ and generalized Gauss-Newton matrix are identical when the model predictive distribution is in the exponential family, such as categorical distribution (for classification) and Gaussian distribution (for regression), justifying the use of the Fisher to approximate the Hessian.

\subsection{Extending OBD and OBS to Structured Pruning}
OBD and OBS were originally used for weight-level pruning. Before introducing our main contributions, we first extend OBD and OBS to structured (channel/filter-level) pruning. The most na\"ive approach is to first compute the importance of every weight, \ie, Eqn.~\eqref{eq:obd-f1-objective} for OBD and Eqn.~\eqref{eq:obs-f-1} for OBS, then sum together the importances within each filter. We use this approach as a baseline, and denote it C-OBD and C-OBS. For C-OBS, because inverting the Hessian/Fisher matrix is computationally intractable, we adopt the K-FAC approximation for efficient inversion, as first proposed by~\citet{zeng2019mlprune} for weight-level pruning.

In the scenario of structured pruning, a more sophisticated approach is to take into account the correlation of the weights within the same filter. For example, we can compute the importance of each filter as follows: 
\begin{equation}\label{eq:obd-f-1}
    \Delta\loss_i = \frac{1}{2} {\vtheta_i^*}^\top \fisher(i) \vtheta_i^*
\end{equation}
where $\vtheta_i^*\in \mathbb{R}^{c_\mathrm{in}k^2}$ and $\fisher(i) \in \mathbb{R}^{c_\mathrm{in}k^2 \times c_\mathrm{in}k^2}$ are the parameters vector and Fisher matrix of $i$-th filter $\mathcal{F}_i$, respectively.
To do this, we would need to store the Fisher matrix for each filter, which is intractable for large convolutional layers. To overcome this problem, we adopt the K-FAC approximation $\mF = \mS \kron \mA$,
and compute the change in weights as well as the importance in the following way:
\begin{equation}
\label{eq:obd-f-2}
    \Delta\vtheta_i = -\vtheta_i^* \;\; \mathrm{and} \;\; \Delta\loss_i = \frac{1}{2} \mS_{ii} {\vtheta_i^*}^\top \mA \vtheta_i^*
\end{equation}
Unlike Eqn.~\eqref{eq:obd-f-1}, the input factor $\mA$ is shared between different filters, and therefore cheap to store. By analogy, we can compute the change in weights and importance of each filter for Kron-OBS as:
\begin{equation}\label{eq:obs-f-2}
    \Delta\vtheta = - \frac{\mS^{-1} \ve_i \kron\vtheta_{i}^*}{[\mS^{-1}]_{ii}}
    \;\; \mathrm{and} \;\; \Delta\loss_i = \frac{1}{2} \frac{{\vtheta_i^*}^\top \mA \vtheta_i^*}{[\mS^{-1}]_{ii}},
\end{equation}
where $\ve_i$ is the selecting vector with 1 for elements of $\mathcal{F}_i$ and 0 elsewhere.
We refer to Eqn.~\eqref{eq:obd-f-2} and Eqn.~\eqref{eq:obs-f-2} as Kron-OBD and Kron-OBS~(See Algorithm~\ref{alg:nnpruning_alg_obs_obd}). See Appendix~\ref{app:derivation} for derivations.
\begin{algorithm}[t]
\caption{Structured pruning algorithms {\color{blue} Kron-OBD} and {\color{red} Kron-OBS}. For simplicity, we focus on a single layer. $\vtheta_i$ below denotes the parameters of filter $\mathcal{F}_i$, which is a vector.}

\label{alg:nnpruning_alg_obs_obd}
\begin{algorithmic}[1]
\REQUIRE pruning ratio $p$ and training data $\mathcal{D}$
\REQUIRE model parameters (pretrained) $\vtheta=\mathrm{vec}\left(\mW\right)$
\STATE Compute Kronecker factors $\mA = \expect [\va \va^{\top}]$ and $\mS = \expect[\{\nabla_{\bs} \loss\}\{\nabla_{\bs} \loss \}^{\top}]$

\FORALL{filter $i$}
    \STATE {\color{blue} $\Delta \loss_{i} =  \frac{1}{2} \mS_{ii} \vtheta_{i}^\top \mA \vtheta_{i}$} or {\color{red} $\Delta \loss_{i} = \frac{1}{2} \frac{\vtheta_{i}^\top \mA \vtheta_{i}}{\left[\mS^{-1}\right]_{ii}}$}
\ENDFOR 
\STATE Compute $p_\mathrm{th}$ percentile of $\Delta \loss$ as $\tau$
\FORALL{filter $i$}
    \IF{$\Delta \loss_i \leq \tau$}
        \STATE {\color{blue}$\vtheta_{i} \leftarrow \mathbf{0}$} or
        {\color{red}$\vtheta \leftarrow \vtheta - \frac{\mS^{-1} \ve_i \kron\vtheta_{i}^*}{[\mS^{-1}]_{ii}}$}%
    \ENDIF
\ENDFOR
\STATE Finetune the network on $\mathcal{D}$ until converge
\end{algorithmic}
\end{algorithm}
\subsection{EigenDamage: Structured Pruning in a KFE}

As argued in Section~\ref{sec:revisit}, weight-level OBD and OBS approximate the posterior distribution with a factorized Gaussian around the mode, which is overly restrictive and cannot capture the correlation between weights. Although we just extended them to filter/channel pruning, which captures correlations of weights within the same filter, the interactions between filters are ignored. In this section, we propose to decorrelate the weights before pruning. In particular, we introduce a novel network reparameterization by breaking each linear operation into three stages. Intuitively, the role of the first and third stages is to rotate to the KFE. %

Considering a single layer with weight $\mW$ with K-FAC Fisher $\SMat \kron \AMat$ (see Section~\ref{subsec:background}), we can decompose the weight matrix $\mW$ as the following form:
\begin{equation}\label{eq:decomp}
    \mathrm{vec}\{\mW \} = (\mQ_\mS \kron \mQ_\mA) \mathrm{vec}\{\mW \} = \mathrm{vec}\{\mQ_{\mA} \mW^\prime \mQ_{\mS}^\top \}
\end{equation}
where $\mathrm{vec}\{\mW' \} = (\mQ_\mS \kron \mQ_\mA)^\top \mathrm{vec}\{\mW \}$.
It is easy to show that the Fisher matrix for $\mW^\prime$ is diagonal if the assumptions of K-FAC are satisfied~\cite{george2018fast}.
We then apply C-OBD (or equivalently C-OBS since the Fisher is close to diagonal) on $\mW^\prime$ for \emph{both} input and output channels. This way, each layer has a bottleneck structure which is a low-rank approximation, which we term \emph{eigenpruning}. (Note that C-OBD and Kron-OBD only prune the output channels, since it automatically results in removal of corresponding input channel in the next layer.)
We refer to our proposed method as EigenDamage~(See Algorithm~\ref{alg:nnpruning_alg_eigen_pruning}).

EigenDamage preserves the input and output shape, and thus can be applied to any convolutional or fully connected architecture without modification, in contrast with~\citet{liu2017learning}, which requires adaptions for networks with cross-layer connections. Furthermore, like all Hessian-based pruning methods, our criterion allows us to set one global compression ratio for the whole network, making it easy to use. Moreover, the introduced eigen-basis $\mQ_\mA$ can be further compressed by the "doubly factored" Kronecker approximation~\citep{ba2016distributed}, and $\mW'$ can be also compressed by depth-wise separable decomposition, as detailed in Sections~\ref{sec:reduce_eigenbasis} and \ref{sec:depthwise_separable}.

\begin{algorithm}[t]
\caption{Pruning in the Kronecker-factored eigenbasis, \ie, {\color{green}EigenDamage}. For simplicity, we focus on a single layer. $\odot$ denotes elementwise mutliplication.}
\label{alg:nnpruning_alg_eigen_pruning}
\begin{algorithmic}[1]
\REQUIRE pruning ratio $p$ and training data $\mathcal{D}$
\REQUIRE model parameters (pretrained) $\vtheta = \mathrm{vec}\left(\mW \right)$
\STATE Compute Kronecker factors $\mA = \expect [\va \va^{\top}]$ and $\mS = \expect[\{\nabla_{\bs} \loss\}\{\nabla_{\bs} \loss \}^{\top}]$
\STATE $\mQ_\mS, \mathbf{\Lambda}_\mS= \mathrm{Eigen}\left(\mS\right)$ and $\mQ_\mA, \mathbf{\Lambda}_\mA= \mathrm{Eigen}\left(\mA\right)$
\STATE Decompose weight $\mW$ according by Eqn.~\eqref{eq:decomp}
\STATE $\mTheta = \mW^\prime \odot \mathrm{diag}(\mathbf{\Lambda}_\mA) \mathrm{diag}(\mathbf{\Lambda}_\mS)^\top \odot \mW^\prime$
\FORALL{row $r$ or column $c$ in $\mTheta$}
    \STATE $\Delta \loss_{r} = \mTheta_{r, \cdot}\mathbf{1}$ and $\Delta \loss_{c} = \mathbf{1}^\top \mTheta_{\cdot, c}$
\ENDFOR
\STATE Compute $p_\mathrm{th}$ percentile of $\Delta \loss$ as $\tau$
\STATE Remove $r_{\mathrm{th}}$ row (or $c_\mathrm{th}$ column) in $\mW^\prime$ and $r_\mathrm{th}$ (or $c_\mathrm{th}$) eigenbasis in $\mQ_\mA$ (or $\mQ_\mS$) if $\Delta \loss_r$ (or $\Delta \loss_c$) $\leq \tau$
\STATE Finetune the network on $\mathcal{D}$ until convergence
\end{algorithmic}

\end{algorithm}
\subsection{Iterative Pruning}
The above method relies heavily on the Taylor expansion~\eqref{eq:change_of_loss}, which may be accurate if we prune only a few filters. Unfortunately, the approximation will break down if we prune a large number of filters. 
In order to handle this issue, we can conduct the pruning process iteratively and only prune a few filters each iteration. Specifically, once we finish pruning the network for the first time, each layer has a bottleneck structure (\ie, $\mQ_\mA, \mW', \mathrm{and}\; \mQ_\mS$). We can then conduct the next pruning iteration (after finetuning) on $\mW'$ in the same manner. This will result in two new eigenbases associated with $\mW^\prime$. Conveniently, we can always merge these two new eigenbases (\ie, $\mQ'_\mA, \mQ'_\mS$) into old ones so as to reduce the model size as well as FLOPs by:
\begin{equation}
	\mQ_\mA \leftarrow \mQ_\mA\mQ_\mA' \;\; \mathrm{and} \;\; \mQ_\mS \leftarrow \mQ_\mS\mQ_\mS'
\end{equation}
This procedure may take several iterations until it reaches desirable compression ratio.

\subsection{Reducing the Parameter Count of the Eigenbasis}
\label{sec:reduce_eigenbasis}
\begin{wrapfigure}[5]{L}{0.21\columnwidth}
    \centering
    \vspace{-0.4cm}
    \includegraphics[width=0.2\columnwidth]{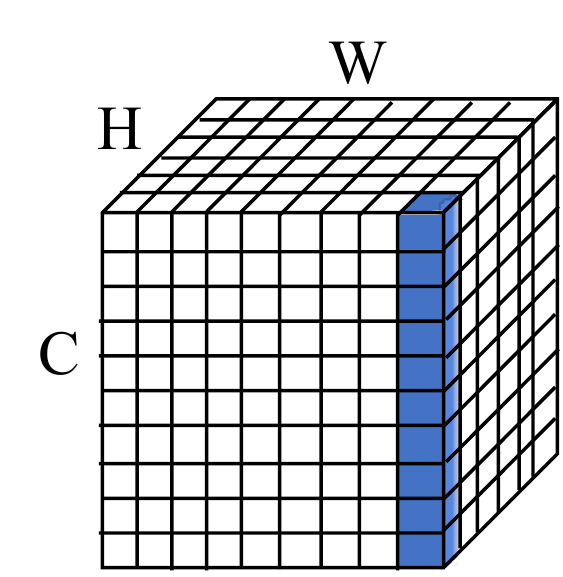}
    \label{fig:a_hat}
\end{wrapfigure}
Since the eigenbasis $\mQ_{\mA}$ can take up a large chunk of memory for convolutional networks\footnote{$\mQ_\mA$ has the shape of $c_\mathrm{in}k^2 \times c_\mathrm{in} k^2$.}, we further leverage the internal structure to reduce the model size. Inspired by~\citet{ba2016distributed}'s ``doubly factored'' Kronecker approximation for layers whose input feature maps are too large, 
we ignore the correlation among the spatial locations within the same input channel. In that case, $\mA \in \mathbb{R}^{c_\mathrm{in} \times c_\mathrm{in}}$ only captures the correlation between different channels. Here we abuse the notation slightly and let $\mA$ denote the covariance matrix along the channel dimension and $\va \in \mathbb{R}^{c_\mathrm{in}}$ (see blue cubes) the activation of each spatial location:
\begin{equation}\label{eq:channel-cov}
    \mA = \expect \left[\va \va^\top  \right] = \frac{1}{N |\mathcal{T}|}\sum_{\vx}\sum_{t \in \mathcal{T}} \va_t(\vx)\va_t(\vx)^\top
\end{equation}
The expectation in Eqn.~\eqref{eq:channel-cov} is taken over training examples $\vx$ and spatial locations $\mathcal{T}$.
We note that with such approximation, $\mQ_\mA$ can be efficiently implemented by $1\times1$ conv, resulting in compact bottleneck structures like ResNet~\citep{he2016deep}, as shown in Figure~\ref{fig:figure_1}. This will greatly reduce the size of eigen-basis to be $1/k^4$ of the original one.

\subsection{Depthwise Separable Decomposition}
\label{sec:depthwise_separable}
\begin{figure}[t]
     \centering
      \includegraphics[width=1.0\columnwidth]{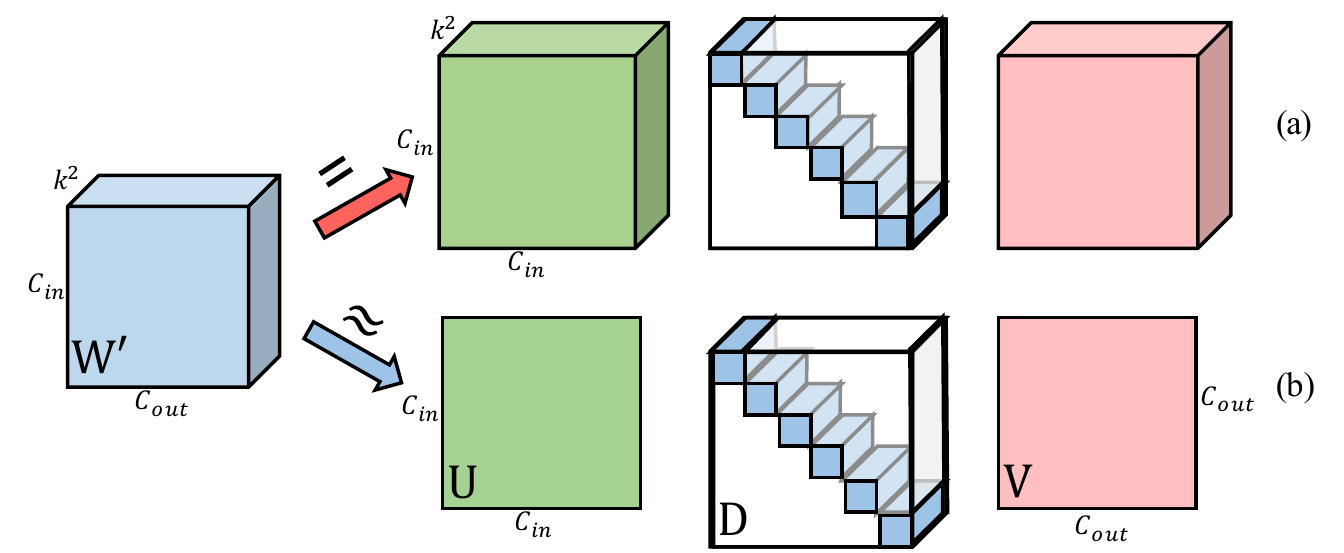}
     \vspace{-0.4cm}
        \caption{Two schemes for depthwise separable decomposition of the convolution layer. The parameters in the blank region are zeros.}
        \label{fig:depthwise_separable}
        \vspace{-0.5cm}
\end{figure}

Depthwise separable convolution has been proven to be effective in designing lightweight models~\citep{howard2017mobilenets, chollet2017xception, Zhang_2018_CVPR, ma2018shufflenet}. 
The idea of separable convolution can be naturally incorporated in our method to further reduce the computational cost and model size. 
For convolution filters $\mW' \in \mathbb{R}^{c_{\mathrm{in}}\times c_{\mathrm{out}}\times k\times k}$, we perform the singular value decomposition (SVD) for every slice $\mW_{:, :, i}' \in \mathbb{R}^{c_\mathrm{in}\times c_\mathrm{out}}$; then we can get a diagonal matrix as well as two new bases, as shown in Figure~\ref{fig:depthwise_separable}~(a). However, such a decomposition will result in more than twice the original parameters due to the two new bases. Therefore, we again ignore the correlation along the spatial dimension of filters,~\ie~sharing the basis for each spatial dimension~(see Figure~\ref{fig:depthwise_separable}~(b)). 
In particular, we solve the following problem:
\begin{equation}\label{eq:frob_norm}
	\min_{\mU, \{\mD_i\}_{i=1}^{k^2}, \mV}\  \frac{1}{2} \sum_{i=1}^{k^2} ||\mU\mD_i\mV^{\top}-\mW_{:,:,i}'||_{\mathrm{Frob}}^2
\end{equation}
where $\mU \in \mathbb{R}^{c_\mathrm{in}\times c_\mathrm{in}}$, $\mD_i \in \mathbb{D}^{c_\mathrm{in} \times c_\mathrm{out}}$\footnote{$\mathbb{D}^{c_\mathrm{in} \times c_\mathrm{out}}$ is the domain of diagonal matrices.} and $\mV \in \mathbb{R}^{c_{\mathrm{out}}\times c_\mathrm{out}}$. We can merge $\mU$ and $\mV$ into $\mQ_\mathrm{A}$ and $\mQ_\mathrm{S}$ respectively, and then replace $\mW'$ with $\mD$, which can be implemented with a depthwise convolution. By doing so, we are able to further reduce the size of the filter to be $1/c_{\mathrm{in}}$ of the original one.

\begin{table*}[t]
\small

\caption{ One-pass pruning on CIFAR10 and CIFAR100 with VGG19, ResNet32 and PreResNet29. To be noted, we cannot control the pruned ratio of parameters since we prune the whole filter and different filters are not of the same size. We run each experiment five times, and present the mean and standard variance.}%
\vspace{-0.4cm}
\label{tab:one_pass_vgg_resnet}
\begin{center}
\resizebox{\textwidth}{!}{
\begin{tabular}{l|c c c|c c c||c c c|c c c}
\toprule
{\normalsize Dataset}
& \multicolumn{6}{c||}{{\normalsize \textbf{CIFAR10}}}
& \multicolumn{6}{c}{{\normalsize \textbf{CIFAR100}}}\\
\hline
Prune Ratio (\%)  
& \multicolumn{3}{c|}{60\%} 
& \multicolumn{3}{c||}{90\%}
& \multicolumn{3}{c|}{60\%}
& \multicolumn{3}{c}{90\%} \\
\hline
\multirow{2}{*}{Method }
& Test & Reduction in & Reduction in & Test & Reduction in & Reduction in & Test & Reduction in & Reduction in & Test & Reduction in & Reduction in \\
& acc (\%) & weights (\%) & FLOPs (\%)       
& acc (\%) & weights (\%) & FLOPs (\%)  
& acc (\%) & weights (\%) & FLOPs (\%)  
& acc (\%) & weights (\%) & FLOPs (\%)   
\\
\hline
\textbf{VGG19(Baseline)}
& 94.17 & - & -
& -     & - & -
& 73.34 & - & -
& -     & - & - \\
NN Slimming~\citep{liu2017learning} 
& 92.84 $\pm$ - & 80.07 $\pm$ - & 42.65 $\pm$ -
& 85.01 $\pm$ - & 97.85 $\pm$ - & 97.89 $\pm$ -
& 71.89 $\pm$ - & 74.60 $\pm$ - & 38.33 $\pm$ -
& 58.69 $\pm$ - & 97.76 $\pm$ - & 94.09 $\pm$ - \\

C-OBD%
& 94.04 $\pm$ 0.12 & 82.01 $\pm$ 0.44 & 38.18 $\pm$ 0.45
& 92.34 $\pm$ 0.18 & 97.68 $\pm$ 0.02 & 77.39 $\pm$ 0.36
& 72.23 $\pm$ 0.15 & 77.03 $\pm$ 0.05 & 33.70 $\pm$ 0.04
& 58.07 $\pm$ 0.60 & 97.97 $\pm$ 0.04 & 77.55 $\pm$ 0.25 \\

C-OBS%
& 94.08 $\pm$ 0.07 & 76.96 $\pm$ 0.14 & 34.73 $\pm$ 0.11
& 91.92 $\pm$ 0.16 & 97.27 $\pm$ 0.04 & 87.53 $\pm$ 0.41
& 72.27 $\pm$ 0.13 & 73.83 $\pm$ 0.03 & 38.09 $\pm$ 0.06
& 58.87 $\pm$ 1.34 & 97.61 $\pm$ 0.01 & 91.94 $\pm$ 0.26\\

Kron-OBD 
& 94.00 $\pm$ 0.11 & 80.40 $\pm$ 0.26 & 38.19 $\pm$ 0.55
& \textbf{92.92 $\pm$ 0.26} & 97.47 $\pm$ 0.02 & 81.44 $\pm$ 0.68
& 72.29 $\pm$ 0.11 & 77.24 $\pm$ 0.10 & 37.90 $\pm$ 0.24
& 60.70 $\pm$ 0.51 & 97.56 $\pm$ 0.08 & 82.55 $\pm$ 0.39 \\

Kron-OBS 
& \textbf{94.09 $\pm$ 0.12} & 79.71 $\pm$ 0.26 & 36.93 $\pm$ 0.15
& 92.56 $\pm$ 0.21 & 97.32 $\pm$ 0.02 & 80.39 $\pm$ 0.21
& 72.12 $\pm$ 0.14 & 74.18 $\pm$ 0.04 & 36.59 $\pm$ 0.11
& 60.66 $\pm$ 0.35 & 97.48 $\pm$ 0.03 & 83.57 $\pm$ 0.27 \\

EigenDamage 
& 93.98 $\pm$ 0.06 & 78.18 $\pm$ 0.12 & 37.13 $\pm$ 0.41
& 92.29 $\pm$ 0.21 & 97.15 $\pm$ 0.04 & 86.51 $\pm$ 0.26
& \textbf{72.90 $\pm$ 0.06} & 76.64 $\pm$ 0.12 & 37.40 $\pm$ 0.11
& \textbf{65.18 $\pm$ 0.10} & 97.31 $\pm$ 0.01 & 88.63 $\pm$ 0.12 \\

\hline
\textbf{VGG19+$L_1$ (Baseline)}
& 93.71 & - & -
& -     & - & -
& 73.08     & - & -
& -     & - & - \\

NN Slimming~\citep{liu2017learning} 
& 93.79 $\pm$ - &  83.45 $\pm$ - & 49.23 $\pm$ -
& \textbf{91.99 $\pm$ -} &  97.93 $\pm$ - & 86.00 $\pm$ -
& 72.78 $\pm$ - & 76.53 $\pm$ - & 39.92 $\pm$ -
& 57.07 $\pm$ - & 97.59 $\pm$ - & 93.86$\pm$ - \\

C-OBD%
& 93.84 $\pm$ 0.04 & 84.19 $\pm$ 0.01 & 47.34 $\pm$ 0.02
& 91.29 $\pm$ 0.30 & 97.88 $\pm$ 0.02 & 81.22 $\pm$ 0.38
& 72.73 $\pm$ 0.09 & 79.47 $\pm$ 0.02 & 39.04 $\pm$ 0.02
& 56.49 $\pm$ 0.06 & 97.96 $\pm$ 0.03 & 80.91 $\pm$ 0.16 \\

C-OBS%
& 93.85 $\pm$ 0.01 & 82.88 $\pm$ 0.02 & 44.58 $\pm$ 0.10
& 91.14 $\pm$ 0.13 & 97.31 $\pm$ 0.03 & 88.18 $\pm$ 0.27
& 72.58 $\pm$ 0.09 & 76.17 $\pm$ 0.01 & 41.61 $\pm$ 0.06
& 44.18 $\pm$ 0.87 & 97.31 $\pm$ 0.02 & 91.90 $\pm$ 0.07 \\ 

Kron-OBD 
& 93.86 $\pm$ 0.06 & 84.78 $\pm$ 0.00 & 50.10 $\pm$ 0.00
& 91.14 $\pm$ 0.26 & 97.74 $\pm$ 0.02 & 83.09 $\pm$ 0.33
& 72.44 $\pm$ 0.03 & 79.99 $\pm$ 0.02 & 43.46 $\pm$ 0.02
& 57.59 $\pm$ 0.21 & 97.53 $\pm$ 0.02 & 85.04 $\pm$ 0.07 \\

Kron-OBS 
& 93.84 $\pm$ 0.04 & 84.33 $\pm$ 0.03 & 48.01 $\pm$ 0.13
& 91.13 $\pm$ 0.17 & 97.37 $\pm$ 0.01 & 81.52 $\pm$ 0.18
& 72.61 $\pm$ 0.15 & 77.27 $\pm$ 0.03 & 40.89 $\pm$ 0.59
& 57.61 $\pm$ 0.67 & 97.51 $\pm$ 0.02 & 86.60 $\pm$ 0.14 \\

EigenDamage 
& \textbf{93.88 $\pm$ 0.04} & 79.50 $\pm$ 0.02 & 39.84 $\pm$ 0.11
& 91.79 $\pm$ 0.16 & 96.84 $\pm$ 0.02 & 84.82 $\pm$ 0.21
& \textbf{73.01 $\pm$ 0.05} & 75.41 $\pm$ 0.03 & 37.46 $\pm$ 0.06
& \textbf{64.91 $\pm$ 0.23} & 97.28 $\pm$ 0.04 & 88.65 $\pm$ 0.06 \\
\bottomrule
\bottomrule

\textbf{ResNet32(Baseline)}
& 95.30 & - & -
& -     & - & -
& 78.17 & - & -
& -     & - & -\\
NN Slimming ~\cite{liu2017learning} 
& N/A & N/A & N/A
& N/A & N/A & N/A
& N/A & N/A & N/A
& N/A & N/A & N/A
\\

C-OBD%

& 95.11 $\pm$ 0.10 & 70.36 $\pm$ 0.39 & 66.18 $\pm$ 0.46
& 91.75 $\pm$ 0.42 & 97.30 $\pm$ 0.06 & 93.50 $\pm$ 0.37 
& 75.70 $\pm$ 0.31 & 66.68 $\pm$ 0.25 & 67.53 $\pm$ 0.25
& 59.52 $\pm$ 0.24 & 97.74 $\pm$ 0.08 & 94.88 $\pm$ 0.08\\

C-OBS%

& 95.04 $\pm$ 0.07 & 67.90 $\pm$ 0.25 & 76.75 $\pm$ 0.36
& 90.04 $\pm$ 0.21 & 95.49 $\pm$ 0.22 & 97.39 $\pm$ 0.04 
& 75.16 $\pm$ 0.32 & 66.83 $\pm$ 0.03 & 76.59 $\pm$ 0.34
& 58.20 $\pm$ 0.56 & 91.99 $\pm$ 0.07 & 96.27 $\pm$ 0.02\\

Kron-OBD 

& 95.11 $\pm$ 0.09 & 63.97 $\pm$ 0.22 & 63.41 $\pm$ 0.42
& 92.57 $\pm$ 0.09 & 96.11 $\pm$ 0.12 & 94.18 $\pm$ 0.17 
& 75.86 $\pm$ 0.37 & 63.92 $\pm$ 0.23 & 62.97 $\pm$ 0.17
& 62.42 $\pm$ 0.41 & 96.42 $\pm$ 0.05 & 95.85 $\pm$ 0.08\\

Kron-OBS 

& 95.14 $\pm$ 0.07 & 64.21 $\pm$ 0.31 & 61.89 $\pm$ 0.79
& 92.76 $\pm$ 0.12 & 96.14 $\pm$ 0.27 & 94.37 $\pm$ 0.54 
& \textbf{75.98 $\pm$ 0.33} & 62.36 $\pm$ 0.40 & 60.41 $\pm$ 1.02
& 63.62 $\pm$ 0.50 & 93.56 $\pm$ 0.14 & 95.65 $\pm$ 0.13\\

EigenDamage 

& \textbf{95.17 $\pm$ 0.12} & 71.99 $\pm$ 0.13 & 70.25 $\pm$ 0.24
& \textbf{93.05 $\pm$ 0.23} & 96.05 $\pm$ 0.03 & 94.74 $\pm$ 0.02
& 75.51 $\pm$ 0.11 & 69.80 $\pm$ 0.11 & 71.62 $\pm$ 0.21
& \textbf{65.72 $\pm$ 0.04} & 95.21 $\pm$ 0.04 & 94.62 $\pm$ 0.06\\

\hline
\textbf{PreResNet29+$L_1$ (Baseline)}
& 94.42 & -  & -
& -     & -  & -
& 75.70 & -  & -
& -     & -  & -\\

NN Slimming~\citep{liu2017learning} 

& 92.32 $\pm$ - & 71.60  $\pm$ - & 80.95 $\pm$ -
& 82.50 $\pm$ - & 93.49  $\pm$ - & 95.88 $\pm$ -
& 68.87 $\pm$ - & 61.68 $\pm$ - & 82.03 $\pm$ -
& 49.48 $\pm$ - & 93.70 $\pm$ - & 96.33 $\pm$ -\\

C-OBD%

& 91.17 $\pm$ 0.16 & 87.48 $\pm$ 0.23 & 78.14 $\pm$ 0.70
& 80.03 $\pm$ 0.21 & 98.45 $\pm$ 0.02 & 96.03 $\pm$ 0.10 
& 62.19 $\pm$ 0.18 & 89.72 $\pm$ 0.01 & 82.24 $\pm$ 0.16
& 36.44 $\pm$ 0.90 & 98.65 $\pm$ 0.00 & 96.81 $\pm$ 0.02\\

C-OBS%

& 91.64 $\pm$ 0.22 & 83.52 $\pm$ 0.12 & 76.33 $\pm$ 0.21
& 76.59 $\pm$ 0.69 & 98.34 $\pm$ 0.02 & 98.47 $\pm$ 0.02 
& 68.10 $\pm$ 0.29 & 81.26 $\pm$ 0.10 & 89.47 $\pm$ 0.04
& 32.77 $\pm$ 0.89 & 97.89 $\pm$ 0.01 & 98.73 $\pm$ 0.00\\

Kron-OBD 

& 90.22 $\pm$ 0.43 & 74.84 $\pm$ 0.20 & 67.83 $\pm$ 0.33
& 82.68 $\pm$ 0.20 & 98.18 $\pm$ 0.04 & 94.90 $\pm$ 0.13 
& 57.76 $\pm$ 0.28 & 76.85 $\pm$ 0.06 & 72.38 $\pm$ 0.02
& 34.26 $\pm$ 1.12 & 98.62 $\pm$ 0.00 & 96.09 $\pm$ 0.00\\

Kron-OBS 

& 89.02 $\pm$ 0.17 & 72.96 $\pm$ 0.20 & 70.14 $\pm$ 0.18
& 81.77 $\pm$ 0.59 & 98.44 $\pm$ 0.01 & 96.85 $\pm$ 0.09 
& 60.28 $\pm$ 0.37 & 70.53 $\pm$ 0.11 & 76.60 $\pm$ 0.14
& 33.45 $\pm$ 0.96 & 98.31 $\pm$ 0.00 & 97.15 $\pm$ 0.01\\

EigenDamage 

& \textbf{93.80 $\pm$ 0.05} & 70.09 $\pm$ 0.12 & 63.13 $\pm$ 0.26
& \textbf{89.10 $\pm$ 0.13} & 93.45 $\pm$ 0.04 & 90.67 $\pm$ 0.06 
& \textbf{73.62 $\pm$ 0.16} & 66.73 $\pm$ 0.17 & 62.86 $\pm$ 0.12
& \textbf{65.11 $\pm$ 0.15} & 92.33 $\pm$ 0.02 & 90.52 $\pm$ 0.02\\

\bottomrule

\end{tabular}
}
\end{center}
\vspace{-0.3cm}
\end{table*}
\section{Experiments}

In this section, we aim to verify the effectiveness of EigenDamage in reducing the test-time resource requirements of a network without significantly sacrificing accuracy. We compare EigenDamage with other compression methods in terms of test accuracy, reduction in weights, reduction in FLOPs, and inference wall-clock time speedup. Wherever possible, we analyze the tradeoff curves involving test accuracy and resource requirements. We find that EigenDamage gives a significantly more favorable tradeoff curve, especially on larger architectures and more difficult datasets.

\subsection{Experimental Setup}
We test our methods on two network architectures: VGGNet~\citep{Simonyan2014VeryDC} and (Pre)ResNet\footnote{For ResNet, we widen the network by a factor of 4, as done in~\citet{zhang2019weightdecay}}~\citep{he2016identity, he2016deep}. We make use of three standard benchmark datasets: CIFAR10, CIFAR100~\citep{krizhevsky2009learning} and Tiny-ImageNet\footnote{https://tiny-imagenet.herokuapp.com}. 
We compare EigenDamage to the extended versions C-OBD/OBS and Kron-OBD/OBS as well as one state-of-the-art channel-level pruning algorithm, NN Slimming~\citep{liu2017learning, liu2018rethinking}, and a low-rank approximation algorithm, CP-Decomposition~\citep{jaderberg2014speeding}.
Note that because NN Slimming requires imposing $L_{1}$ loss on the scaling weights of BatchNorm~\citep{ioffe2015batch}, we train the networks with two different settings, \ie, with and without $L_1$ loss, for fair comparison.

For networks with skip connections, NN Slimming can only be applied to specially designed network architectures. Therefore, in addition to ResNet32,
we also test on PreResNet-29~\citep{he2016identity}, which is in the same family of architectures considered by~\citet{liu2017learning}.
In our experiments, all the baseline (i.e.~unpruned) networks are trained from scratch with SGD. We train the networks for 150 epochs for CIFAR datasets and 300 epochs for Tiny-ImageNet with an initial learning rate of $0.1$ and weight decay of $2e^{-4}$. The learning rate is decayed by a factor of 10 at $\frac{1}{2}$ and $\frac{3}{4}$ of the total number of training epochs. For the networks trained with $L_1$ sparsity on BatchNorm,
we followed the same settings as in~\citet{liu2017learning}.

\subsection{One-pass Pruning Results}

We first consider the single-pass setting, where we perform a single round of pruning, and then fine-tune the network. 
Specifically, we compare eigenpruning\footnote{For EigenDamage, we count both the parameters of $\mW'$ and two eigenbasis.}~(EigenDamage) against our proposed baselines C-OBD, C-OBS, Kron-OBD, Kron-OBS and a state-of-the-art channel-level pruning method, NN Slimming, on CIFAR10 and CIFAR100 with VGGNet and (Pre)ResNet.
For all methods, we test a variety of pruning ratios, ranging from $0.5$ to $0.9$. Due to the space limit, please refer to Appendix~\ref{app:additional_one_pass_results} for the full results. In order to avoiding pruning all the channels in some layers, we constrain that at most $95\%$ of the channels can be pruned at each layer. After pruning, the network is finetuned for 150 epochs with an initial learning rate of $1e^{-3}$ and weight decay of $1e^{-4}$. The learning rate decay follows the same scheme as in training. We run each experiment $5$ times in order to reduce the variance of the results.

\textbf{Results on CIFAR datasets.} The results on CIFAR datasets are presented in Table~\ref{tab:one_pass_vgg_resnet}. It shows that even C-OBD and C-OBS can almost match NN slimming on CIFAR10 and CIFAR100 with VGGNet, if trained with $L_1$ sparsity on BatchNorm, and outperform when trained without it. Moreover, when the pruning ratio is $90\%$, two channel-level variants outperform NN Slimming on CIFAR100 with VGGNet by $\sim2\%$ in terms of test accuracy. For the experiments on ResNet, EigenDamage achieves better performance ($\sim2\%$) than others when the pruning ratio is $90\%$ on CIFAR-100 dataset. Besides, for the experiments on PreResNet, EigenDamage achieves the best performance in terms of test accuracy on all configurations and outperforms other baselines by a bigger margin.
\begin{table}[t]
\small
\centering
\vspace{-0.2cm}
\caption{One pass pruning on Tiny-ImageNet with VGG19. To be noted, the network for NN Slimming is pretrained with $L_1$ loss as required by the method. See Appendix~\ref{app:additional_one_pass_results} for the full results.}\label{tab:tiny_imagenet}
\vspace{-0.15cm}
\begin{center}
\resizebox{0.95\columnwidth}{!}{
\begin{tabular}{l|c c c}
\toprule
Prune Ratio (\%)  
& \multicolumn{3}{c}{50\%} 
\\
\hline
\multirow{2}{*}{Method }
& Test & Reduction in & Reduction in %
\\
& acc (\%) & weights (\%) & FLOPs (\%) %
\\
\hline
VGG19(Baseline) 
& 61.56 & - & - %
\\
VGG19+$L_1$(Baseline) 
& 60.68 & - & - %
\\
NN Slimming ~\cite{liu2017learning} 
& 50.90 $\pm$ - & 60.14 $\pm$ - & 85.42 $\pm$ - %
\\

C-OBD%
& 51.10 $\pm$ 0.60 & 69.27 $\pm$ 0.22 & 63.61 $\pm$ 0.19 %
\\

C-OBS%
& 53.13 $\pm$ 0.47 & 57.99 $\pm$ 0.52 & 78.51 $\pm$ 0.56 %
\\

Kron-OBD 
& 53.82 $\pm$ 0.32 & 67.22 $\pm$ 0.19 & 76.11 $\pm$ 0.24 %
\\

Kron-OBS 
& 53.54 $\pm$ 0.32 & 64.51 $\pm$ 0.23 & 74.57 $\pm$ 0.29 %
\\

EigenDamage 
& \textbf{58.20 $\pm$ 0.30} & 61.87 $\pm$ 0.11 & 66.21 $\pm$ 0.15 %
\\

\bottomrule
\end{tabular}
}
\end{center}
\vspace{-0.3cm}
\end{table}

\begin{figure}[t]

    \centering
    \includegraphics[width=1.0\columnwidth]{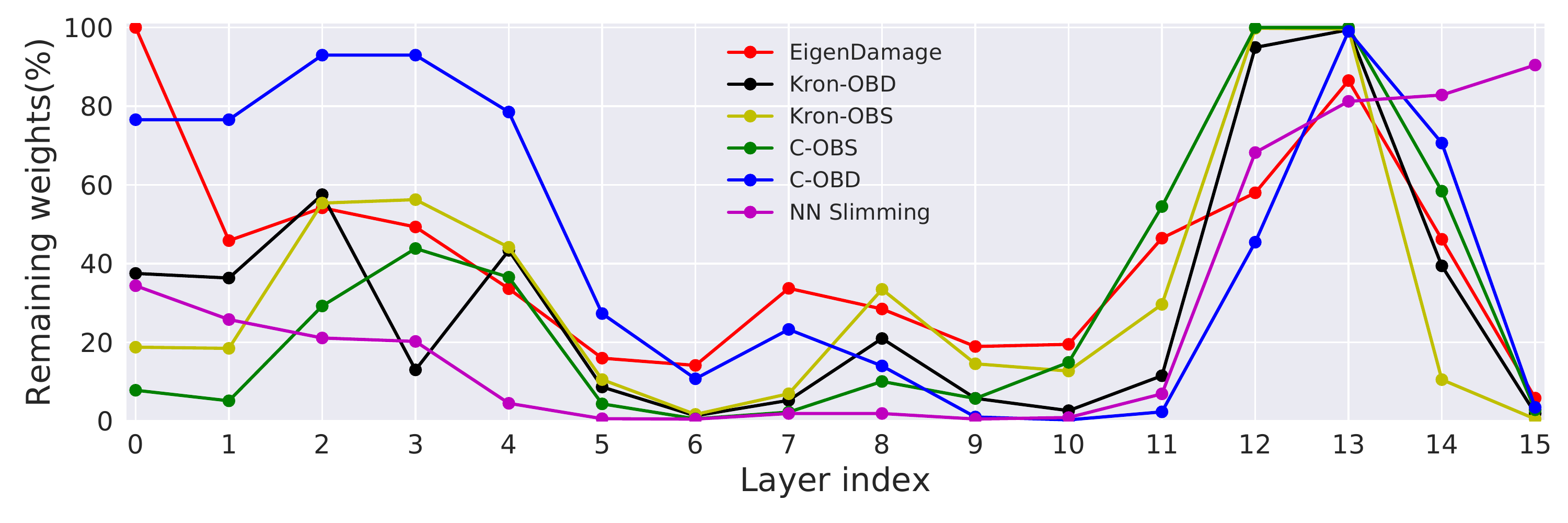}
    \vspace{-0.6cm}
    \caption{The percentage of remaining weights at each conv layer after one-pass pruning with a ratio of $0.5$ on Tiny-ImageNet with VGG19. The legend is sorted in descending order of test accuracy.}
    \label{fig:pruning-raio}
    \vspace{-0.5cm}
\end{figure}
\begin{figure*}[t]
     \centering
     \begin{subfigure}[b]{0.245\textwidth}
         \centering
         \includegraphics[width=\textwidth]{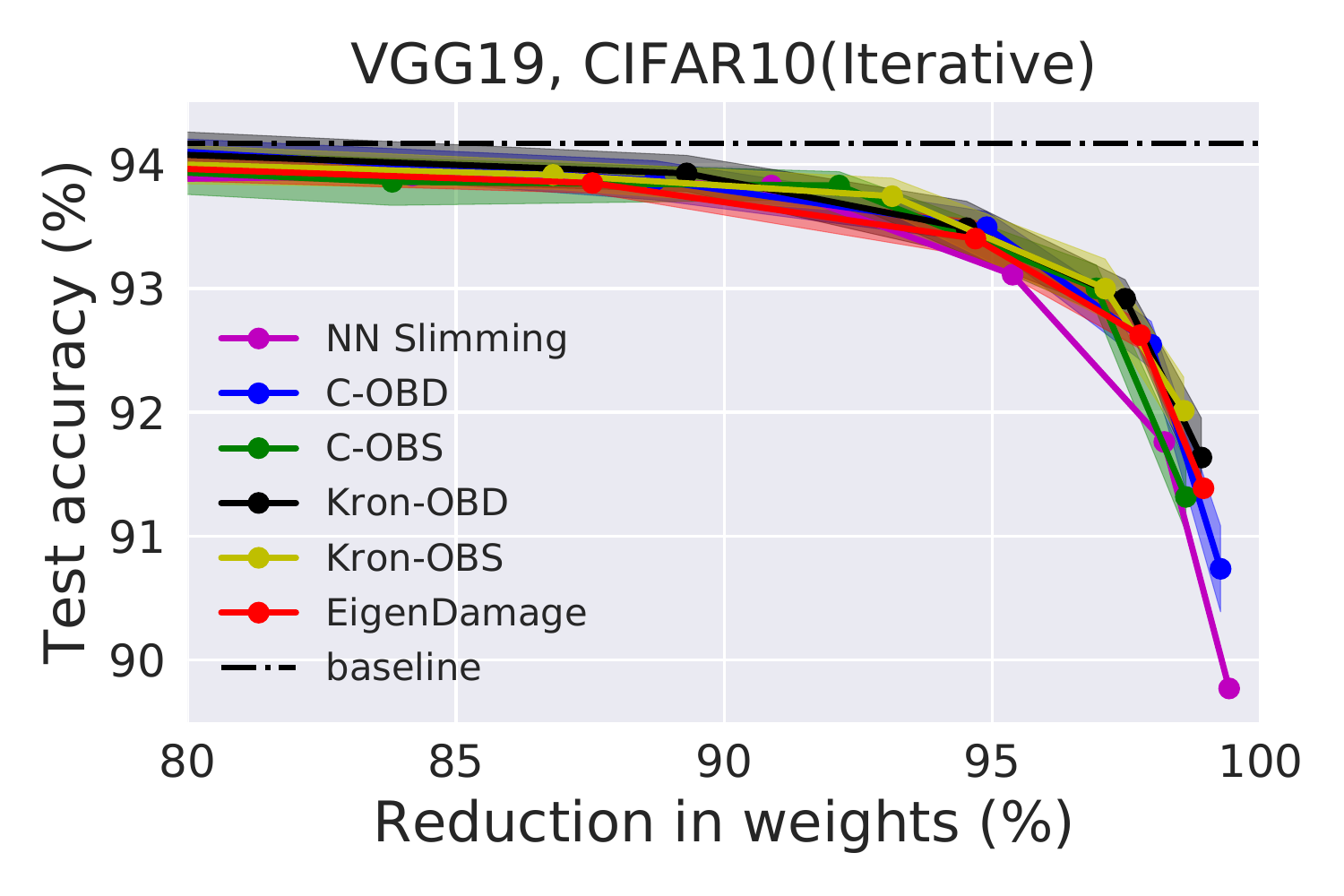}
     \end{subfigure}
     \hfill
     \begin{subfigure}[b]{0.245\textwidth}
         \centering
         \includegraphics[width=\textwidth]{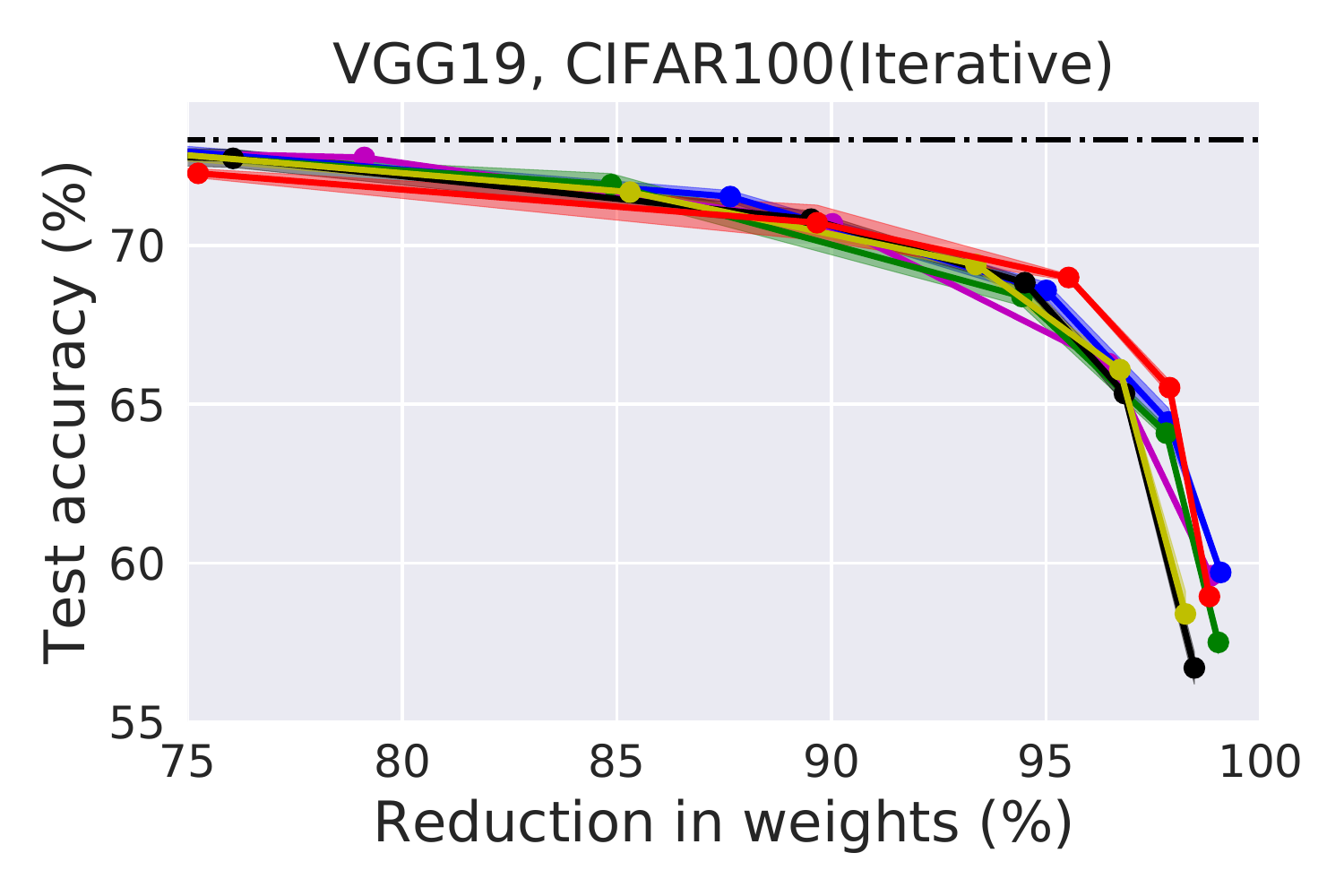}
     \end{subfigure}
     \hfill
     \begin{subfigure}[b]{0.245\textwidth}
         \centering
         \includegraphics[width=\textwidth]{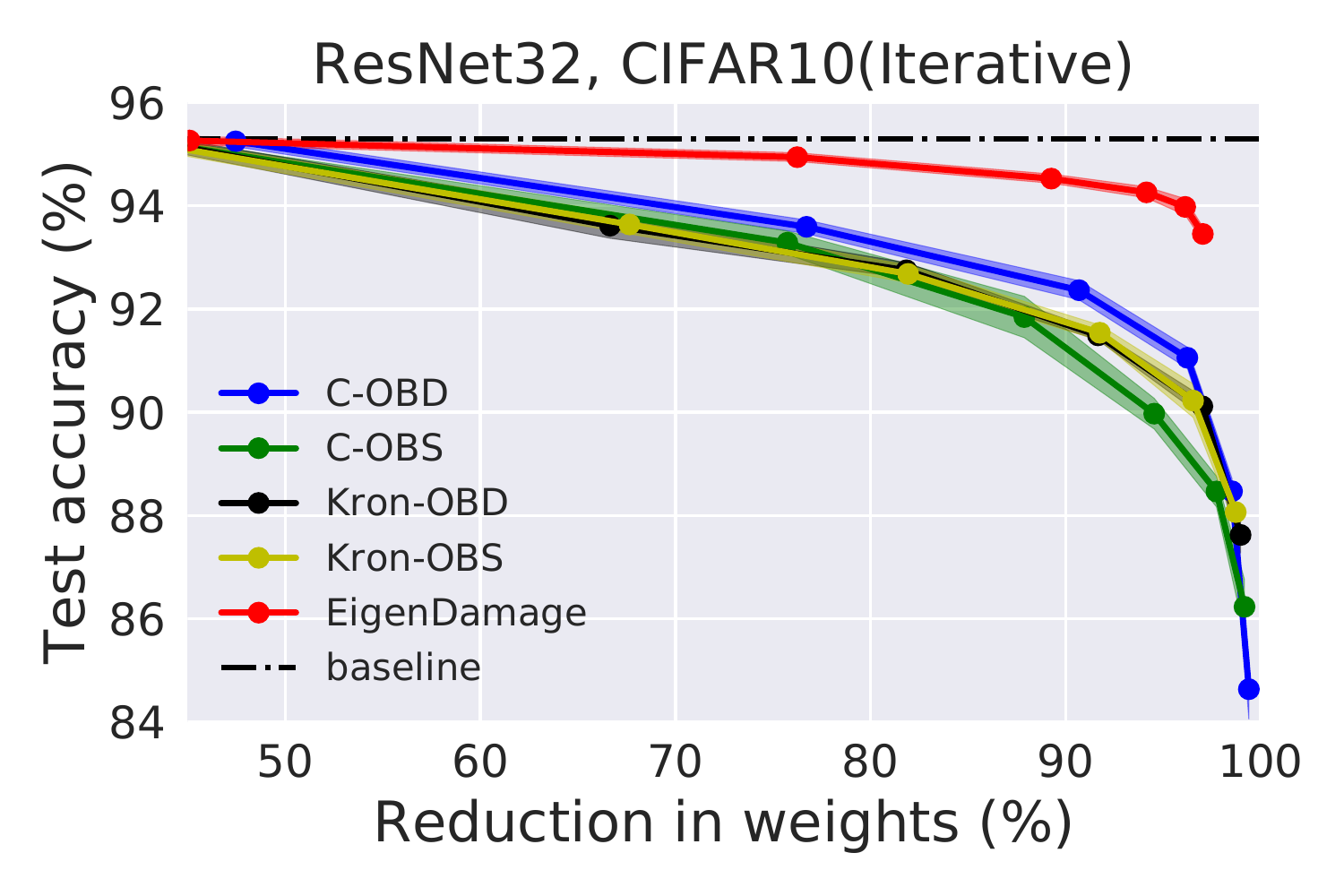}
     \end{subfigure}
     \hfill
     \begin{subfigure}[b]{0.245\textwidth}
         \centering
         \includegraphics[width=\textwidth]{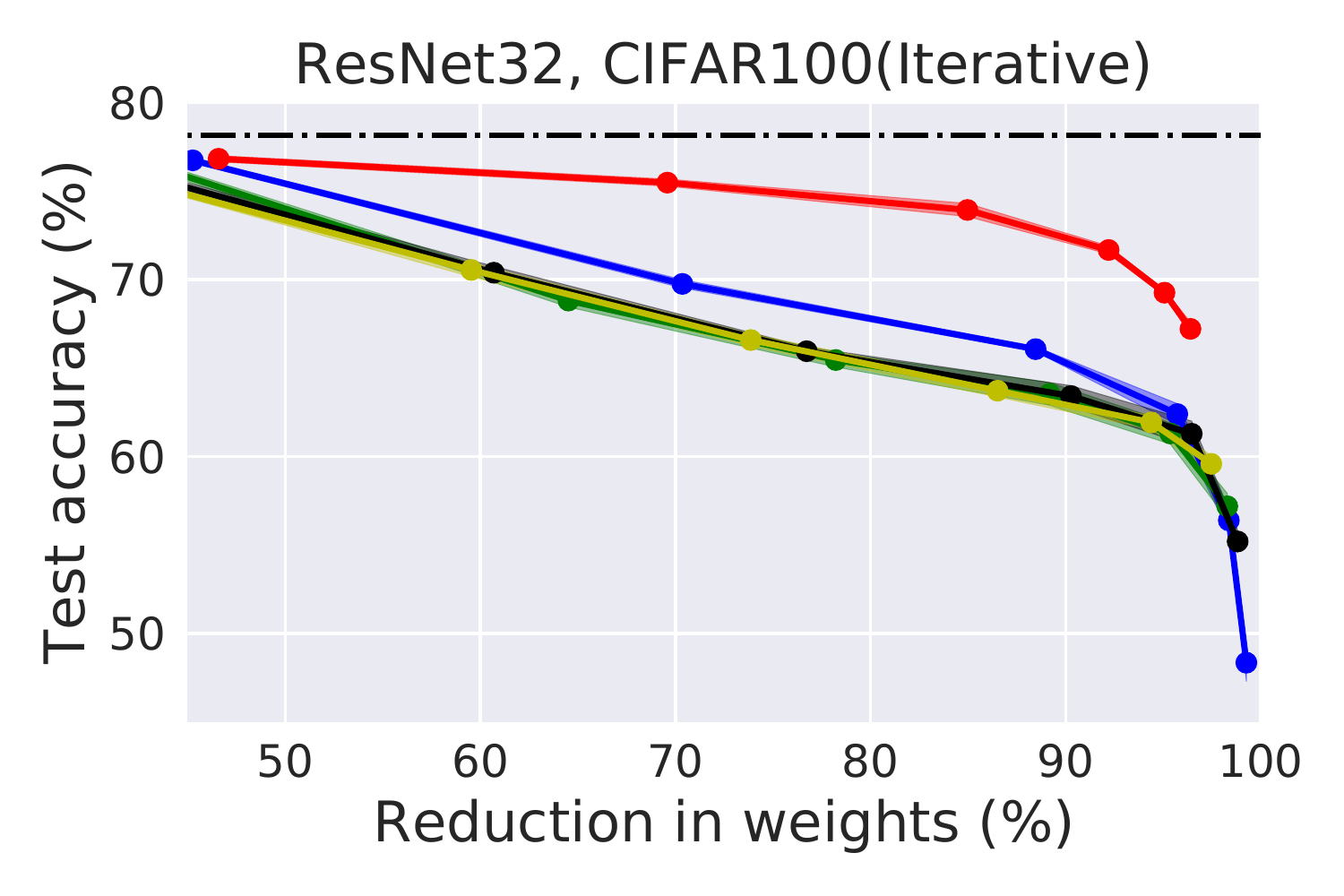}
     \end{subfigure}
     \begin{subfigure}[b]{0.245\textwidth}
         \centering
         \includegraphics[width=\textwidth]{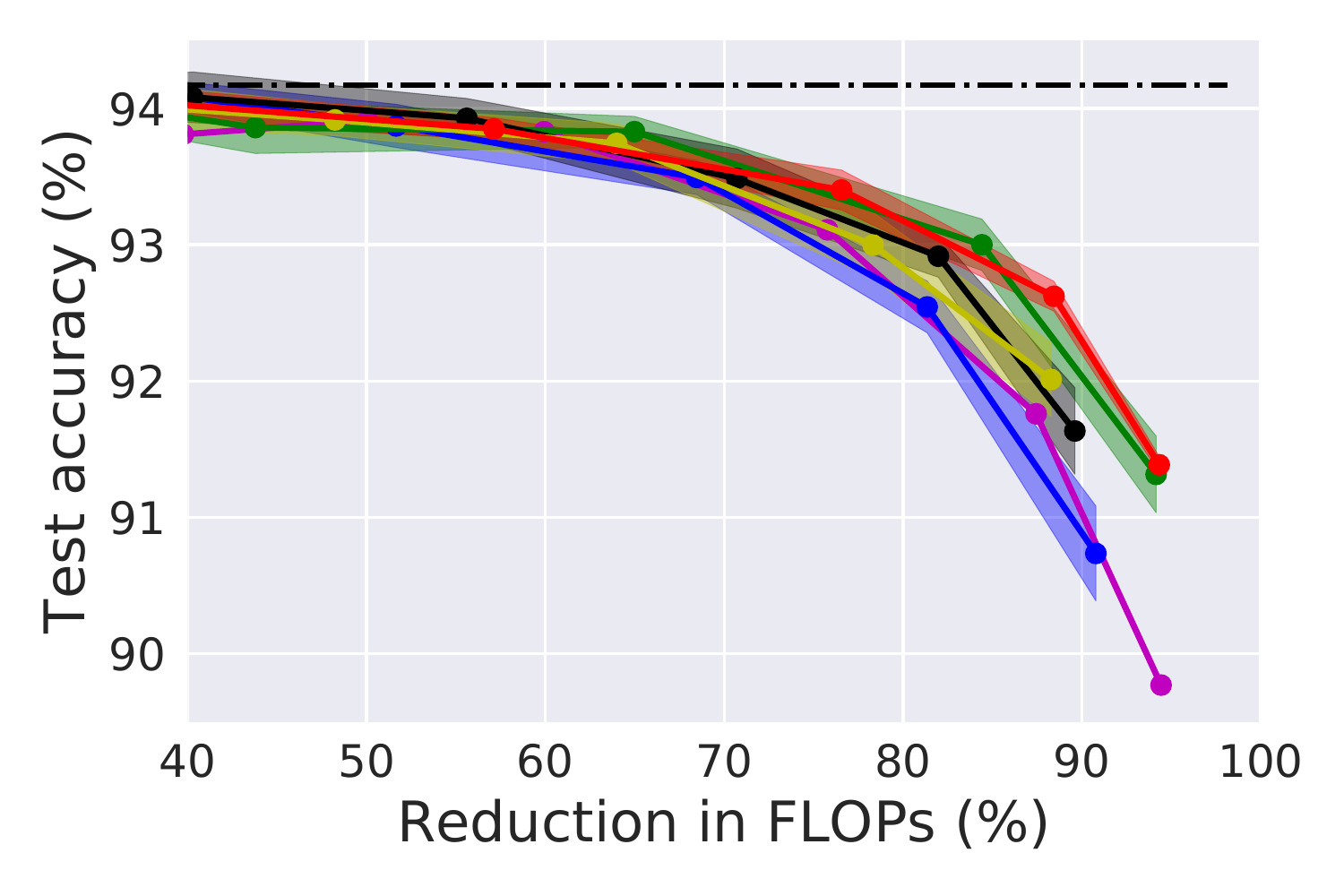}
     \end{subfigure}
     \hfill
     \begin{subfigure}[b]{0.245\textwidth}
         \centering
         \includegraphics[width=\textwidth]{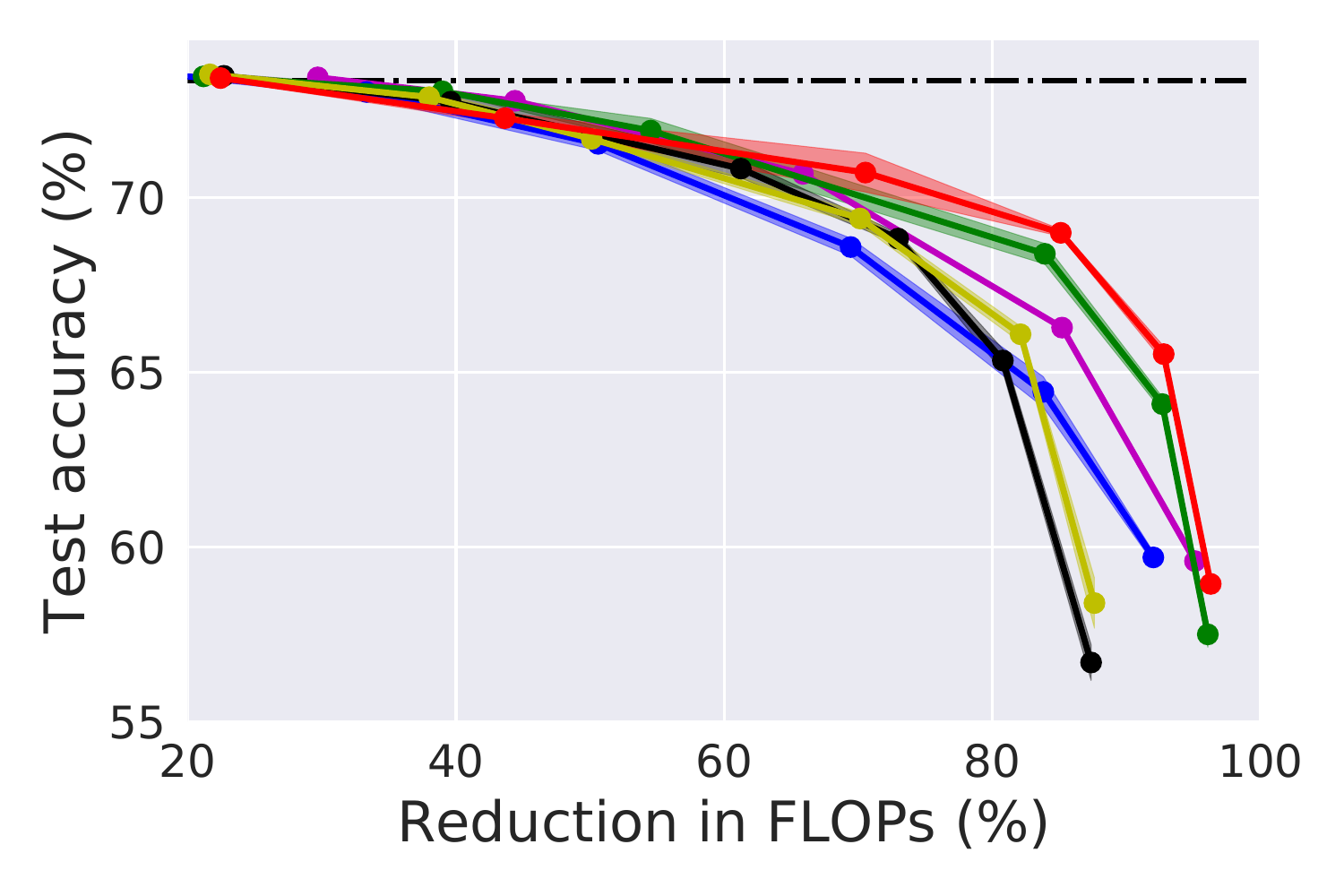}
     \end{subfigure}
     \hfill
     \begin{subfigure}[b]{0.245\textwidth}
         \centering
         \includegraphics[width=\textwidth]{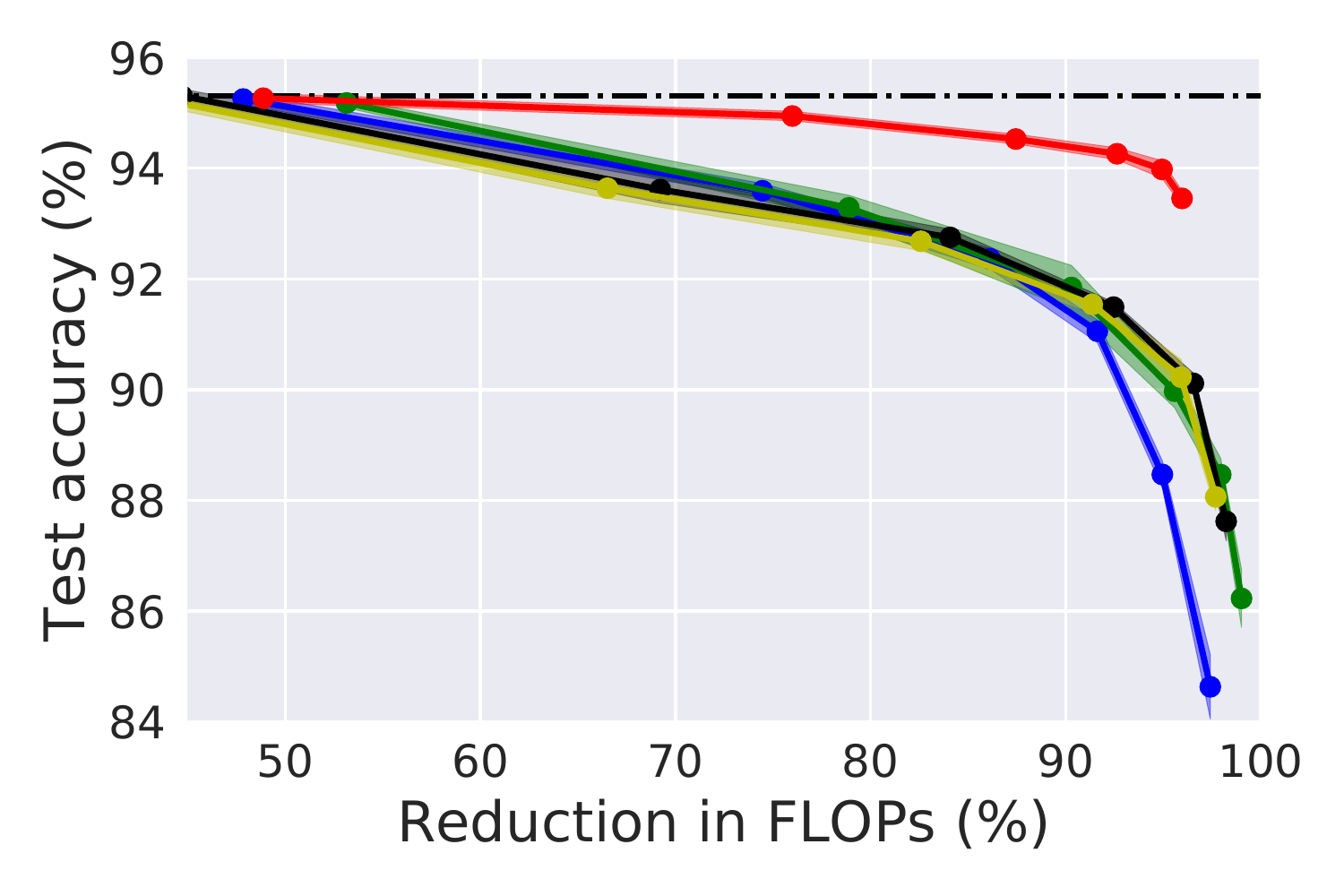}
     \end{subfigure}
     \hfill
     \begin{subfigure}[b]{0.245\textwidth}
         \centering
         \includegraphics[width=\textwidth]{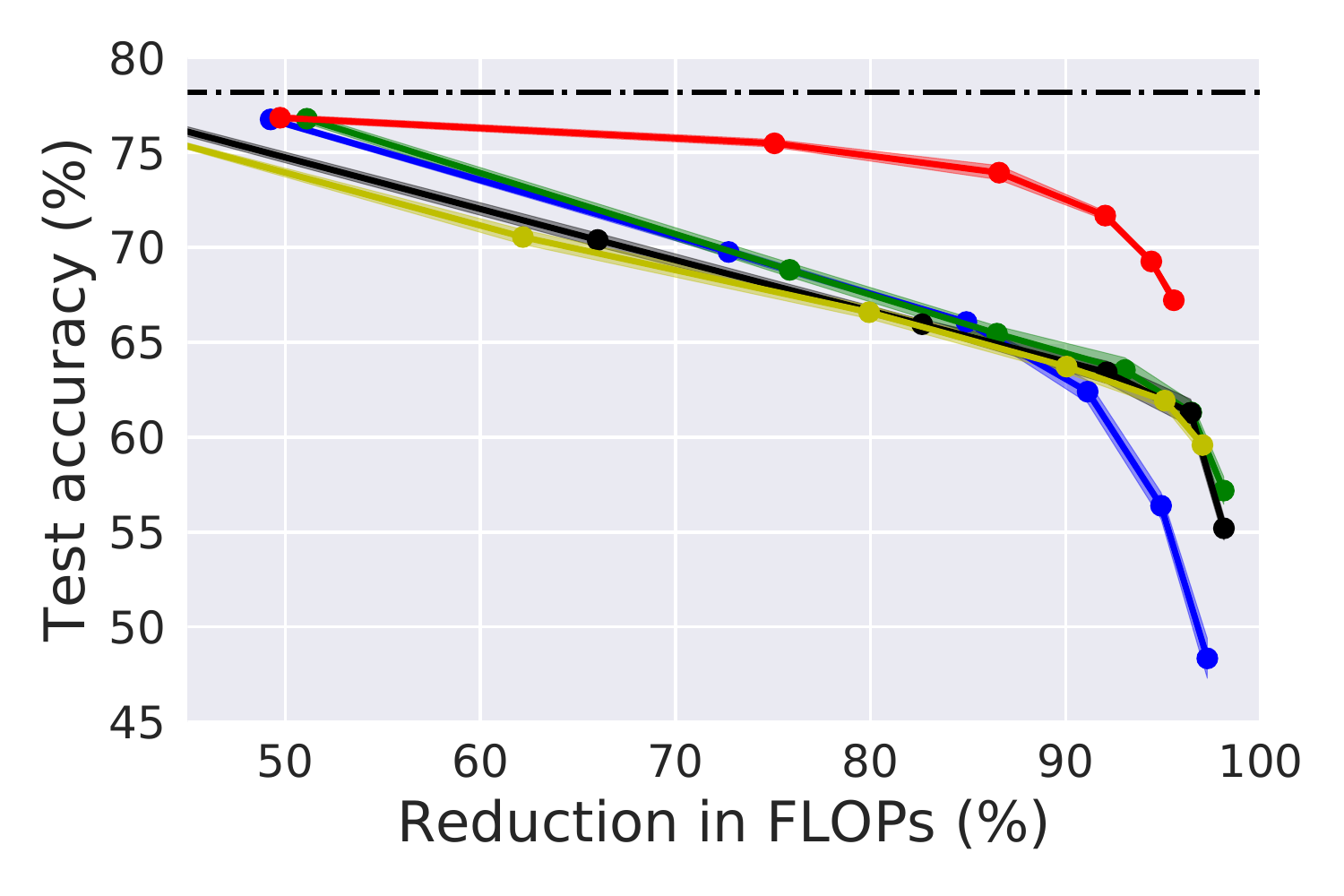}
     \end{subfigure}
     \vspace{-0.3cm}
        \caption{The results of iterative pruning. The first row are the curves of reduction in weights vs.~test accuracy, and second row are the curves of pruned FLOPs vs.~test accuracy of VGGNet and ResNet trained on CIFAR10 and CIFAR100 datasets. The shaded areas represent the variance over five runs. }
        \label{fig:iterative_compare}
    \vspace{-0.3cm}
\end{figure*}

To summarize, EigenDamage performs the best across almost all the settings, and the improvements become more significant when the pruning ratio is high, \eg$90\%$, especially on more complicated networks, \eg(Pre)ResNet, which demonstrates the effectiveness of pruning in the KFE. Moreover, EigenDamage adopts the bottleneck structure, which preserves the input and output dimension, as illustrated in Figure~\ref{fig:figure_1}, and thus can be trivially applied to any fully connected or convolution layer without modification.

\begin{figure}[t]
     \centering
      \includegraphics[width=1.0\columnwidth]{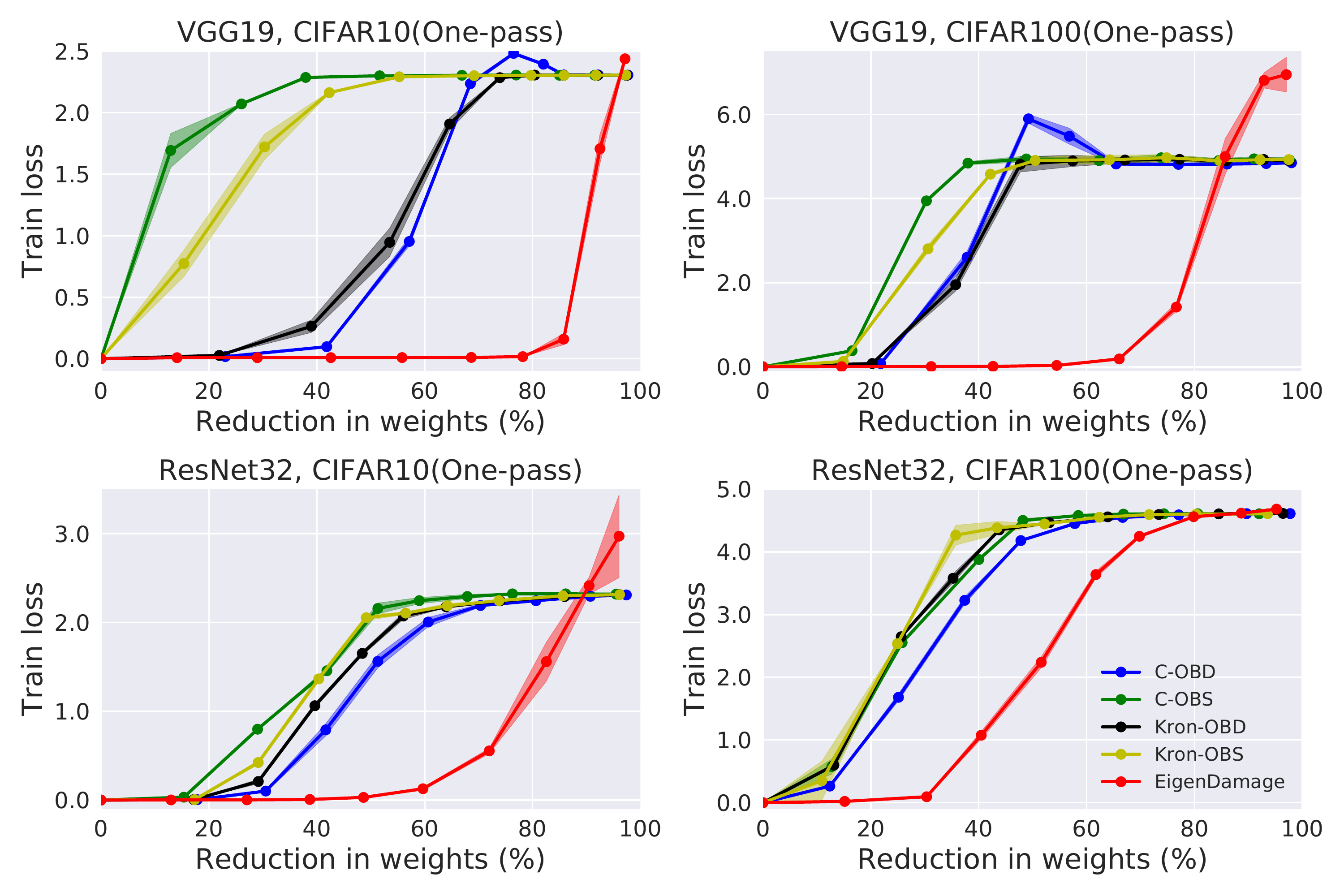}
     \vspace{-0.5cm}
        \caption{The above four figures show the training loss after one-pass pruning (without finetuning) vs.~reduction in weights. The network pruned by EigenDamage achieves significantly lower loss on the training set. This shows that pruning in the KFE is very accurate in reflecting the sensitivity of loss to the weights.}
    \label{fig:train_loss_fintune}
        \label{fig:train_loss_compare}
\end{figure}

As we mentioned in Section~\ref{sec:background_revisit}, the success of loss-aware pruning algorithms relies on the approximation to the loss function for identifying unimportant weights/filters. Therefore, we visualize the loss on training set after one-pass pruning (without fintuning) in Figure~\ref{fig:train_loss_compare}. For EigenDamage, we can see that for VGG19 on CIFAR10, when even prune $80\%$ of the weights, the increase in loss is negligible, and for other settings, the loss is also significantly lower than for other methods. For the remaining methods, which conduct pruning in the original weight space, they all result in a large increase in loss, and the resulting network performs similarly to uniform predictions in terms of loss. 

\begin{figure*}[t]
     \centering
     \includegraphics[width=1.0\textwidth]{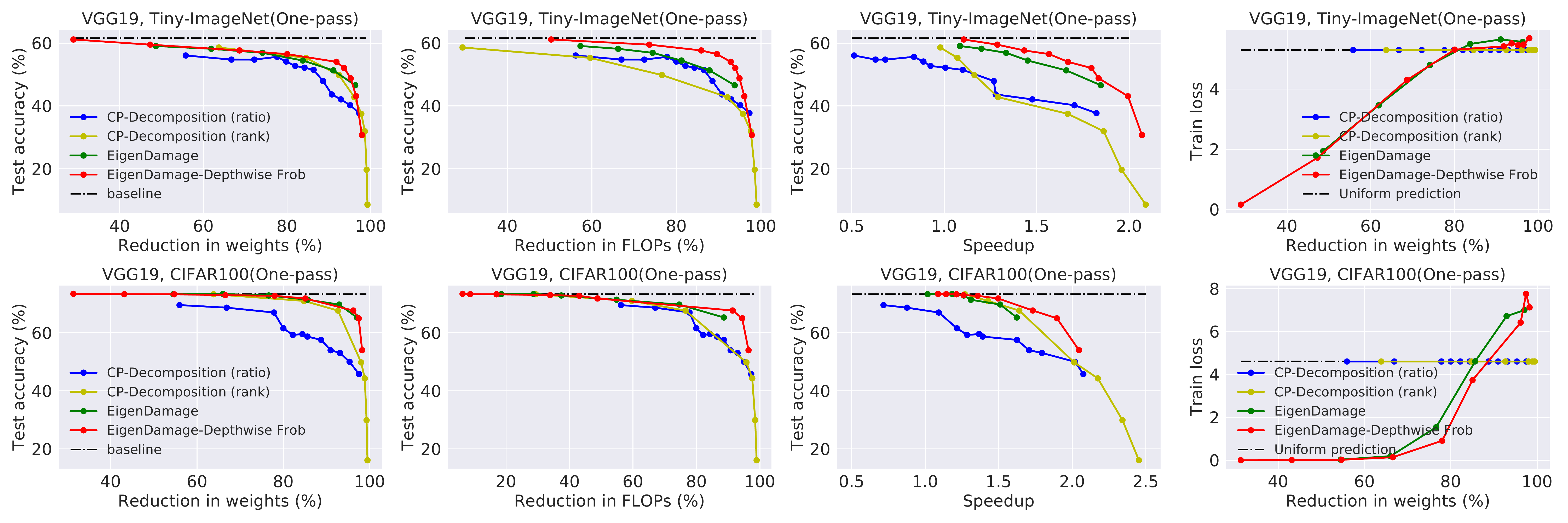}

     \vspace{-0.3cm}
        \caption{Low-rank approximation results on VGG19 on CIFAR100 and Tiny-ImageNet. The results are obtained by varying either the ranks of approximation or the pruning ratios. }
        \label{fig:low_rank}
\end{figure*}
\begin{figure}[t]
    \centering
    \begin{subfigure}[b]{0.49\columnwidth}
        \includegraphics[width=\textwidth]{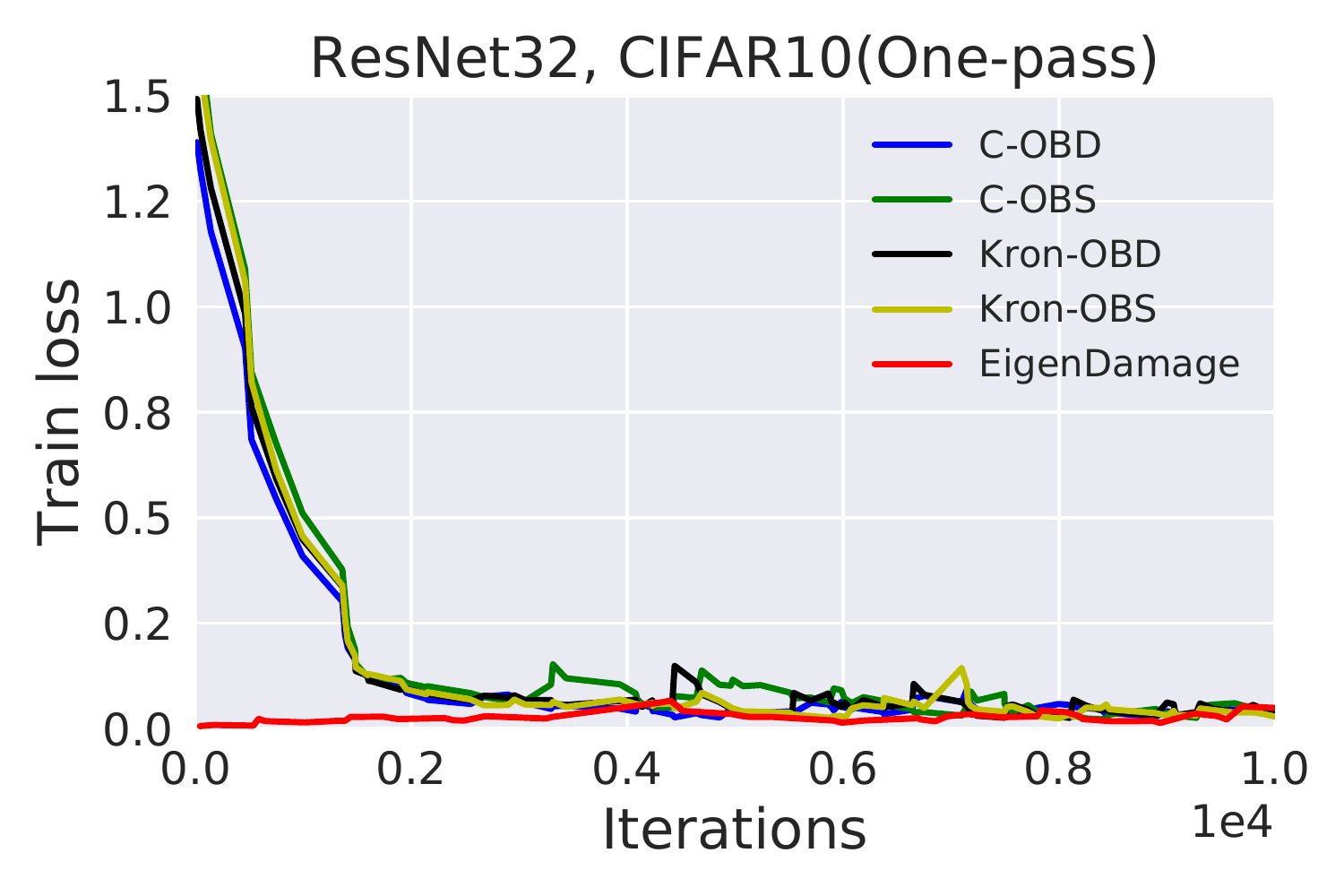}
    \end{subfigure}
    \begin{subfigure}[b]{0.49\columnwidth}
        \includegraphics[width=\textwidth]{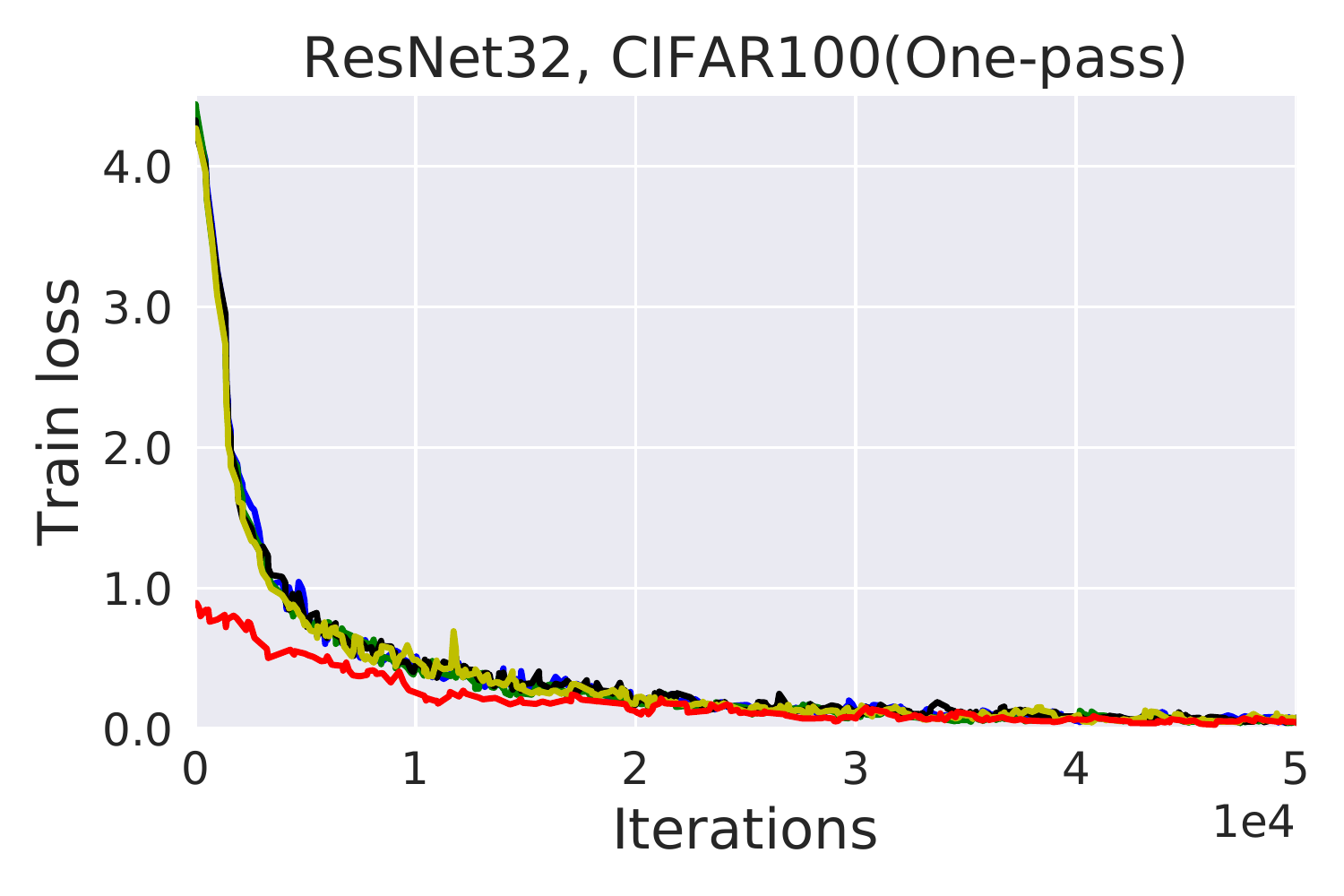}
    \end{subfigure}
    \vspace{-0.5cm}
    \caption{Loss on training set when finetuning the network after  pruning~(with a ratio of $0.5$) with ResNet32 on CIFAR10 and CIFAR100 datasets.}
    \label{fig:finetuning_curve}
\vspace{-0.5cm}
\end{figure}

\textbf{Results on Tiny-ImageNet dataset.} Apart from the results on CIFAR datasets, we futher test our methods on a more challenging dataset, Tiny-ImageNet, with VGGNet. Tiny-ImageNet consists of 200 classes and 500 images per class for training, and 10,000 images for testing, which are downsampled from the original ImageNet dataset. The results are in Table~\ref{tab:tiny_imagenet}. Again, EigenDamage outperforms all the baselines by a significant margin.

We further plot the pruning ratio in each convolution layer for a detailed analysis. As shown in Figure~\ref{fig:pruning-raio}, NN Slimming tends to prune more in the bottom layers but retain most of the filters in the top layers, which is undesirable since neural networks typically learn compact representations in the top. This may explain why NN Slimming performs worse than other methods in Tiny-ImageNet~(see Table~\ref{tab:tiny_imagenet}). By contrast, EigenDamage yields a balanced pruning ratio across different layers (retains most filters in bottom layers while pruning most redundant weights in top layer).

\subsection{Iterative Pruning Results}

We further experiment with the iterative setting, where the pruning can be conducted iteratively until it reaches a desired model size or FLOPs.
Concretely, the iterative pruning is conducted for $6$ times with a pruning ratio of $0.5$ at each iteration for simulating the process. 
In order to avoiding pruning the entire layer, we also adopt the same strategy as in~\citet{liu2017learning}, \ie we constrain that at most $50\%$ of the channels can be pruned in each layer for each iteration.

We compare EigenDamage to C-OBD, C-OBS, Kron-OBD and Kron-OBS. The results are summarized in Figure~\ref{fig:iterative_compare}. We notice that EigenDamage performs slightly better than other baselines with VGGNet and achieves significantly higher performance on ResNet.
Specifically, for the results on CIFAR10 dataset with VGGNet, nearly all the methods achieved similar results due to the simplicity of CIFAR10 and VGGNet. However, the performance gap is a bit more clear as the dataset becoming more challenging, \eg, CIFAR100. 
On a more sophisticated network, ResNet, the performance improvements of EigenDamage were especially significant on CIFAR10 or CIFAR100. Furthermore, EigenDamage was especially effective in reducing the number of FLOPs, due to the bottleneck structure.

\subsection{Comparisons with Low-rank Approximation}

Since EigenDamage can also be viewed as low-rank approximation, we compared it with a state-of-the-art low-rank method, CP-Decomposition~\citep{lebedev2014speeding}, which computes a low-rank decomposition of the filter into a sum of rank-one tensors. We experimented low-rank approximation for VGG19 on CIFAR100 and Tiny-ImageNet. For CP-Decomposition, we tested it under two settings: (1) we varied the ranks from $\{0.1, \dots, 1.0, 1.25, 1.5, 2.0\}$ times of the original rank at each layer; (2) we varied ranks in $\{4, 8, \dots, 512\}$ for computing the approximation\footnote{We choose the minimum of the target rank and the original rank of the convolution filter as the rank for approximation.}. For EigenDamage, we chose different pruning ratios in the range of $\{0,4, \dots, 0.9\}$, and EigenDamage-Depthwise Frob is obtained by applying depthwise separable decomposition on the network obtained by EigenDamage.

The results are presented in Figure~\ref{fig:low_rank}. EigenDamage outperforms CP-Decomposition significantly in terms of speedup and accuracy. Moreover, CP-Decomposition approximates the original weights under the Frobenius norm in the original weight coordinates, which does not precisely reflect the sensitivity to the training loss. In contrast, EigenDamage is loss-aware, and thus the resulting approximation will achieve lower training loss when only pruning is applied, \ie~without finetuning, as is shown in the Figure~\ref{fig:low_rank}. Note that EigenDamage will determine the approximation rank for each layer automatically given a global pruning ratio. However, CP-Decomposition requires pre-determined approximation rank for each layer and thus the search complexity will grow exponentially in the number of layers.

\section{Conclusion}
In this paper, we introduced a novel network reparameterization based on the Kronecker-factored eigenbasis, in which the entrywise independence assumption is approximately satisfied. This lets us prune the weights effectively using Hessian-based pruning methods. The pruned networks give low-rank (bottleneck structure) which allows for fast computation.
Empirically, EigenDamage outperforms strong baselines which do pruning in original parameter coordinates, especially on more chanllenging datasets and networks.

\section*{Acknowledgements}
We thank Shengyang Sun, Ricky Chen, David Duvenaud, Jonathan Lorraine for their feedback on early drafts. GZ was funded by an MRIS Early Researcher Award.
\newpage
\nocite{langley00}

\bibliography{example_paper}
\bibliographystyle{icml2019}

\onecolumn
\appendix
\icmltitle{Supplementary Material}

\section{Derivation of Kron-OBD and Kron-OBS}\label{app:derivation}
\textbf{Derivation of Kron-OBD.} Assuming that the weight of a conv layer is $\vtheta=\mathrm{vec}\left(\mW\right)$, where $\mW \in \mathbb{R}^{n \times m}$,  $n=c_\mathrm{in}k^2$ and $m=c_\mathrm{out}$, and the two Kronecker factors are $\mS\in \mathbb{R}^{m\times m}$ and $\mA \in \mathbb{R}^{n \times n}$. Then the Fisher information matrix of $\vtheta$ can be approximated by $\mF = \mS\kron \mA$. Substituting the Hessian with K-FAC Fisher in eqn.~\eqref{eq:obd-f1-objective}, we get:
\begin{equation}
\begin{aligned}
    \Delta \loss &= \frac{1}{2}\Delta \vtheta^\top \left(\mS \kron \mA\right) \Delta \vtheta = \frac{1}{2}\mathrm{Tr}\left(\Delta\mW^\top\mA\Delta\mW\mS\right) = \frac{1}{2}\sum_{i,j}\mS_{ij}\Delta \vtheta_i^\top\mA\Delta\vtheta_j
\end{aligned}
\end{equation}
where $\Delta\vtheta_i$ represents the change in ${\vtheta_i^*}$, and ${\vtheta_i^*} \in \mathbb{R}^n$ is the weight of $i$-th filter $\mathcal{F}_i$, \ie, $i$-th column of $\mW$.
Under the assumption that each filter is independent to each other, and thus $\mS$ is diagonal. So, we can get the importance of each filter and the corresponding change in weights are:
\begin{equation}\label{app:obd_objective}
    \Delta \loss_i = \frac{1}{2} \mS_{ii}{\vtheta_i^*}^\top\mA{\vtheta_i^*} \;\; \mathrm{and} \;\; \Delta \vtheta_i = -{\vtheta_i^*}
\end{equation}

\textbf{Derivation of Kron-OBS.} Under the assumption of Kron-OBS that different filters are correlated to each other, $\mS$ is no longer diagonal. Then, similar to eqn.~\eqref{app:obd_objective}, the corresponding structured version of eqn.\eqref{eq:OBS_objective} becomes:
\begin{equation}\label{app:obs_objective}
    \min_{i}\left\{\min_{\Delta\mW}\frac{1}{2}\mathrm{Tr}\left(\Delta\mW^\top\mA\Delta\mW\mS\right)\right\} \;\; \mathrm{s.t.} \;\; \Delta\mW\ve_i + {\vtheta_i^*} = \mathbf{0}
\end{equation}
We can solve the above constrained optimization problem with Lagrange multiplier:
\begin{equation}
    \min_{i}\left\{\min_{\Delta \mW}\frac{1}{2}\mathrm{Tr}\left(\Delta\mW^\top\mA\Delta\mW\mS\right) - \vlambda^\top\left(\Delta\mW\ve_i + {\vtheta_i^*}\right)\right\}
\end{equation}
Taking the derivatives w.r.t to $\Delta\mW$ and set it to $\mathbf{0}$, we get:
\begin{equation}\label{app:theta}
    \Delta \mW = \mA^{-1}\vlambda\ve_i^\top\mS^{-1}
\end{equation}
Substitute it back to the constrain to solve the equation, we get:
\begin{equation}\label{app:lambda}
    \vlambda = \frac{-\mA{\vtheta_i^*}}{[\mS^{-1}]_{ii}}
\end{equation}
Then substitute eqn.~\eqref{app:lambda} back to eqn.~\eqref{app:theta}, we can finally get the optimal change in weights if we remove filter $\mathcal{F}_i$:
\begin{equation}\label{app:obs_delta_theta}
    \Delta\mW = -\frac{{\vtheta_i^*}}{[\mS^{-1}]_{ii}}\ve_i^\top\mS^{-1} \;\; \mathrm{and} \;\; \Delta\vtheta = - \frac{\mS^{-1}\ve_i\kron{\vtheta_i^*}}{[\mS^{-1}]_{ii}}
\end{equation}
In order to evaluating the importance of each filter, we can substitute eqn.~\eqref{app:obs_delta_theta} back to eqn.~\eqref{app:obs_objective}:
\begin{equation}
    \begin{aligned}
        \Delta \loss_i &= \frac{1}{2}\mathrm{Tr}\left(\mS^{-1}\ve_i\frac{{\vtheta_i^*}^\top}{[\mS^{-1}]_{ii}}\mA\frac{\vtheta_i^*}{[\mS^{-1}]_{ii}}\ve_i^\top\mS^{-1}\mS\right)
        = \frac{1}{2}\mathrm{Tr}\left(\frac{{\vtheta_i^*}^\top \mA {\vtheta_i^*}}{[\mS^{-1}]_{ii}^{2}}\mS^{-1}\ve_i\ve_i^\top\right)\\
        &= \frac{1}{2}\mathrm{Tr}\left(\frac{{\vtheta_i^*}^\top\mA{\vtheta_i^*}}{[\mS^{-1}]_{ii}^2}[\mS^{-1}]_{ii}\right)
        = \frac{1}{2}\frac{{\vtheta_i^*}^\top\mA{\vtheta_i^*}}{[\mS^{-1}]_{ii}}
    \end{aligned}
\end{equation}

\section{Algorithm for Solving eqn.~\eqref{eq:frob_norm}}
\label{app:als}
In this section, we will introduce the algorithm for solving the optimization problem in eqn.~\eqref{eq:frob_norm}.

\textbf{Khatri-Rao product.} The  Khatri-Rao product $\odot$ of two matrices $\mA \in \mathbb{R}^{m\times r}$ and $\mB \in \mathbb{R}^{n\times r}$ is the column-wise Kronecker product, that is:
\begin{equation}\label{eq:khatri_rao_product}
	\mA\odot \mB = \begin{pmatrix} a_{11}b_{11} & a_{12}b_{12} & \cdots & a_{1r}b_{1r} \\ a_{11}b_{21} & a_{12}b_{22} & \cdots & a_{1r}b_{2r} \\ \vdots & \vdots & \ddots & \vdots \\ a_{m1}b_{n1} & a_{m2}b_{n2} & \cdots & a_{mr}b_{nr} \end{pmatrix} \in \mathbb{R}^{mn\times r}
\end{equation}

\textbf{Kruskal tensor notation.} Suppose $\mT\in \mathbb{R}^{n_1\times n_2\times \cdots \times n_d}$ has low-rank  Canonical Polyadic~(CP) structure. Following~\cite{bader2007efficient}, we refer to it as a \emph{Kruskal tensor}. Normally, it can be defined by a collection of factor matrices, $\mA_k \in \mathbb{R}^{n_k\times r}$ for $k=1, ..., d$, such that:
\begin{equation}
	\mT(i_1,i_2,\cdots,i_d) = \sum_{j=1}^r \mA_1(i_1, j)\mA_2(i_2, j)\cdots\mA_d(i_d, j) \; \; \;  \; \mathrm{for\; all \;} (i_1, i_2, \cdots, i_d) \in \mathcal{I}
\end{equation}
where $\mathcal{I} \equiv \{1,\cdots,n_1\}\kron\{1,\cdots\,n_2\}\kron\cdots\kron\{1,\cdots,n_d\}$. Denote $\mT_{(k)} \in \mathbb{R}^{n_k\times (n_d\cdots n_{k-1}n_{k+1}\cdots n_1)}$ is the mode-$k$ unfolding of a Kruskal tensor, which has the following form that depends on the Khatri-Rao products of the factor matrices:
\begin{equation}
	\mT_{(k)} = \mA_k\mZ_k^\top \; \; \; \; \mathrm{where} \; \; \mZ_k \equiv \mA_d \odot \cdots \odot \mA_{k+1}\odot \mA_{k-1}\odot\cdots\odot\mA_1
\end{equation}

\textbf{Alternating Least Squares~(ALS).} We can use ALS to solve problems similar to eqn.~\eqref{eq:frob_norm}. Suppose we are approximating $\mT$ using $\mA_1, \cdots, \mA_d$. Specifically, for fixed $\mA_1,\cdots,\mA_{k-1},\mA_{k+1},\cdots,\mA_d$, there is a closed form solution for $\mA_k$. Specifically, we can update update $\mA_1, \cdots, \mA_d$ by the following update rule:
\begin{equation}
	\mA_k^\top =  \mZ_k^\dagger\mT_{(k)}^\top  \; \; \; \; \mathrm{for\; }k=1,\cdots,d
\end{equation}
alternatively until converge or reach the maximum number of iterations. For the Mahalanobis norm case~(with $\mF$ as the metric tensor), if we take the derivative with respect to $\mA_k$ to be $\mathbf{0}$,
\begin{equation}
	\mathrm{unvec}(\mF\mathrm{vec}(\mA_{k}\mZ_{k}^\top-\mT_{(k)}))\mZ = \mathbf{0}
\end{equation}
we can get the corresponding update rule for $\mA_k$:
\begin{equation}
	\mA_{k}^\top = \mZ_k^\dagger\left(\mT_{(k)} + \mathrm{unvec}(\mF^{-1}\mathrm{vec}(\mP))\right)^\top
\end{equation}
where unvec and vec are inverse operators to each other, and in our case, unvec operation is to convert the vectorized matrix back to the original matrix form. $\mZ^\dagger=(\mZ^\top\mZ)^{-1}\mZ^\top$ and $\mP$ has the same shape with $\mT_{k}$, and for each column $\mP_i \in \mathrm{Null}(\mZ_k^\top)$.

\section{Additional Results on One-pass Pruning}
We present the additional results on one-pass pruning in the following tables. We also present the data in tables as trade-off curves in terms of acc vs. reduction in weight and acc vs. reduction in FLOPs for making it easy to tell the difference in performances of each method.
\label{app:additional_one_pass_results}
\begin{table*}[h]
\small
\vspace{-0.5cm}
\caption{One pass pruning on CIFAR-10 with VGG19}
\vspace{-0.3cm}
\label{tab:classification}
\begin{center}
\resizebox{\textwidth}{!}{
\begin{tabular}{l|c c c|c c c|c c c}
\toprule
Prune Ratio (\%)  
& \multicolumn{3}{c|}{50\%} 
& \multicolumn{3}{c|}{70\%}
& \multicolumn{3}{c}{80\%} \\
\hline
\multirow{2}{*}{Method }
 & Test & Reduction in & Reduction in & Test & Reduction in & Reduction in & Test & Reduction in & Reduction in \\
& acc (\%) & weights (\%) & FLOPs (\%)       
& acc (\%) & weights (\%) & FLOPs (\%)  
& acc (\%) & weights (\%) & FLOPs (\%)    \\
\hline
VGG19(Baseline) 
& 94.17 & - & -
& -     & - & -
& -     & - & -\\
NN Slimming~\citep{liu2017learning} 
& 92.84 $\pm$ - & 73.84 $\pm$ - & 38.88 $\pm$ -
& 92.89 $\pm$ - & 84.30 $\pm$ - & 54.83 $\pm$ -
& 91.92 $\pm$ - & 91.77 $\pm$ - & 76.43 $\pm$ - \\

C-OBD%
& 94.01 $\pm$ 0.15 & 76.84 $\pm$ 0.30 & 35.07 $\pm$ 0.38
& 94.04 $\pm$ 0.09 & 85.88 $\pm$ 0.10 & 41.17 $\pm$ 0.23
& 93.70 $\pm$ 0.07 & 92.17 $\pm$ 0.07 & 56.87 $\pm$ 0.33 \\

C-OBS%
& \textbf{94.19 $\pm$ 0.10} & 66.91 $\pm$ 0.08 & 26.12 $\pm$ 0.13
& 93.97 $\pm$ 0.16 & 84.97 $\pm$ 0.02 & 43.16 $\pm$ 0.20
& 93.77 $\pm$ 0.12 & 91.52 $\pm$ 0.09 & 63.64 $\pm$ 0.13 \\

Kron-OBD 
& 93.91 $\pm$ 0.16 & 73.93 $\pm$ 0.42 & 33.71 $\pm$ 0.69
& 93.95 $\pm$ 0.12 & 85.80 $\pm$ 0.09 & 43.78 $\pm$ 0.24
& 93.78 $\pm$ 0.17 & 92.04 $\pm$ 0.04 & 60.81 $\pm$ 0.26 \\

Kron-OBS 
& 94.03 $\pm$ 0.13 & 69.17 $\pm$ 0.20 & 28.02 $\pm$ 0.26
& 94.10 $\pm$ 0.15 & 85.83 $\pm$ 0.09 & 42.56 $\pm$ 0.14
& \textbf{93.87 $\pm$ 0.14} & 92.00 $\pm$ 0.04 & 60.19 $\pm$ 0.38 \\

EigenDamage 
& 94.15 $\pm$ 0.05 & 68.64 $\pm$ 0.19 & 28.09 $\pm$ 0.21
& \textbf{94.15 $\pm$ 0.14} & 85.78 $\pm$ 0.06 & 45.68 $\pm$ 0.31
& 93.68 $\pm$ 0.22 & 92.51 $\pm$ 0.05 & 66.98 $\pm$ 0.36 \\

\hline
VGG19+$L_1$ (Baseline) 
& 93.71 & -  & -
& -     & -  & -
& -     & -  & -\\

NN Slimming~\citep{liu2017learning} 
& 93.79 $\pm$ - &  77.44 $\pm$ - & 45.19 $\pm$ -
& 93.74 $\pm$ - &  88.81 $\pm$ - & 52.15 $\pm$ -
& \textbf{93.48 $\pm$ -} &  92.60 $\pm$ - & 62.23 $\pm$ - \\

C-OBD%
& 93.85 $\pm$ 0.03 & 76.83 $\pm$ 0.01 & 41.14 $\pm$ 0.05 
& 93.88 $\pm$ 0.03 & 89.04 $\pm$ 0.03 & 52.73 $\pm$ 0.14
& 93.38 $\pm$ 0.04 & 93.29 $\pm$ 0.05 & 63.74 $\pm$ 0.12 \\

C-OBS%
& \textbf{93.88 $\pm$ 0.04} & 74.95 $\pm$ 0.03 & 37.56 $\pm$ 0.06
& 93.84 $\pm$ 0.04 & 88.53 $\pm$ 0.20 & 51.88 $\pm$ 0.00
& 93.27 $\pm$ 0.04 & 92.01 $\pm$ 0.02 & 63.96 $\pm$ 0.10 \\

Kron-OBD 
& \textbf{93.88 $\pm$ 0.01} & 83.43 $\pm$ 0.00 & 49.58 $\pm$ 0.00
& \textbf{93.89 $\pm$ 0.03} & 89.02 $\pm$ 0.01 & 53.40 $\pm$ 0.09
& 93.33 $\pm$ 0.05 & 93.55 $\pm$ 0.06 & 67.01 $\pm$ 0.30 \\

Kron-OBS 
& 93.85 $\pm$ 0.03 & 76.95 $\pm$ 0.01 & 42.04 $\pm$ 0.10
& 93.88 $\pm$ 0.04 & 88.69 $\pm$ 0.02 & 52.38 $\pm$ 0.08
& 93.44 $\pm$ 0.07 & 92.66 $\pm$ 0.05 & 63.77 $\pm$ 0.27\\

EigenDamage 
& 93.84 $\pm$ 0.04 & 78.14 $\pm$ 0.11 & 39.02 $\pm$ 0.30
& 93.85 $\pm$ 0.04 & 85.71 $\pm$ 0.01 & 46.56 $\pm$ 0.03
& 93.40 $\pm$ 0.07 & 91.48 $\pm$ 0.06 & 62.18 $\pm$ 0.29\\

\bottomrule
\end{tabular}
}
\end{center}
\vspace{-0.3cm}
\end{table*}

\begin{table*}[h]
\small
\vspace{-0.5cm}
\caption{One pass pruning on CIFAR-100 with VGG19}
\vspace{-0.3cm}
\label{tab:classification}
\begin{center}
\resizebox{\textwidth}{!}{
\begin{tabular}{l|c c c|c c c|c c c}
\toprule
Prune Ratio (\%)  
& \multicolumn{3}{c|}{50\%} 
& \multicolumn{3}{c|}{70\%}
& \multicolumn{3}{c}{80\%} \\
\hline
\multirow{2}{*}{Method }
 & Test & Reduction in & Reduction in & Test & Reduction in & Reduction in & Test & Reduction in & Reduction in \\
& acc (\%) & weights (\%) & FLOPs (\%)       
& acc (\%) & weights (\%) & FLOPs (\%)  
& acc (\%) & weights (\%) & FLOPs (\%)      \\
\hline
VGG19(Baseline) 
& 73.34 & - & -
& -     & - & -
& -     & - & - \\
NN Slimming~\citep{liu2017learning} 
& 72.77 $\pm$ - & 66.50 $\pm$ - & 30.61 $\pm$ -
& 69.98 $\pm$ - & 85.56 $\pm$ - & 54.51 $\pm$ -
& 66.09 $\pm$ - & 92.33 $\pm$ - & 76.76 $\pm$ - \\

C-OBD%
& 72.82 $\pm$ 0.15 & 65.47 $\pm$ 0.13 & 24.24 $\pm$ 0.10
& 71.10 $\pm$ 0.22 & 86.06 $\pm$ 0.04 & 41.18 $\pm$ 0.04
& 67.46 $\pm$ 0.26 & 93.31 $\pm$ 0.06 & 60.39 $\pm$ 0.25 \\

C-OBS%
& 72.73 $\pm$ 0.17 & 62.31 $\pm$ 0.05 & 25.50 $\pm$ 0.06
& 71.25 $\pm$ 0.21 & 84.49 $\pm$ 0.04 & 49.25 $\pm$ 0.52
& 67.47 $\pm$ 0.13 & 91.04 $\pm$ 0.06 & 68.38 $\pm$ 0.26 \\

Kron-OBD 
& 72.88 $\pm$ 0.12 & 67.11 $\pm$ 0.21 & 28.57 $\pm$ 0.19
& 71.16 $\pm$ 0.11 & 85.83 $\pm$ 0.10 & 47.19 $\pm$ 0.35
& 67.70 $\pm$ 0.32 & 92.86 $\pm$ 0.05 & 65.26 $\pm$ 0.26 \\

Kron-OBS 
& 72.89 $\pm$ 0.12 & 67.26 $\pm$ 0.08 & 25.80 $\pm$ 0.16
& 71.36 $\pm$ 0.17 & 84.75 $\pm$ 0.02 & 45.74 $\pm$ 0.17
& 68.17 $\pm$ 0.34 & 92.16 $\pm$ 0.03 & 63.95 $\pm$ 0.19 \\

EigenDamage 
& \textbf{73.39 $\pm$ 0.12} & 66.05 $\pm$ 0.11 & 28.55 $\pm$ 0.11
& \textbf{71.62 $\pm$ 0.14} & 85.69 $\pm$ 0.05 & 54.83 $\pm$ 0.79
& \textbf{69.50 $\pm$ 0.22} & 92.92 $\pm$ 0.03 & 74.55 $\pm$ 0.33 \\

\hline
VGG19+$L_1$ (Baseline) 
& 73.08 & - & -
& -     & - & -
& -     & - & - \\

NN Slimming~\citep{liu2017learning} 
& 73.24 $\pm$ - & 72.68 $\pm$ - & 35.37 $\pm$ -
& 71.55 $\pm$ - & 84.38 $\pm$ - & 51.59 $\pm$ -
& 66.55 $\pm$ - & 92.48 $\pm$ - & 76.54 $\pm$ - \\

C-OBD%
& 73.39 $\pm$ 0.05 & 74.16 $\pm$ 0.01 & 35.13 $\pm$ 0.01
& 71.40 $\pm$ 0.09 & 86.13 $\pm$ 0.01 & 45.47 $\pm$ 0.10
& 67.56 $\pm$ 0.16 & 93.00 $\pm$ 0.01 & 63.42 $\pm$ 0.11 \\

C-OBS%
& \textbf{73.44 $\pm$ 0.04} & 71.17 $\pm$ 0.03 & 33.77 $\pm$ 0.67
& 71.30 $\pm$ 0.12 & 84.07 $\pm$ 0.01 & 56.74 $\pm$ 0.13
& 66.90 $\pm$ 0.23 & 91.20 $\pm$ 0.04 & 73.39 $\pm$ 0.31 \\

Kron-OBD 
& 73.24 $\pm$ 0.05 & 74.00 $\pm$ 0.03 & 36.56 $\pm$ 0.03
& 71.01 $\pm$ 0.13 & 86.66 $\pm$ 0.05 & 52.66 $\pm$ 0.21
& 67.24 $\pm$ 0.20 & 92.90 $\pm$ 0.05 & 68.62 $\pm$ 0.21 \\

Kron-OBS 
& 73.20 $\pm$ 0.12 & 72.27 $\pm$ 0.03 & 36.45 $\pm$ 0.66
& \textbf{71.88 $\pm$ 0.11} & 84.77 $\pm$ 0.01 & 50.53 $\pm$ 0.08
& 67.75 $\pm$ 0.14 & 92.08 $\pm$ 0.01 & 67.39 $\pm$ 0.17 \\

EigenDamage 
& 73.23 $\pm$ 0.08 & 66.80 $\pm$ 0.02 & 29.49 $\pm$ 0.03
& 71.81 $\pm$ 0.13 & 84.27 $\pm$ 0.04 & 52.75 $\pm$ 0.21
& \textbf{69.83 $\pm$ 0.24} & 92.36 $\pm$ 0.01 & 73.68 $\pm$ 0.13 \\

\bottomrule
\end{tabular}
}
\end{center}
\vspace{-0.3cm}
\end{table*}

\begin{table*}[h]
\small
\vspace{-0.5cm}
\caption{One pass pruning on CIFAR-10 with ResNet}
\vspace{-0.3cm}
\label{tab:classification}
\begin{center}
\resizebox{\textwidth}{!}{
\begin{tabular}{l|c c c|c c c|c c c}
\toprule
Prune Ratio (\%)  
& \multicolumn{3}{c|}{50\%} 
& \multicolumn{3}{c|}{70\%}
& \multicolumn{3}{c}{80\%} \\
\hline
\multirow{2}{*}{Method }
 & Test & Reduction in & Reduction in & Test & Reduction in & Reduction in & Test & Reduction in & Reduction in \\
& acc (\%) & weights (\%) & FLOPs (\%)       
& acc (\%) & weights (\%) & FLOPs (\%)  
& acc (\%) & weights (\%) & FLOPs (\%)      \\
\hline
ResNet32(Baseline) 
& 95.30 & - & -
& -     & - & -
& -     & - & -\\

C-OBD%
& 95.27 $\pm$ 0.10 & 60.67 $\pm$ 0.44 & 55.46 $\pm$ 0.35 
& \textbf{95.00 $\pm$ 0.17} & 80.64 $\pm$ 0.40 & 76.78 $\pm$ 0.63
& 94.41 $\pm$ 0.10 & 90.73 $\pm$ 0.11 & 86.66 $\pm$ 0.34 \\

C-OBS%
& 95.30 $\pm$ 0.15 & 58.99 $\pm$ 0.02 & 65.54 $\pm$ 0.25
& 94.43 $\pm$ 0.17 & 76.27 $\pm$ 0.35 & 85.89 $\pm$ 0.23
& 93.45 $\pm$ 0.25 & 86.15 $\pm$ 0.68 & 92.77 $\pm$ 0.05 \\

Kron-OBD 
& 95.30 $\pm$ 0.09 & 56.05 $\pm$ 0.24 & 52.21 $\pm$ 0.36
& 94.94 $\pm$ 0.02 & 73.98 $\pm$ 0.46 & 74.97 $\pm$ 0.63
& \textbf{94.60 $\pm$ 0.14} & 85.96 $\pm$ 0.41 & 86.36 $\pm$ 0.38 \\

Kron-OBS 
& \textbf{95.46 $\pm$ 0.08} & 56.48 $\pm$ 0.26 & 50.93 $\pm$ 0.46
& 94.92 $\pm$ 0.11 & 73.77 $\pm$ 0.24 & 74.58 $\pm$ 0.44
& 94.44 $\pm$ 0.08 & 85.65 $\pm$ 0.46 & 86.05 $\pm$ 0.55 \\

EigenDamage 
& 95.28 $\pm$ 0.16 & 59.68 $\pm$ 0.28 & 58.32 $\pm$ 0.23
& 94.86 $\pm$ 0.11 & 82.57 $\pm$ 0.27 & 80.88 $\pm$ 0.36
& 94.23 $\pm$ 0.13 & 90.48 $\pm$ 0.35 & 88.86 $\pm$ 0.50 \\

\hline
PreResNet29+$L_1$ (Baseline) 
& 94.42 & -  & -
& -     & -  & -
& -     & -  & -\\

NN Slimming~\citep{liu2017learning} 
& 92.60 $\pm$ - &  58.12 $\pm$ - & 69.88 $\pm$ -
& 90.60 $\pm$ - & 81.99  $\pm$ - & 87.05 $\pm$ -
& 87.24 $\pm$ - & 86.68  $\pm$ - & 91.33 $\pm$ - \\

C-OBD%
& 92.85 $\pm$ 0.14 & 79.67 $\pm$ 0.41 & 68.90 $\pm$ 0.49
& 88.74 $\pm$ 0.32 & 93.22 $\pm$ 0.08 & 86.37 $\pm$ 0.32
& 85.75 $\pm$ 0.66 & 96.14 $\pm$ 0.04 & 91.39 $\pm$ 0.16 \\

C-OBS%
& 92.43 $\pm$ 0.03 & 73.81 $\pm$ 0.14 & 76.33 $\pm$ 0.21
& 85.85 $\pm$ 0.36 & 91.74 $\pm$ 0.17 & 92.39 $\pm$ 0.07
& 78.97 $\pm$ 0.84 & 96.55 $\pm$ 0.02 & 96.60 $\pm$ 0.06 \\

Kron-OBD 
& 93.19 $\pm$ 0.06 & 59.72 $\pm$ 0.31 & 54.26 $\pm$ 0.34
& 87.99 $\pm$ 0.37 & 86.50 $\pm$ 0.05 & 78.96 $\pm$ 0.17
& 86.40 $\pm$ 0.28 & 95.02 $\pm$ 0.04 & 88.58 $\pm$ 0.04 \\

Kron-OBS 
& 92.88 $\pm$ 0.08 & 58.95 $\pm$ 0.14 & 57.61 $\pm$ 0.12
& 87.71 $\pm$ 0.22 & 85.38 $\pm$ 0.03 & 82.00 $\pm$ 0.10
& 85.67 $\pm$ 0.31 & 93.93 $\pm$ 0.07 & 90.74 $\pm$ 0.31 \\

EigenDamage 
& \textbf{94.15 $\pm$ 0.07} & 62.06 $\pm$ 0.15 & 54.40 $\pm$ 0.10 
& \textbf{93.33 $\pm$ 0.07} & 77.71 $\pm$ 0.11 & 71.92 $\pm$ 0.15
& \textbf{92.30 $\pm$ 0.15} & 86.27 $\pm$ 0.04 & 81.59 $\pm$ 0.07 \\

\bottomrule
\end{tabular}
}
\end{center}
\vspace{-0.5cm}
\end{table*}
\begin{table*}[h]
\small
\vspace{-0.5cm}
\caption{One pass pruning on CIFAR-100 with ResNet}
\vspace{-0.3cm}
\begin{center}
\resizebox{\textwidth}{!}{
\begin{tabular}{l|c c c|c c c|c c c}
\toprule
Prune Ratio (\%)  
& \multicolumn{3}{c|}{50\%} 
& \multicolumn{3}{c|}{70\%}
& \multicolumn{3}{c}{80\%} \\
\hline
\multirow{2}{*}{Method }
 & Test & Reduction in & Reduction in & Test & Reduction in & Reduction in & Test & Reduction in & Reduction in \\
& acc (\%) & weights (\%) & FLOPs (\%)       
& acc (\%) & weights (\%) & FLOPs (\%)  
& acc (\%) & weights (\%) & FLOPs (\%)    \\
\hline
ResNet32(Baseline) 
& 78.17 & - & -
& -     & - & -
& -     & - & - \\

C-OBD%
& 76.59 $\pm$ 0.06 & 57.82 $\pm$ 0.48 & 56.88 $\pm$ 0.37
& 73.74 $\pm$ 0.42 & 77.08 $\pm$ 0.32 & 78.64 $\pm$ 0.32
& 68.86 $\pm$ 0.40 & 89.67 $\pm$ 0.19 & 88.45 $\pm$ 0.19 \\

C-OBS%
& 76.26 $\pm$ 0.25 & 58.47 $\pm$ 0.22 & 63.81 $\pm$ 0.26
& 73.06 $\pm$ 0.23 & 74.33 $\pm$ 0.15 & 86.06 $\pm$ 0.03
& 63.15 $\pm$ 0.41 & 80.61 $\pm$ 0.15 & 91.10 $\pm$ 0.06 \\

Kron-OBD 
& 76.41  $\pm$ 0.29 & 53.08 $\pm$ 0.35 & 52.06 $\pm$ 0.27
& 72.82 $\pm$ 0.28 & 73.40 $\pm$ 0.11 & 75.25 $\pm$ 0.15
& 69.62 $\pm$ 0.38 & 84.50 $\pm$ 0.16 & 88.04 $\pm$ 0.10 \\

Kron-OBS 
& \textbf{76.74 $\pm$ 0.32} & 52.21 $\pm$ 0.20  & 49.40 $\pm$ 0.13
& 73.00 $\pm$ 0.19 & 71.60 $\pm$ 0.15 & 73.33 $\pm$ 0.48
& 70.42 $\pm$ 0.20 & 80.34 $\pm$ 0.25 & 86.96 $\pm$ 0.27 \\

EigenDamage 
& 76.12 $\pm$ 0.12 & 61.73 $\pm$ 0.12 & 60.87 $\pm$ 0.38
& \textbf{73.82 $\pm$ 0.07} & 79.85 $\pm$ 0.07 & 81.03 $\pm$ 0.11
& \textbf{70.78 $\pm$ 0.08} & 88.68 $\pm$ 0.08 & 88.68 $\pm$ 0.06 \\

\hline
PreResNet29+$L_1$ (Baseline) 
& 75.70 & - & -
& -     & - & -
& -     & - & - \\

NN Slimming~\citep{liu2017learning} 
& 71.93 $\pm$ - & 47.55 $\pm$ - & 73.44 $\pm$ -
& 63.47 $\pm$ - & 78.41 $\pm$ - & 89.92 $\pm$ -
& 61.64 $\pm$ - & 85.47 $\pm$ - & 92.38 $\pm$ -\\

C-OBD%
& 67.18 $\pm$ 0.10 & 84.72 $\pm$ 0.05 & 75.48 $\pm$ 0.11
& 55.02 $\pm$ 0.50 & 93.42 $\pm$ 0.01 & 88.12 $\pm$ 0.06
& 47.87 $\pm$ 0.57 & 96.57 $\pm$ 0.01 & 93.48 $\pm$ 0.02 \\

C-OBS%
& 71.17 $\pm$ 0.11 & 65.67 $\pm$ 0.12 & 80.77 $\pm$ 0.05
& 51.45 $\pm$ 0.74 & 91.88 $\pm$ 0.06 & 95.00 $\pm$ 0.03
& 41.55 $\pm$ 0.91 & 95.64 $\pm$ 0.02 & 97.20 $\pm$ 0.01 \\

Kron-OBD 
& 69.64 $\pm$ 0.19 & 56.63 $\pm$ 0.20 & 58.24 $\pm$ 0.05
& 46.46 $\pm$ 0.72 & 89.50 $\pm$ 0.07 & 82.44 $\pm$ 0.05
& 41.64 $\pm$ 0.86 & 95.83 $\pm$ 0.01 & 89.63 $\pm$ 0.06 \\

Kron-OBS 
& 69.87 $\pm$ 0.17 & 49.09 $\pm$ 0.13 & 62.59 $\pm$ 0.05
& 49.00 $\pm$ 0.68 & 87.03 $\pm$ 0.05 & 87.82 $\pm$ 0.10
& 40.51 $\pm$ 1.04 & 94.72 $\pm$ 0.01 & 93.86 $\pm$ 0.01\\

EigenDamage 
& \textbf{74.50 $\pm$ 0.13} & 57.98 $\pm$ 0.15 & 53.87 $\pm$ 0.14
& \textbf{72.41 $\pm$ 0.16} & 75.79 $\pm$ 0.04 & 71.92 $\pm$ 0.12
& \textbf{70.09 $\pm$ 0.11} & 85.34 $\pm$ 0.02 & 81.61 $\pm$ 0.06\\
\bottomrule
\end{tabular}
}
\end{center}
\vspace{-0.3cm}
\end{table*}
\begin{table*}[h]
\small
\vspace{-0.5cm}
\caption{One pass pruning on Tiny-ImageNet with VGG19. N/A denotes the network failed to converge, and achieves random guess performance on the test set.}
\vspace{-0.3cm}
\begin{center}
\resizebox{\textwidth}{!}{
\begin{tabular}{l|c c c|c c c|c c c|c c c}
\toprule

Prune Ratio (\%)  
& \multicolumn{3}{c|}{40\%} 
& \multicolumn{3}{c|}{60\%}
& \multicolumn{3}{c|}{70\%} 
& \multicolumn{3}{c}{80\%} 
\\
\hline
\multirow{2}{*}{Method }
& Test & Reduction in & Reduction in & Test & Reduction in & Reduction in & Test & Reduction in & Reduction in & Test & Reduction in & Reduction in \\
& acc (\%) & weights (\%) & FLOPs (\%)
& acc (\%) & weights (\%) & FLOPs (\%)       
& acc (\%) & weights (\%) & FLOPs (\%)  
& acc (\%) & weights (\%) & FLOPs (\%)     \\
\hline
VGG19(Baseline) 
& 61.56 & - & -
& -     & - & -
& -     & - & -
& -     & - & - \\
NN Slimming ~\cite{liu2017learning} 
& 57.40 $\pm$ - & 52.61 $\pm$ - & 76.99 $\pm$ -
& 40.05 $\pm$ - & 71.04 $\pm$ - & 90.04 $\pm$ -
& N/A $\pm$ - & 84.63 $\pm$ - & 94.09 $\pm$ -
& N/A $\pm$ - & 93.02 $\pm$ - & 95.45 $\pm$ -
\\

C-OBD%
& 56.87 $\pm$ 0.13 & 58.95 $\pm$ 0.06 & 55.49 $\pm$ 0.50
& 47.36 $\pm$ 0.47 & 79.10 $\pm$ 0.32 & 69.74 $\pm$ 0.38
& 43.61 $\pm$ 0.31 & 88.57 $\pm$ 0.14 & 74.90 $\pm$ 0.31
& 42.29 $\pm$ 0.53 & 95.62 $\pm$ 0.13 & 78.45 $\pm$ 0.46
\\

C-OBS%
& 56.72 $\pm$ 0.21 & 46.90 $\pm$ 0.18 & 68.87 $\pm$ 0.14
& 39.80 $\pm$ 0.53 & 67.46 $\pm$ 0.34 & 86.81 $\pm$ 0.25
& 35.30 $\pm$ 0.33 & 76.47 $\pm$ 0.07 & 92.71 $\pm$ 0.07
& 31.52 $\pm$ 0.66 & 86.19 $\pm$ 0.07 & 96.66 $\pm$ 0.04
\\

Kron-OBD 
& 56.81 $\pm$ 0.22 & 56.75 $\pm$ 0.27 & 66.96 $\pm$ 0.65
& 44.41 $\pm$ 0.82 & 76.55 $\pm$ 0.15 & 81.27 $\pm$ 0.16
& 41.03 $\pm$ 0.50 & 85.28 $\pm$ 0.11 & 86.12 $\pm$ 0.05
& 38.88 $\pm$ 0.43 & 95.02 $\pm$ 0.33 & 90.98 $\pm$ 0.34 
\\

Kron-OBS 
& 56.47 $\pm$ 0.20 & 50.67 $\pm$ 0.11 & 62.28 $\pm$ 0.66
& 44.54 $\pm$ 0.43 & 73.88 $\pm$ 0.10 & 83.27 $\pm$ 0.13
& 41.44 $\pm$ 0.41 & 82.61 $\pm$ 0.41 & 87.91 $\pm$ 0.22
& 39.54 $\pm$ 0.20 & 92.77 $\pm$ 0.19 & 91.91 $\pm$ 0.33
\\

EigenDamage 
& \textbf{59.09 $\pm$ 0.05} & 48.62 $\pm$ 0.06 & 57.30 $\pm$ 0.12
& \textbf{56.92 $\pm$ 0.23} & 74.12 $\pm$ 0.15 & 74.37 $\pm$ 0.13
& \textbf{54.46 $\pm$ 0.32} & 83.77 $\pm$ 0.02 & 81.19 $\pm$ 0.16
& \textbf{51.34 $\pm$ 0.37} & 91.05 $\pm$ 0.06 & 87.82 $\pm$ 0.16 
\\

\bottomrule
\end{tabular}
}
\end{center}
\vspace{-0.3cm}
\end{table*}

\newpage
\begin{figure*}[t]
     \centering
     \begin{subfigure}[b]{0.245\textwidth}
         \centering
         \includegraphics[width=\textwidth]{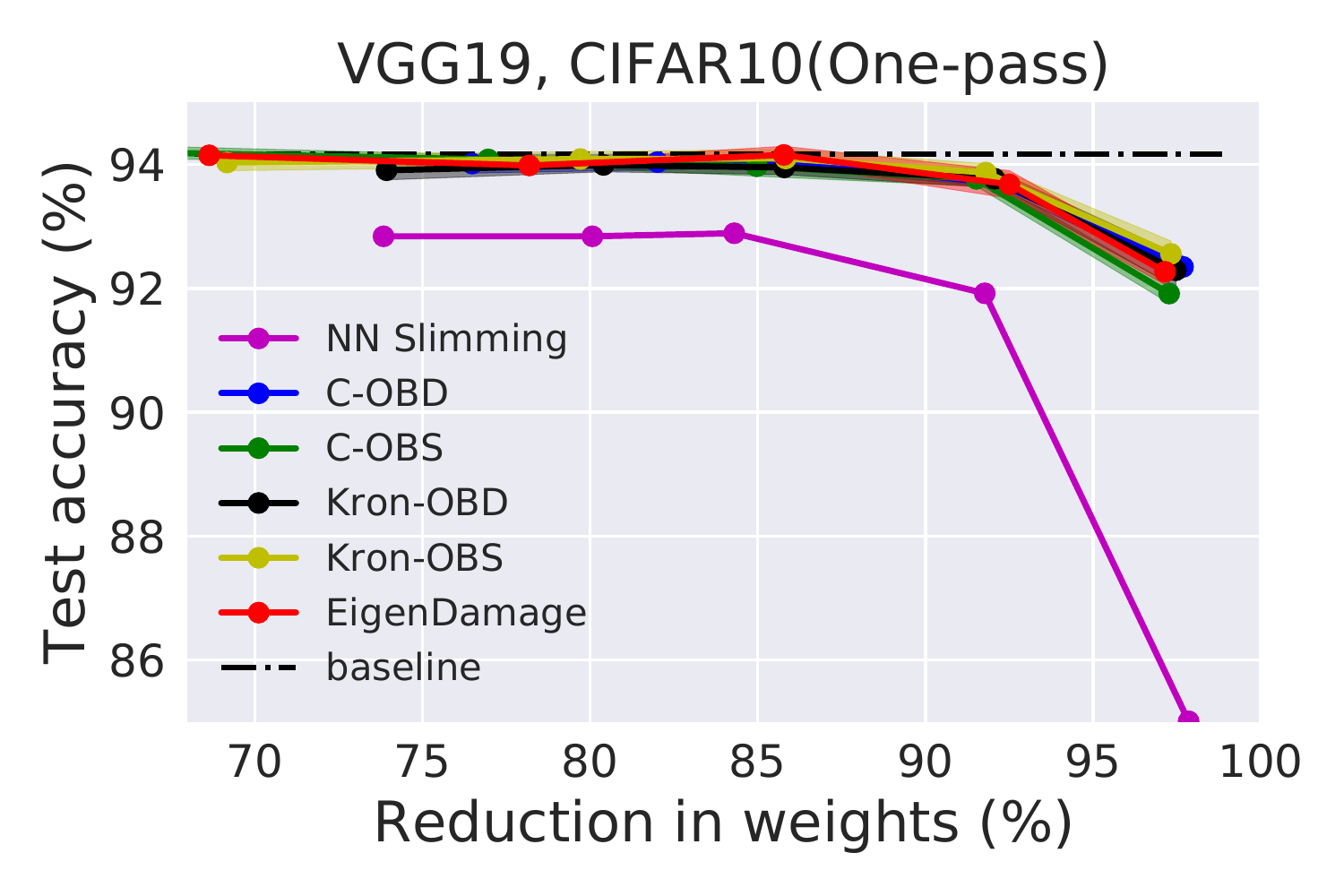}
     \end{subfigure}
     \hfill
     \begin{subfigure}[b]{0.245\textwidth}
         \centering
         \includegraphics[width=\textwidth]{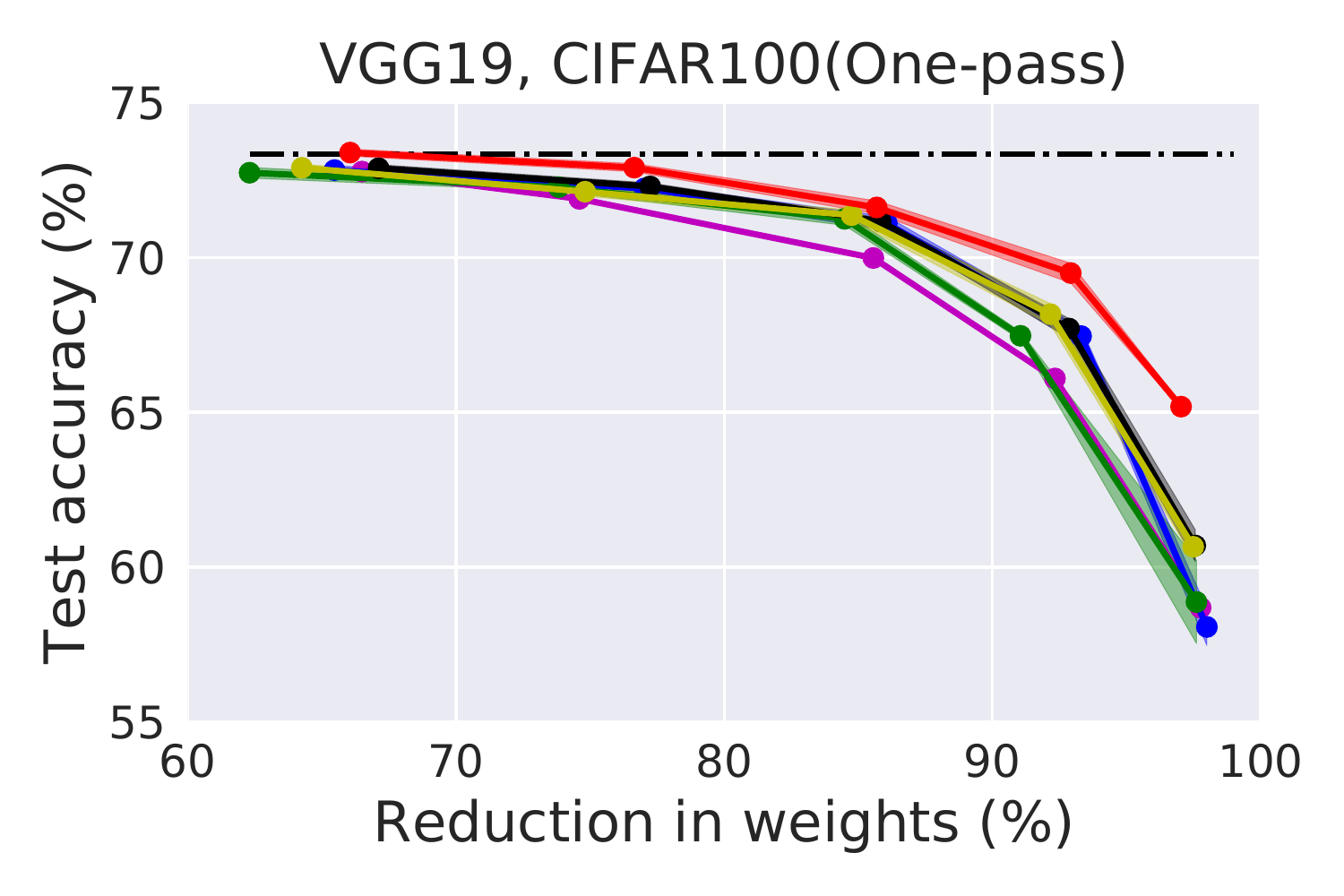}
     \end{subfigure}
     \hfill
     \begin{subfigure}[b]{0.245\textwidth}
         \centering
         \includegraphics[width=\textwidth]{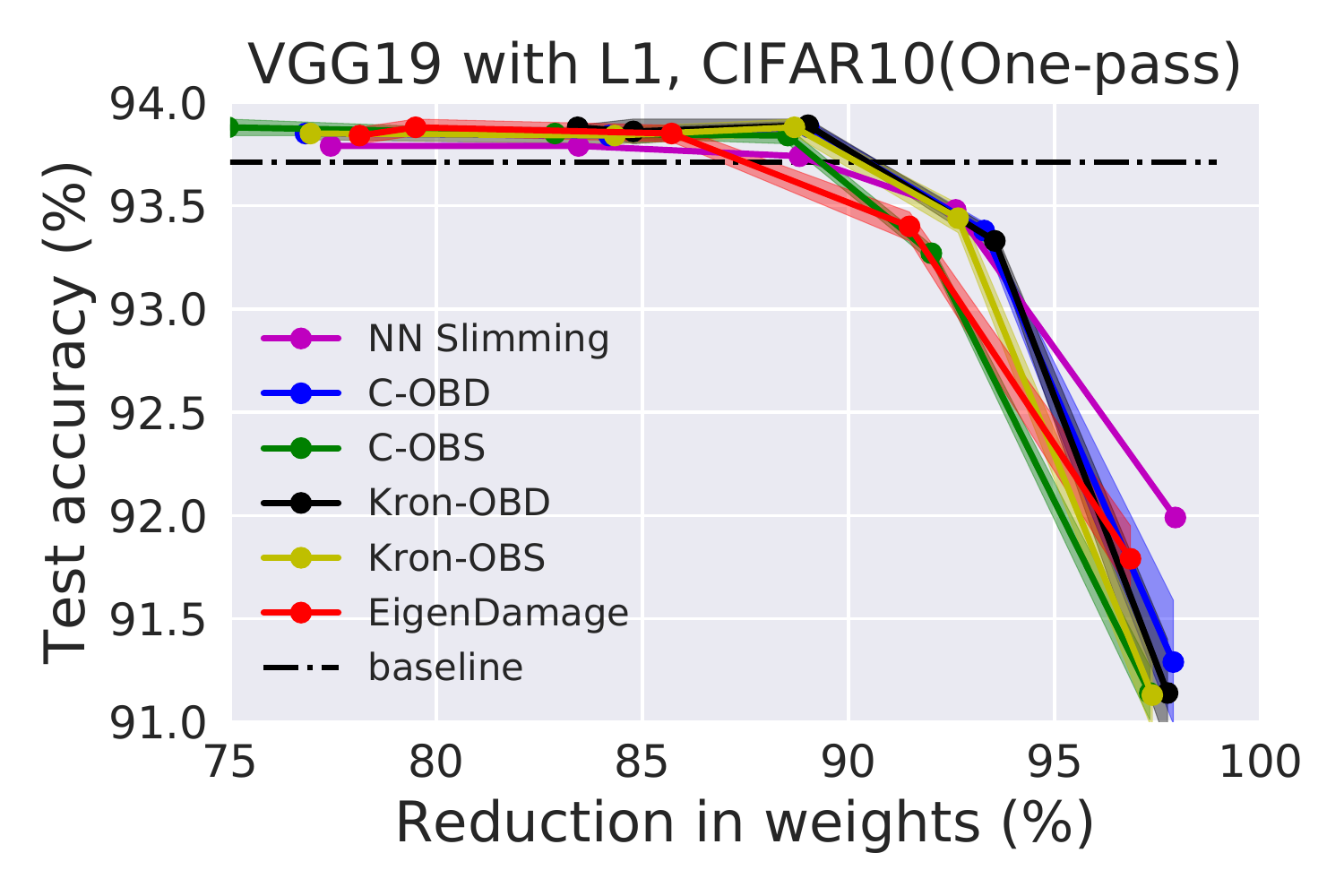}
     \end{subfigure}
     \hfill
     \begin{subfigure}[b]{0.245\textwidth}
         \centering
         \includegraphics[width=\textwidth]{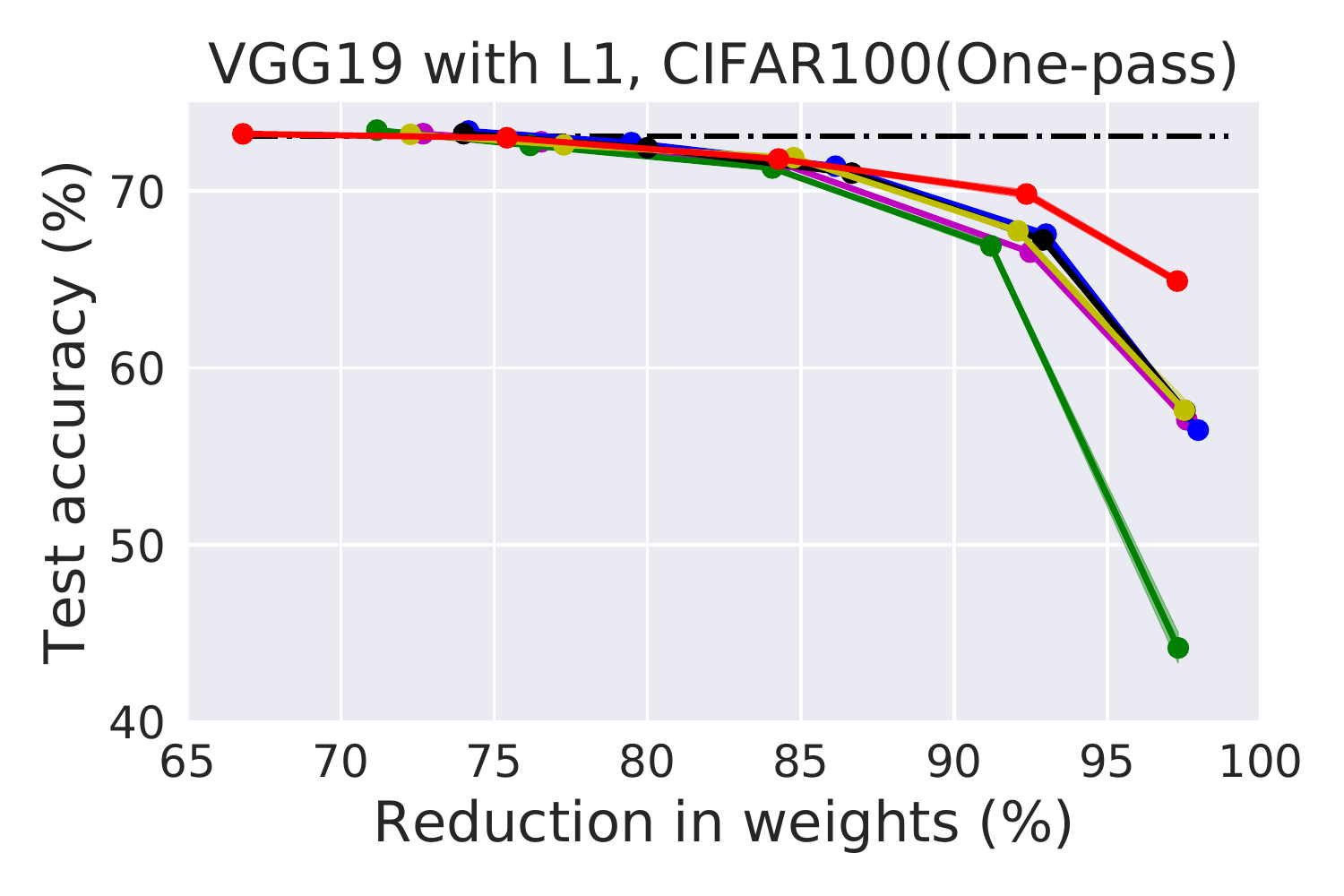}
     \end{subfigure}
     \begin{subfigure}[b]{0.245\textwidth}
         \centering
         \includegraphics[width=\textwidth]{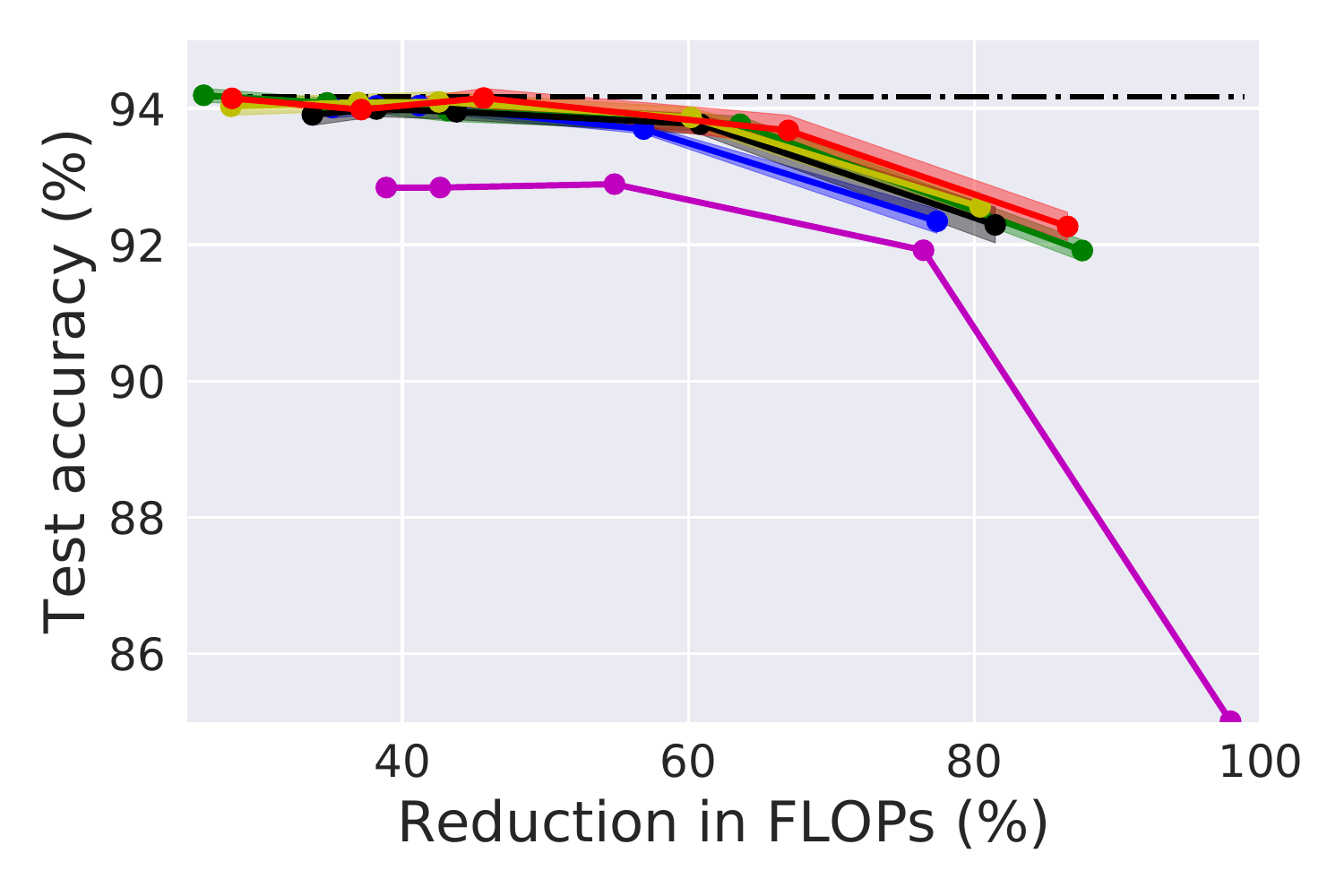}
     \end{subfigure}
     \hfill
     \begin{subfigure}[b]{0.245\textwidth}
         \centering
         \includegraphics[width=\textwidth]{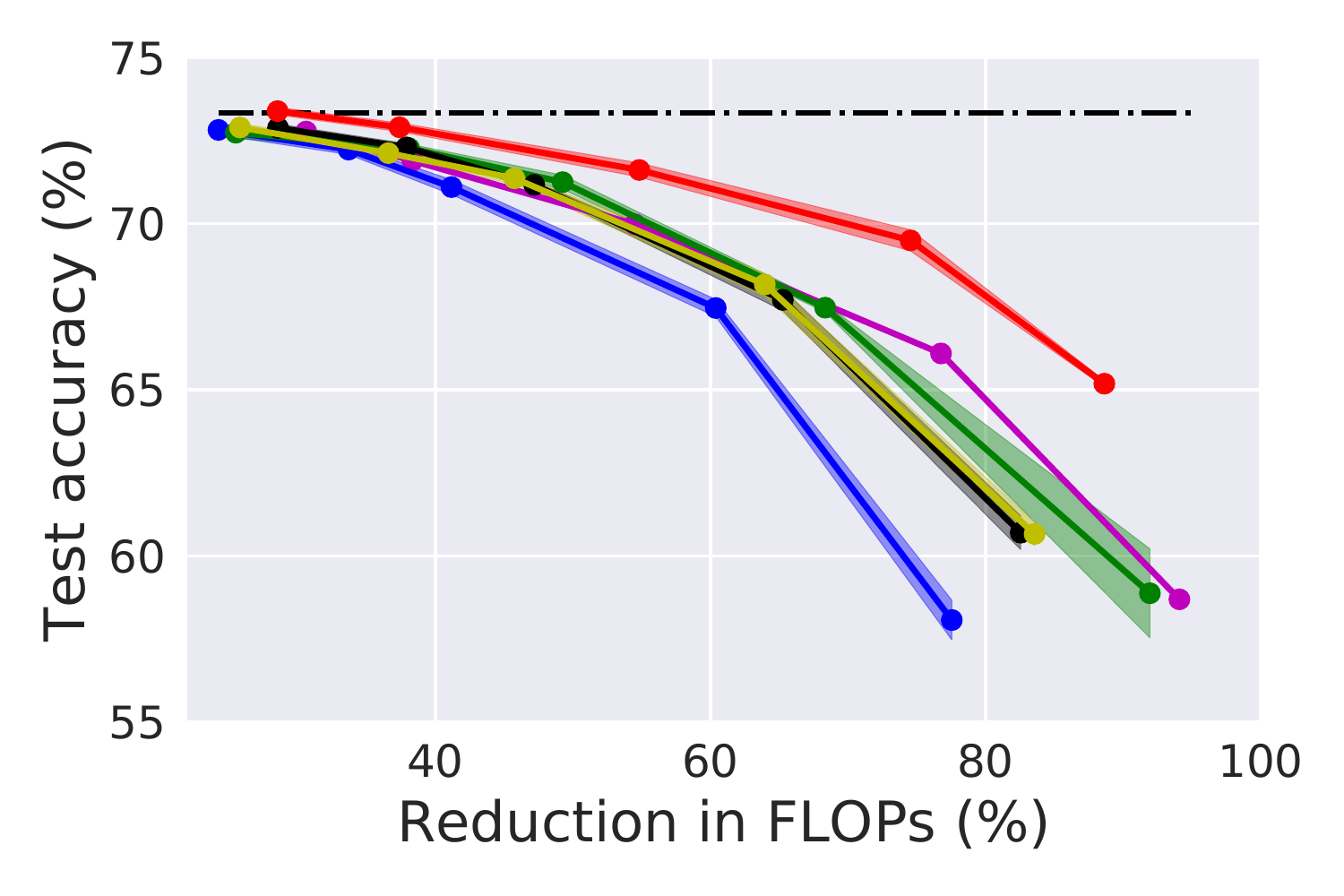}
     \end{subfigure}
     \hfill
     \begin{subfigure}[b]{0.245\textwidth}
         \centering
         \includegraphics[width=\textwidth]{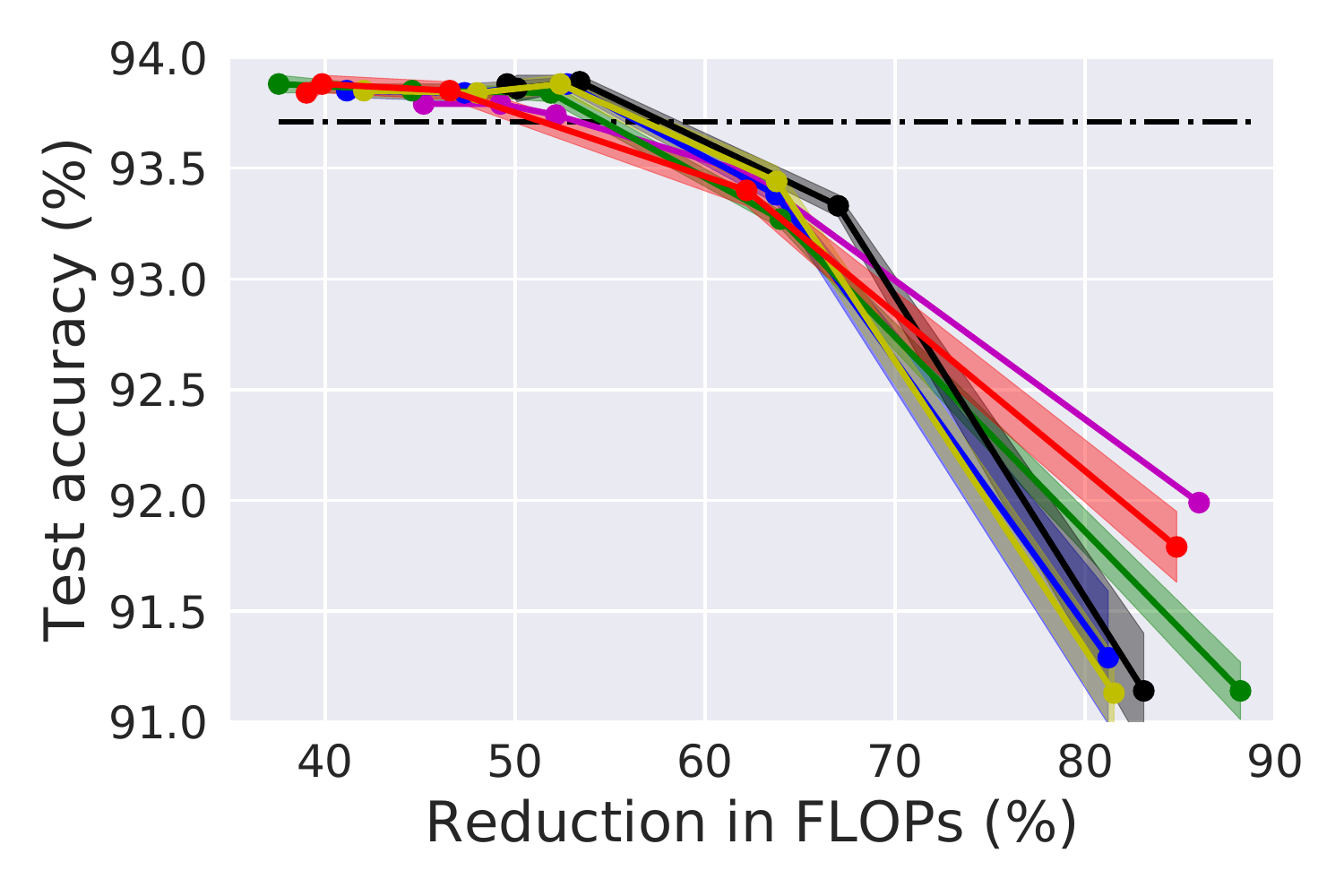}
     \end{subfigure}
     \hfill
     \begin{subfigure}[b]{0.245\textwidth}
         \centering
         \includegraphics[width=\textwidth]{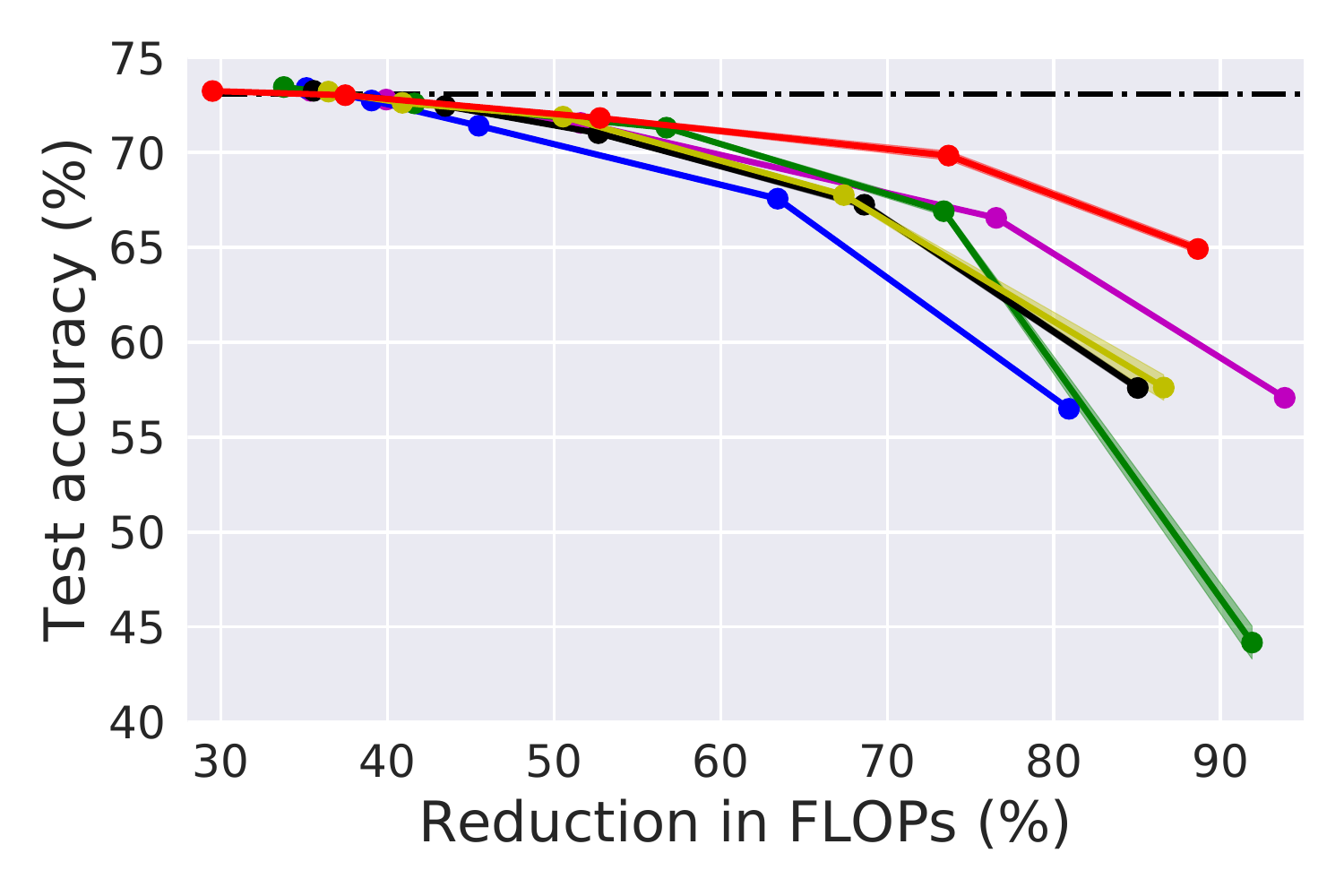}
     \end{subfigure}
     \vspace{-0.3cm}
        \caption{The results of one pass pruning, which are plotted based on the results in Tables. The first row are the curves of reduction in weights vs.~test accuracy, and second row are the curves of pruned FLOPs vs.~test accuracy of VGGNet trained on CIFAR10 and CIFAR100 dataset under the settings of with and without $L_1$ sparsity on BatchNorm. The shaded areas represent the standard variance over five runs. }
        \label{fig:onepass_figures_vgg19_cifar}
    \vspace{-0.3cm}
\end{figure*}

\begin{figure*}[t]
     \centering
     \begin{subfigure}[b]{0.245\textwidth}
         \centering
         \includegraphics[width=\textwidth]{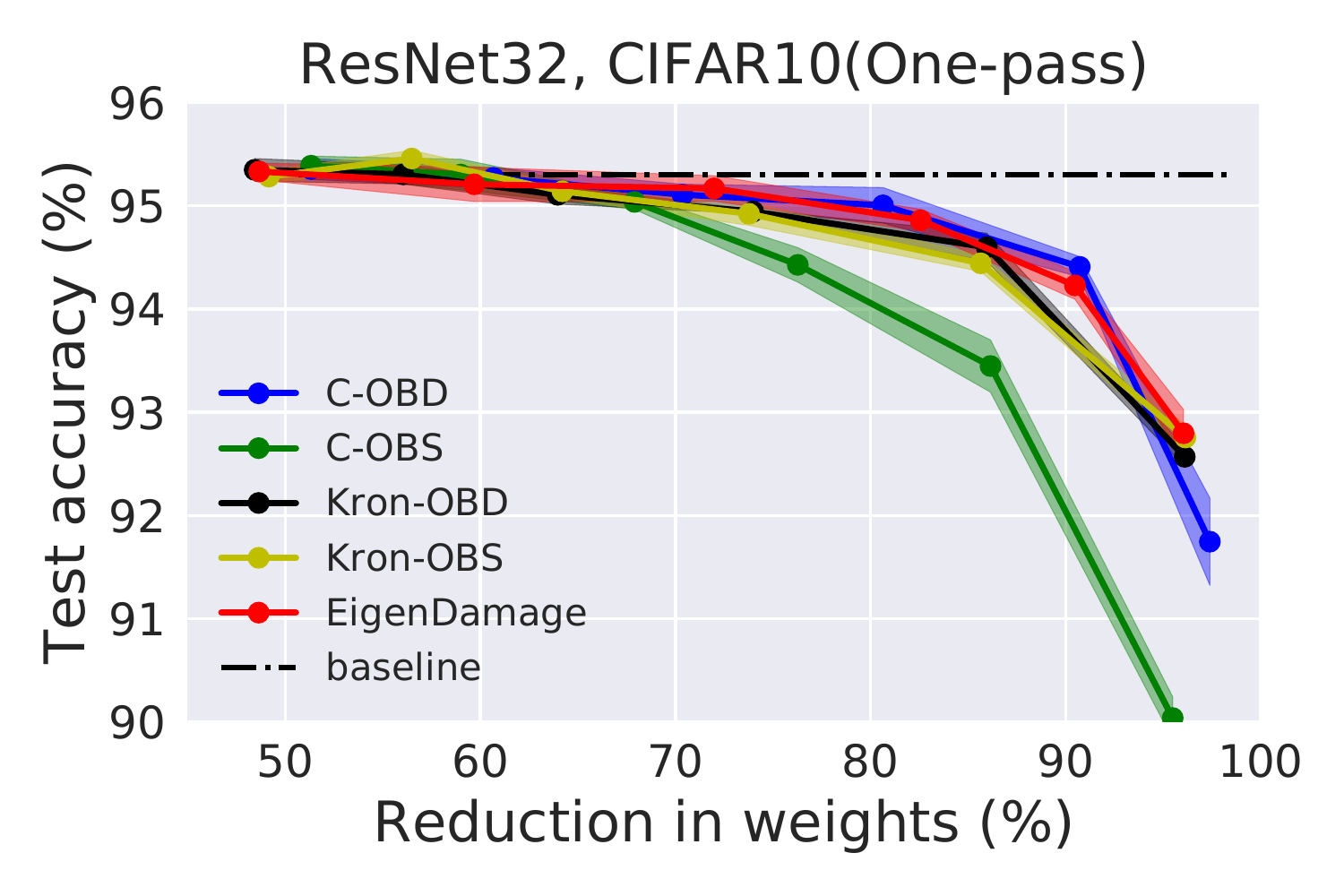}
     \end{subfigure}
     \hfill
     \begin{subfigure}[b]{0.245\textwidth}
         \centering
         \includegraphics[width=\textwidth]{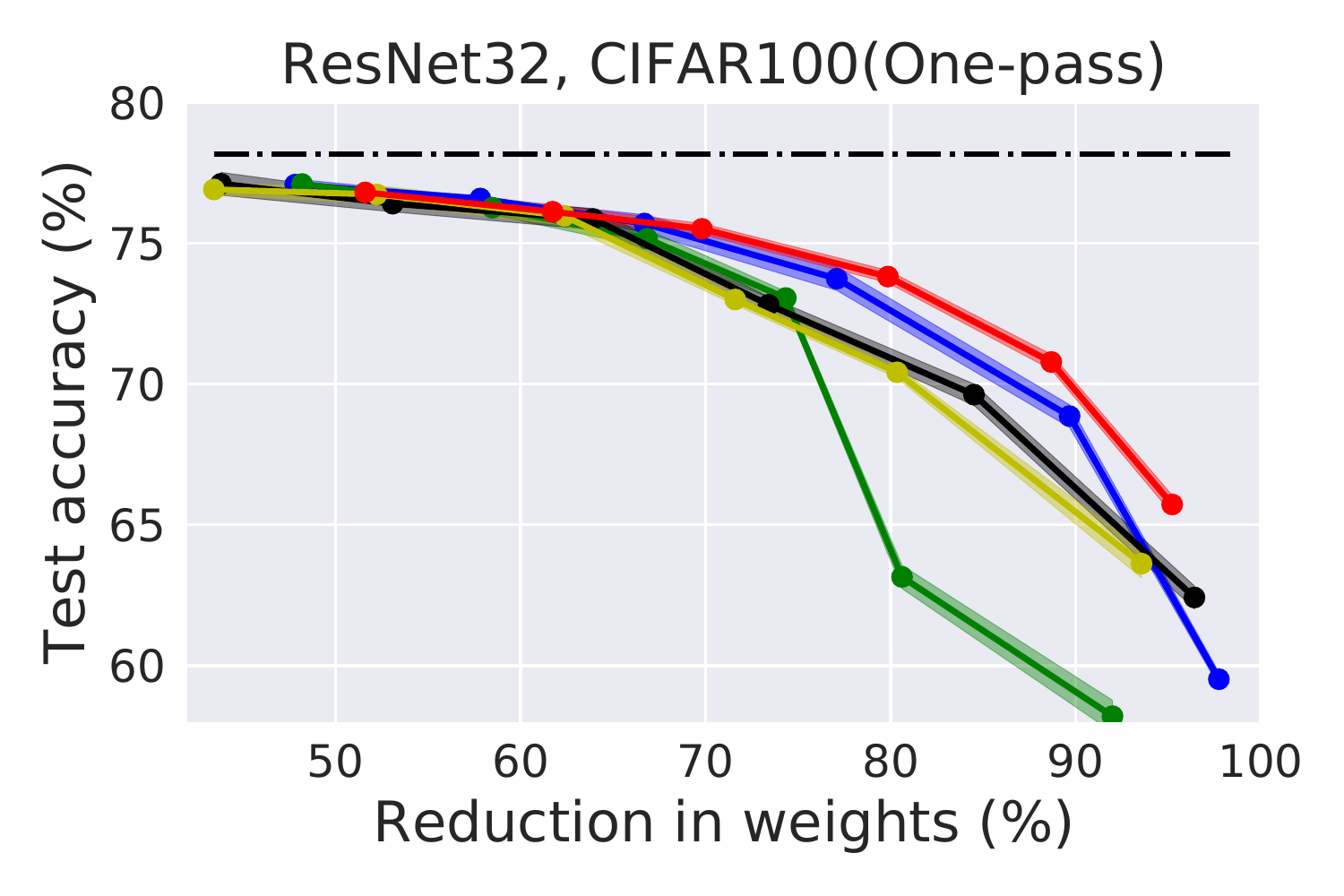}
     \end{subfigure}
     \hfill
     \begin{subfigure}[b]{0.245\textwidth}
         \centering
         \includegraphics[width=\textwidth]{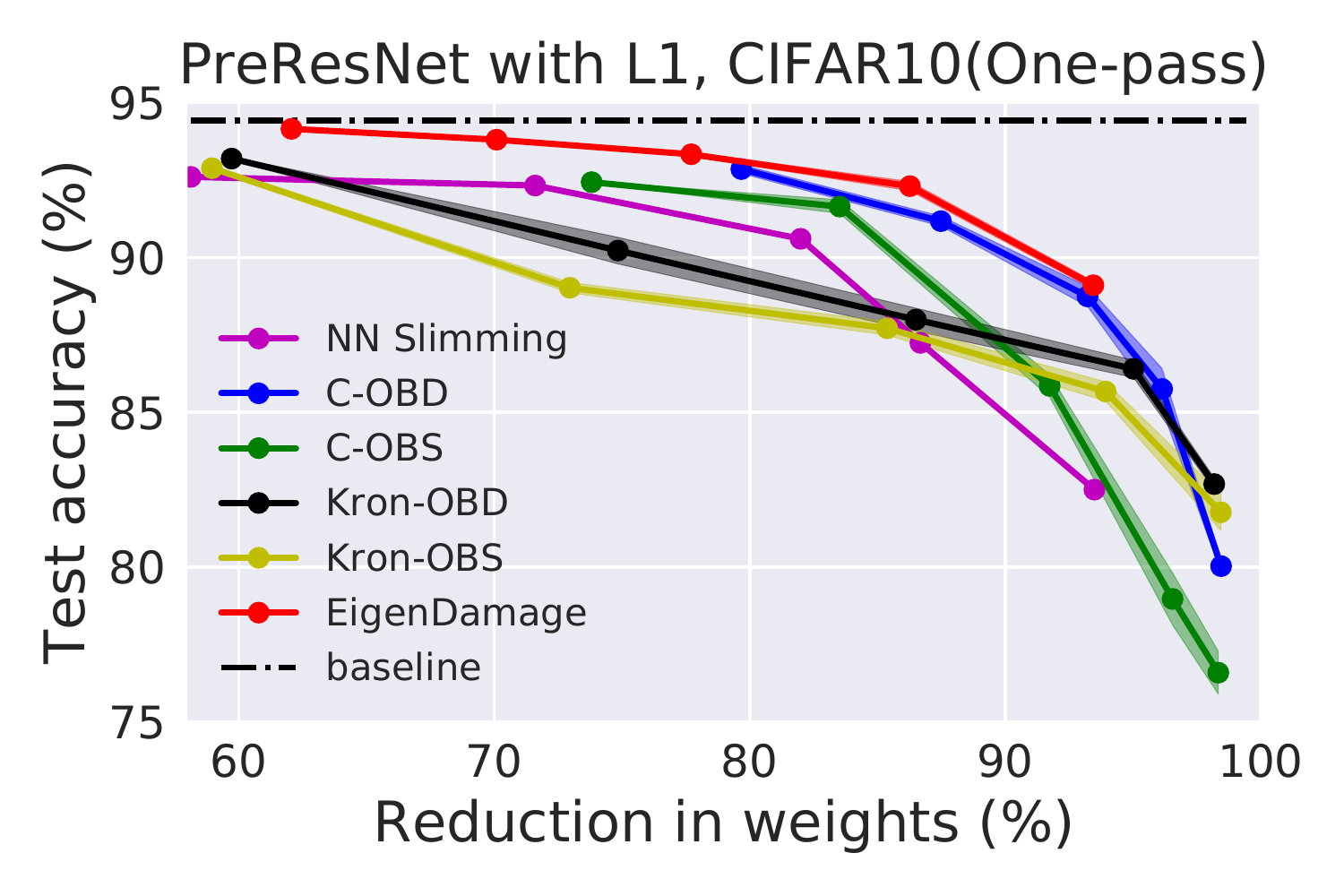}
     \end{subfigure}
     \hfill
     \begin{subfigure}[b]{0.245\textwidth}
         \centering
         \includegraphics[width=\textwidth]{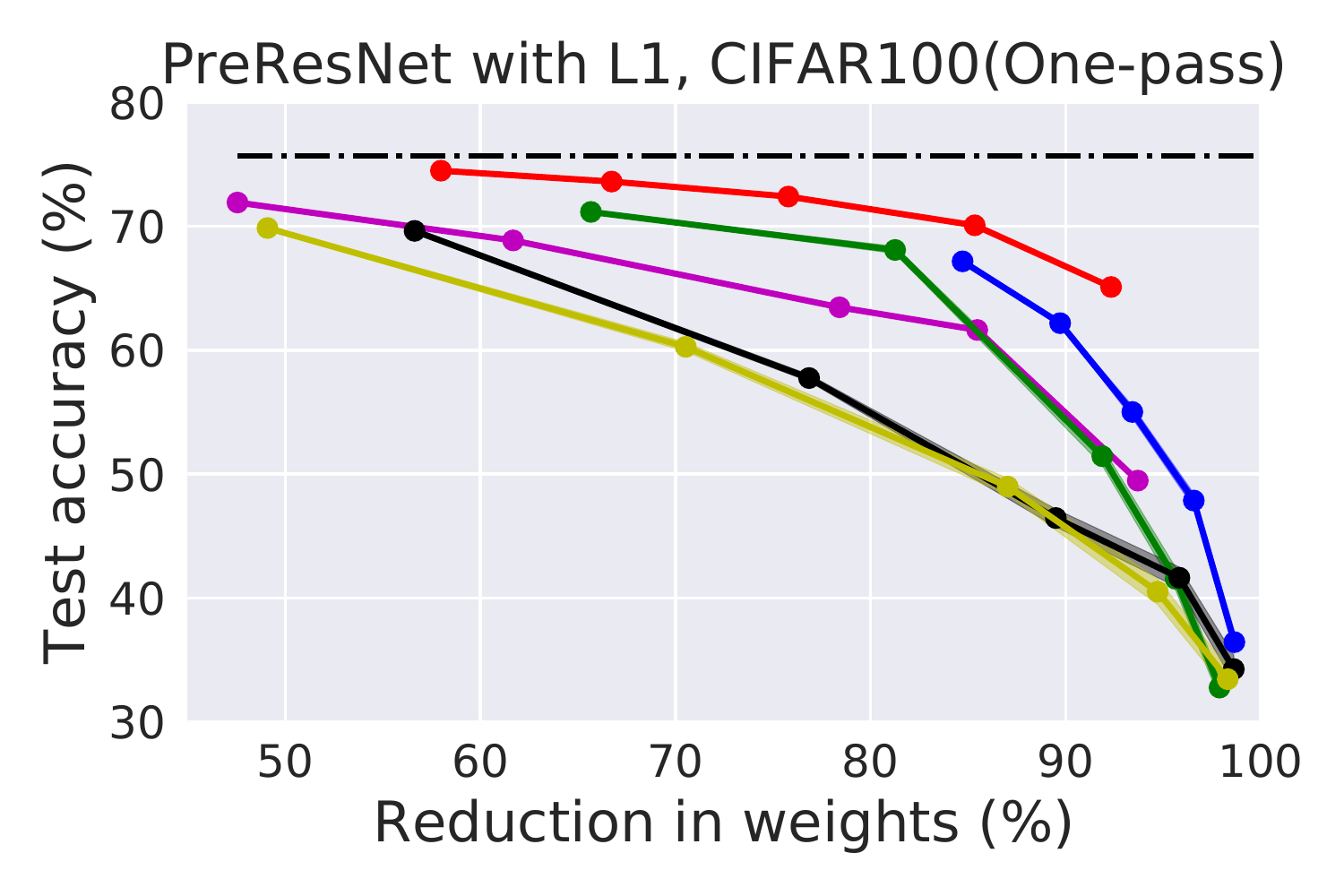}
     \end{subfigure}
     \begin{subfigure}[b]{0.245\textwidth}
         \centering
         \includegraphics[width=\textwidth]{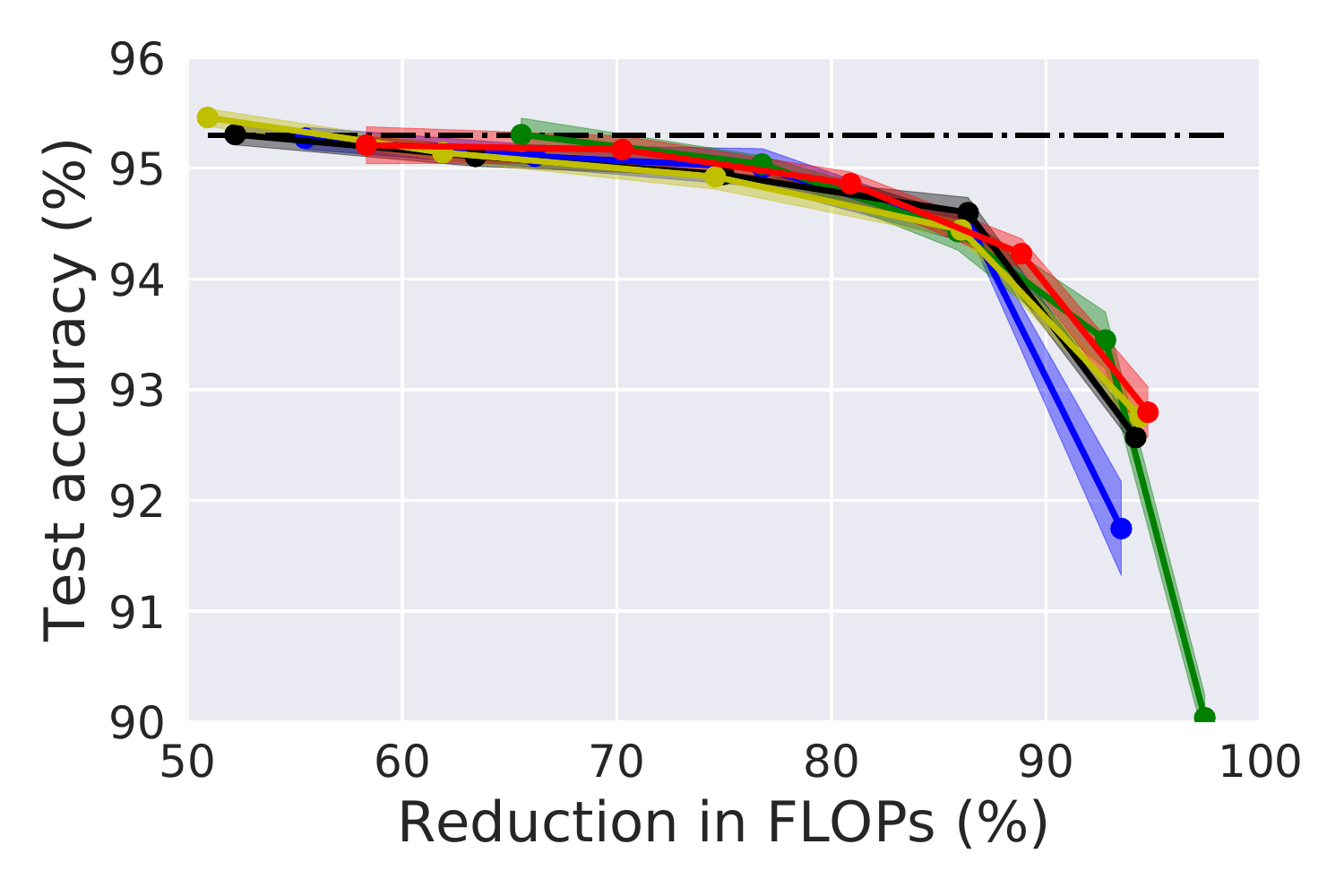}
     \end{subfigure}
     \hfill
     \begin{subfigure}[b]{0.245\textwidth}
         \centering
         \includegraphics[width=\textwidth]{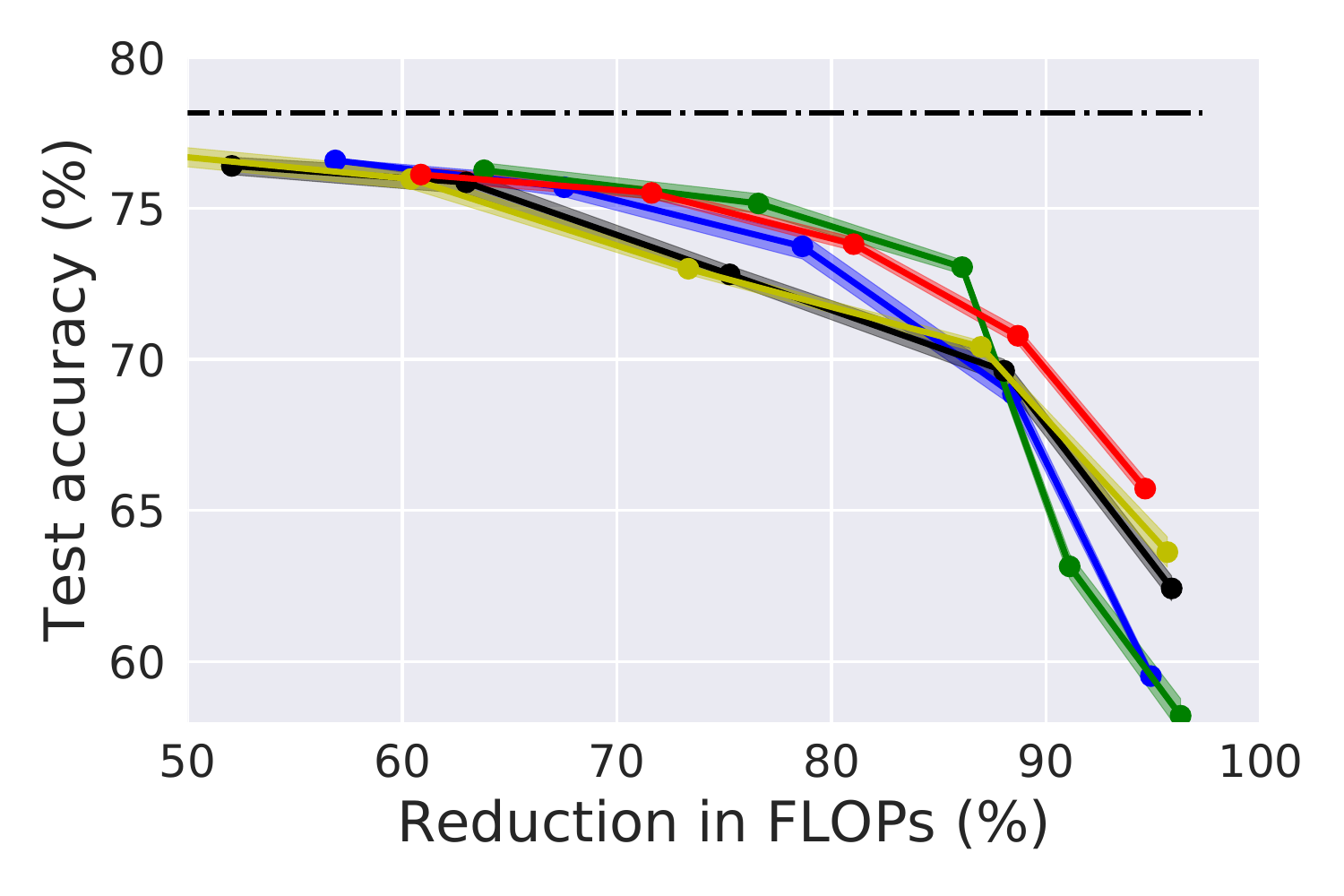}
     \end{subfigure}
     \hfill
     \begin{subfigure}[b]{0.245\textwidth}
         \centering
         \includegraphics[width=\textwidth]{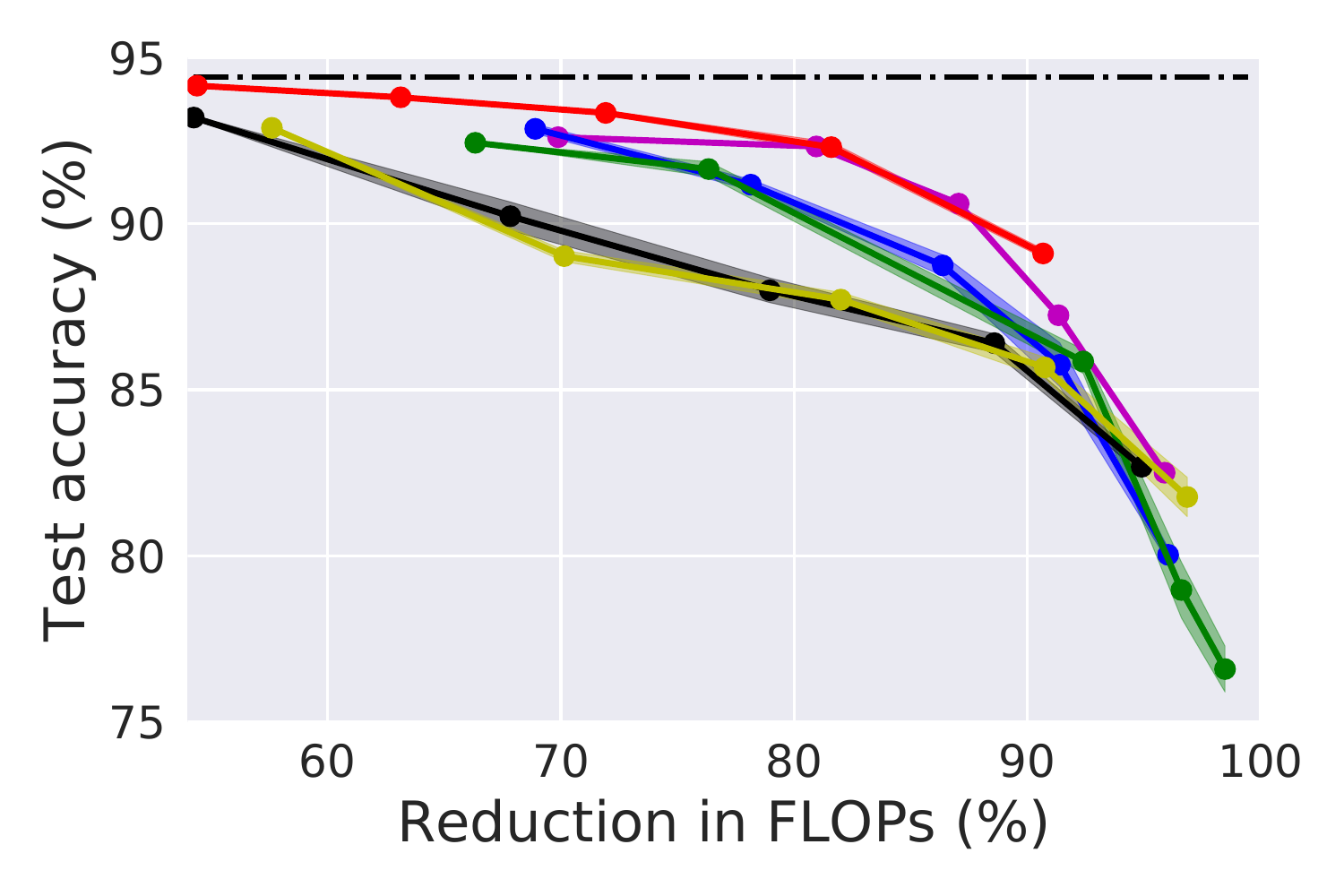}
     \end{subfigure}
     \hfill
     \begin{subfigure}[b]{0.245\textwidth}
         \centering
         \includegraphics[width=\textwidth]{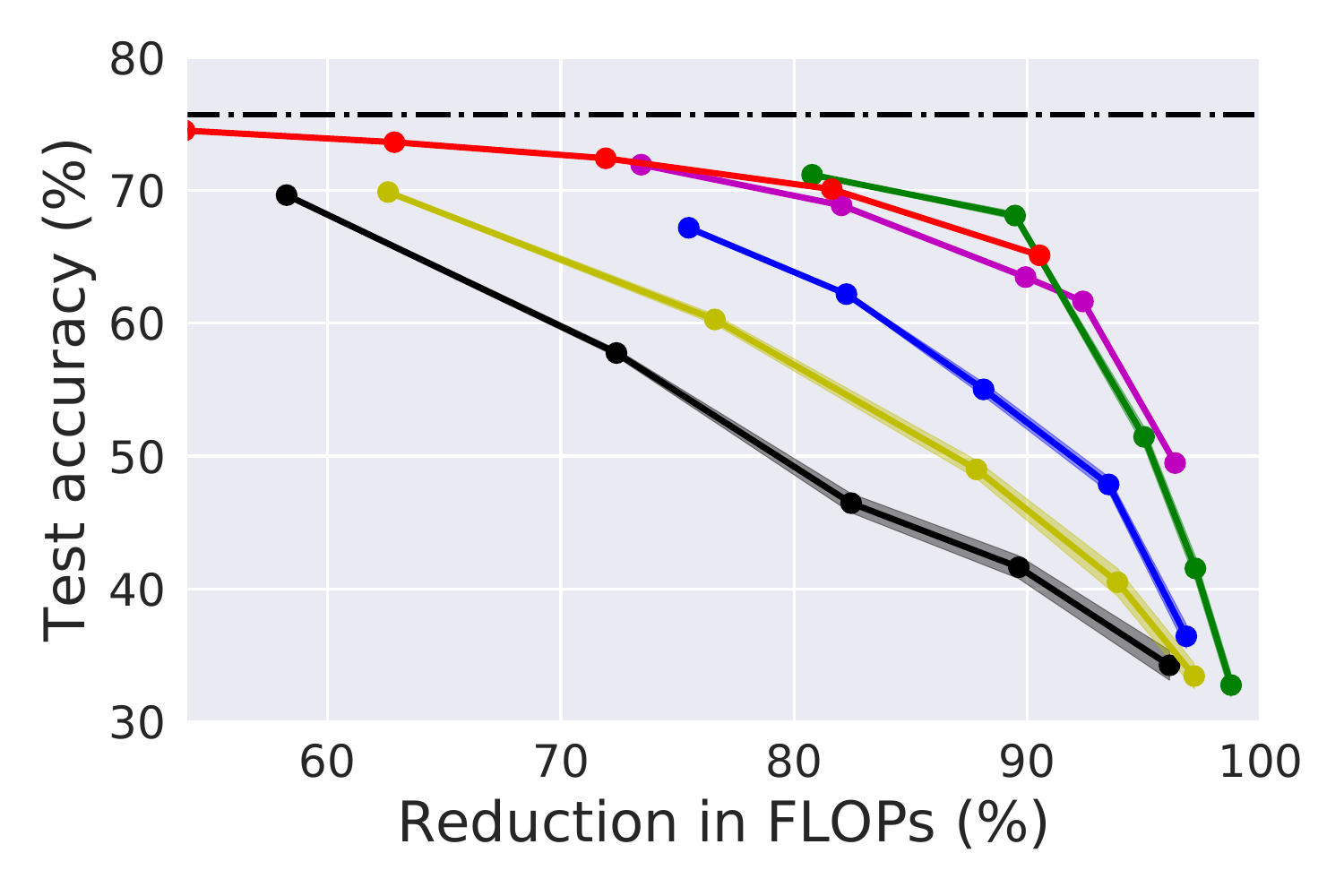}
     \end{subfigure}
     \vspace{-0.3cm}
        \caption{The results of one pass pruning, which are plotted based on the results in Tables. The first row are the curves of reduction in weights vs.~test accuracy, and second row are the curves of pruned FLOPs vs.~test accuracy of (Pre)ResNet trained on CIFAR10 and CIFAR100 dataset under the settings of with and without $L_1$ sparsity on BatchNorm. The shaded areas represent the standard variance over five runs. }
        \label{fig:onepass_figures_resnet_cifar}
    \vspace{-0.3cm}
\end{figure*}

\begin{figure*}[t]
    \centering
    \begin{subfigure}[b]{0.3\linewidth}
        \includegraphics[width=\textwidth]{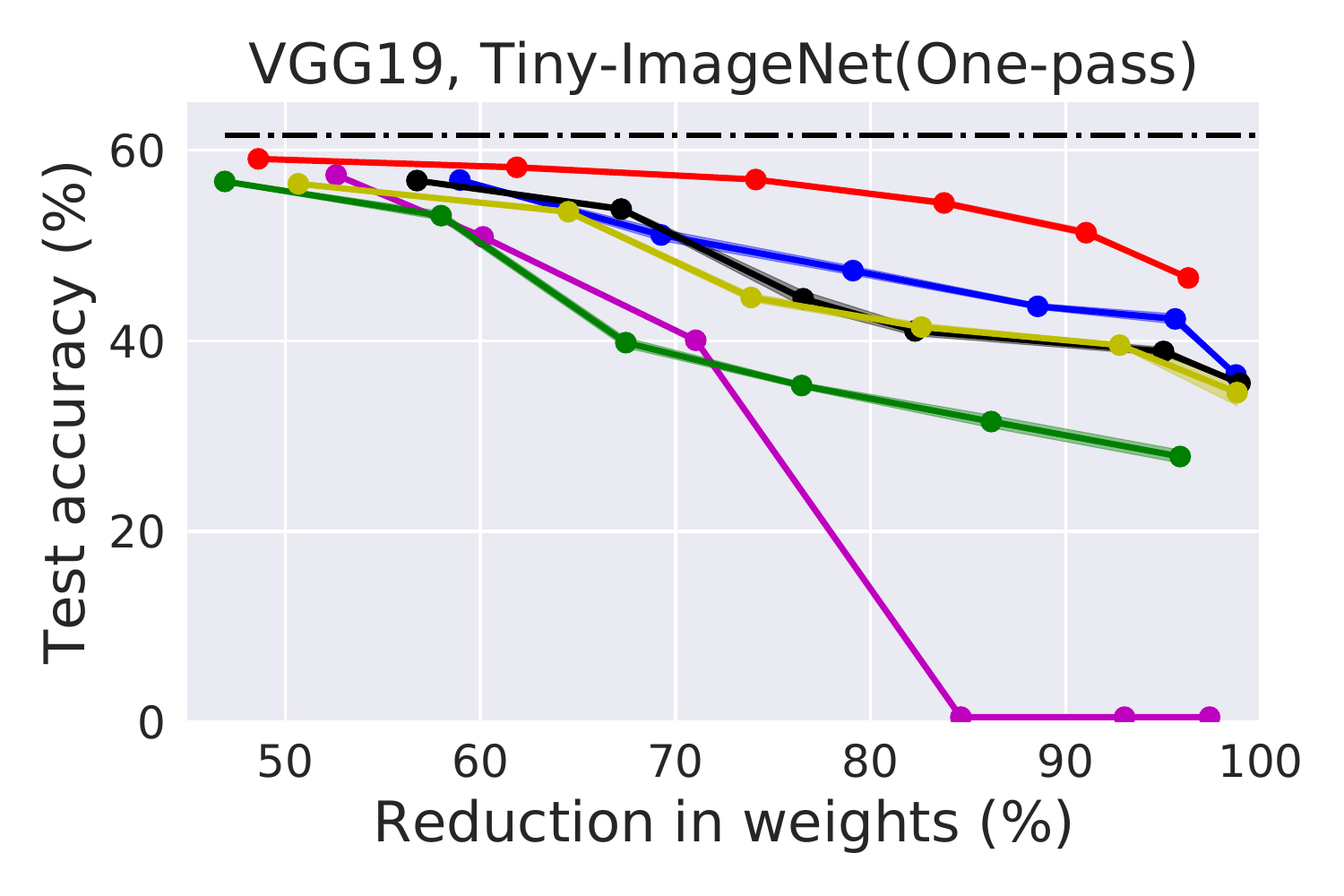}
    \end{subfigure}
    \begin{subfigure}[b]{0.3\linewidth}
        \includegraphics[width=\textwidth]{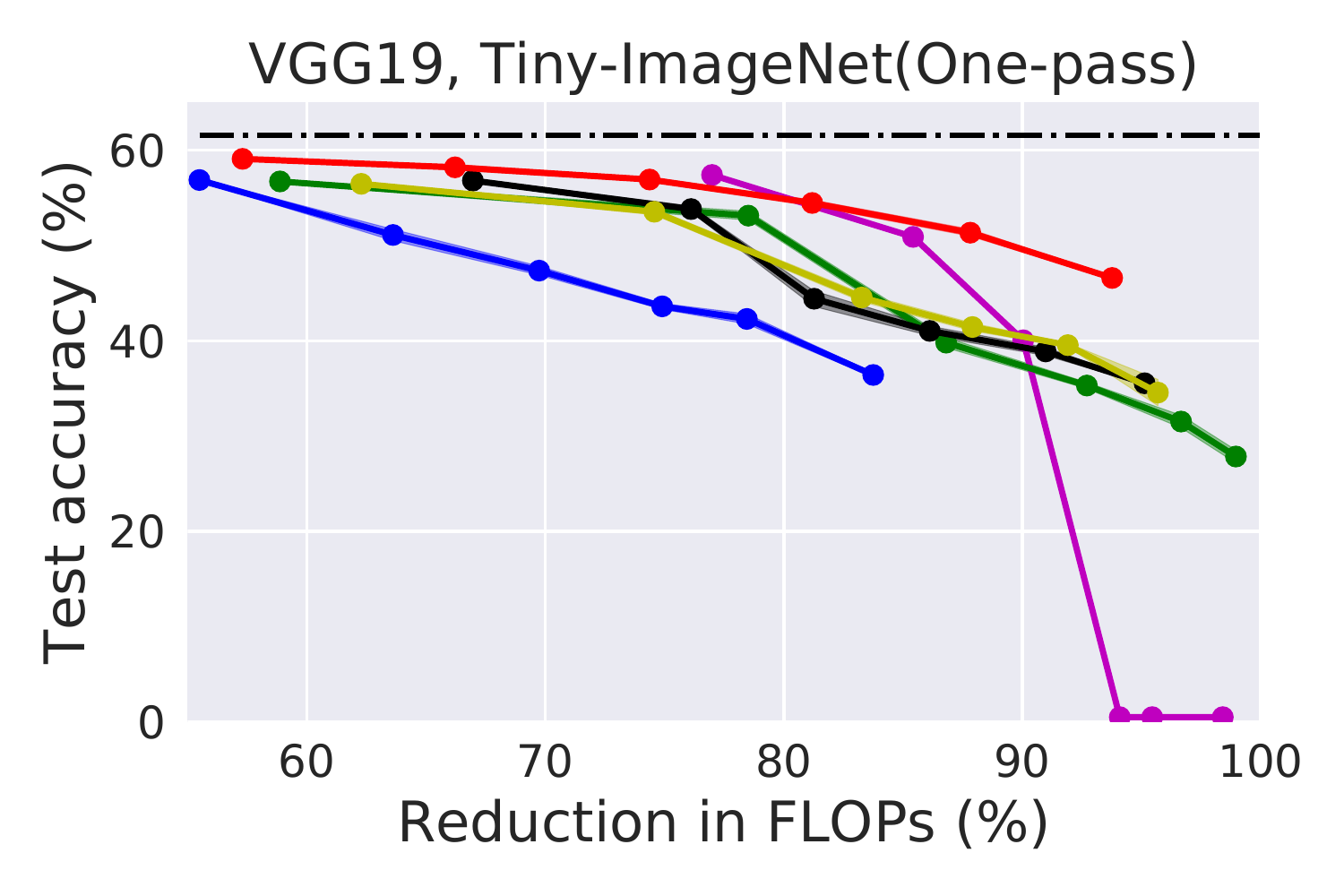}
    \end{subfigure}
    \vspace{-0.5cm}
    \caption{The results of one pass pruning, which are plotted based on the results in Tables. The base network for NN Slimming is pre-trained with $L_1$ sparsity on BatchNorm as required, and the others are normally pre-trained. }
    \label{fig:train_loss_fintune}
\vspace{-0.4cm}
\end{figure*}

\end{document}